\documentclass[11pt]{article}
\usepackage[a4paper,margin=2.5cm]{geometry}
\usepackage[T1]{fontenc}
\usepackage{lmodern}
\usepackage{mathptmx}
\usepackage{amsmath,amssymb}
\usepackage{booktabs,longtable,array,adjustbox}
\newcolumntype{P}[1]{>{\raggedright\arraybackslash}p{#1}}
\usepackage{graphicx}
\usepackage{xcolor}
\usepackage[numbers,sort&compress]{natbib}
\usepackage{url}
\usepackage{setspace}
\usepackage{xfp}
\usepackage{caption}
\DeclareCaptionLabelSeparator{mid}{ $\mid$ }
\usepackage{etoolbox}
\AtBeginEnvironment{table}{\footnotesize}
\usepackage[hidelinks]{hyperref}
\hypersetup{pdftitle={Spontaneous reasoning reorders language models' gains from chain-of-thought},pdfauthor={Dean Ariel, Eyal Hadad, Harel Navon, Tomer Hamuel, Ran Balicer, Seffi Cohen}}
\newcommand{\nm}[1]{#1}
\newcommand{\pp}{\,pp}

\newcommand{\pctof}[1]{\fpeval{round(100*(#1),0)}}
\providecommand{\aeBbhReversedPct}{18}
\providecommand{\aeGsmReversedPct}{20}
\providecommand{\aeHqPairs}{406}
\providecommand{\aeHqReversed}{142}
\providecommand{\aeHqReversedContent}{84}
\providecommand{\aeHqReversedPctContent}{21}
\providecommand{\aeMathGapFivePairs}{314}
\providecommand{\aeMathGapFiveReversed}{50}
\providecommand{\aeMathReversed}{84}
\providecommand{\aeMathReversedPct}{21}
\providecommand{\aeRecentM}{14}
\providecommand{\attrBbhLongRevShared}{27}
\providecommand{\attrBbhLongRevSharedContent}{77}
\providecommand{\attrBbhObsRev}{68}
\providecommand{\attrBbhObsRevContent}{85}
\providecommand{\attrBbhShortRevShared}{45}
\providecommand{\attrBbhShortRevSharedContent}{5}
\providecommand{\attrContentShareMax}{\nm{91}}
\providecommand{\attrContentShareMin}{\nm{78}}
\providecommand{\attrGsmLongRevShared}{58}
\providecommand{\attrGsmLongRevSharedContent}{68}
\providecommand{\attrGsmObsRev}{83}
\providecommand{\attrGsmObsRevContent}{87}
\providecommand{\attrGsmShortRevShared}{12}
\providecommand{\attrGsmShortRevSharedContent}{11}
\providecommand{\attrMathLongRevShared}{66}
\providecommand{\attrMathLongRevSharedContent}{69}
\providecommand{\attrMathObsRev}{84}
\providecommand{\attrMathObsRevContent}{80}
\providecommand{\attrMathShortRevShared}{13}
\providecommand{\attrMathShortRevSharedContent}{5}
\providecommand{\awGsmArmSwitchClauseFour}{0.279}
\providecommand{\awGsmArmSwitchClauseOne}{0.202}
\providecommand{\awGsmArmSwitchClauseThree}{0.224}
\providecommand{\awGsmArmSwitchClauseTwo}{0.279}
\providecommand{\awGsmClauseSwitchFdReversed}{0.013}
\providecommand{\awGsmClauseSwitchReqReversed}{0.132}
\providecommand{\awGsmConfigSwitchReqReversed}{0.057}
\providecommand{\awGsmDecArmExtra}{2.46}
\providecommand{\awGsmDecClauseExtra}{3.76}
\providecommand{\awGsmDecClauseExtraHi}{4.55}
\providecommand{\awGsmDecClauseExtraLo}{2.98}
\providecommand{\awGsmDecTriggerExtra}{0.00}
\providecommand{\awGsmGapClauseFourMean}{12.92}
\providecommand{\awGsmGapClauseOneMean}{8.04}
\providecommand{\awGsmGapClauseThreeMean}{6.78}
\providecommand{\awGsmGapClauseTwoMean}{17.79}
\providecommand{\awGsmTriggerSwitchReqReversed}{0.017}
\providecommand{\awMathArmSwitchClauseFour}{0.259}
\providecommand{\awMathArmSwitchClauseOne}{0.211}
\providecommand{\awMathArmSwitchClauseThree}{0.207}
\providecommand{\awMathArmSwitchClauseTwo}{0.230}
\providecommand{\awMathClauseSwitchFdReversed}{0.013}
\providecommand{\awMathClauseSwitchReqReversed}{0.093}
\providecommand{\awMathConfigSwitchReqReversed}{0.039}
\providecommand{\awMathDecArmExtra}{2.54}
\providecommand{\awMathDecClauseExtra}{0.13}
\providecommand{\awMathTriggerSwitchReqReversed}{0.019}
\providecommand{\awMechGsmArmClauses}{4}
\providecommand{\awMechGsmArmHoldsCount}{4}
\providecommand{\awMechGsmClauseObs}{13.2}
\providecommand{\awMechGsmClausePredHi}{14.9}
\providecommand{\awMechGsmClausePredLo}{10.3}
\providecommand{\awMechGsmFdLess}{28}
\providecommand{\awMechGsmModels}{29}
\providecommand{\awMechGsmScreenCollCorr}{0.99}
\providecommand{\awMechGsmSlopeModels}{15}
\providecommand{\awMechGsmSlopePositive}{15}
\providecommand{\awMechMathArmClauses}{4}
\providecommand{\awMechMathArmHoldsCount}{4}
\providecommand{\awMechMathClauseObs}{9.3}
\providecommand{\awMechMathClausePredHi}{9.8}
\providecommand{\awMechMathClausePredLo}{7.2}
\providecommand{\awMechMathFdLess}{28}
\providecommand{\awMechMathScreenCollCorr}{0.99}
\providecommand{\awMechMathSlopeModels}{17}
\providecommand{\awMechMathSlopePositive}{16}
\providecommand{\awxFirmText}{Give only the final answer. Do not show any working under any circumstances.}
\providecommand{\awxGsmCurFd}{25.1}
\providecommand{\awxGsmDefFd}{22.8}
\providecommand{\awxGsmDefSwitchPct}{49.8}
\providecommand{\awxGsmDefW}{1.00}
\providecommand{\awxGsmFirmSwitchPct}{16.3}
\providecommand{\awxGsmFirmW}{0.33}
\providecommand{\awxModels}{29}
\providecommand{\bbhBankConfigs}{64}
\providecommand{\bbhConflictBoxes}{1{,}305}
\providecommand{\bbhConflictDecMaxChange}{0.02}
\providecommand{\bbhConflictMaxRevChange}{0}
\providecommand{\bbhConflictOtherRevMaxChange}{1}
\providecommand{\bbhItemsPerTask}{54}
\providecommand{\bhM}{28}
\providecommand{\bhRben}{0.95}
\providecommand{\bmixBbhWMeanPct}{32}
\providecommand{\bmixDecBbhChoiceValCorr}{1.1}
\providecommand{\bmixDecBbhChoiceValShort}{0.8}
\providecommand{\bmixDecBbhVarLongPct}{53}
\providecommand{\bmixDecBbhVarLongPctContent}{95}
\providecommand{\bmixDecGsmChoiceValCorr}{2.4}
\providecommand{\bmixDecGsmChoiceValFd}{9.8}
\providecommand{\bmixDecGsmChoiceValShort}{1.7}
\providecommand{\bmixDecGsmShortGap}{$-$3.4}
\providecommand{\bmixDecGsmVarLongPct}{92}
\providecommand{\bmixDecMathChoiceValCorr}{0.8}
\providecommand{\bmixDecMathChoiceValFd}{13.2}
\providecommand{\bmixDecMathChoiceValShort}{1.0}
\providecommand{\bmixDecMathRecov}{0.47}
\providecommand{\bmixDecMathShortGap}{$-$0.1}
\providecommand{\bmixDecMathVarLongPct}{83}
\providecommand{\bmixGsmRevCorr}{48}
\providecommand{\bmixGsmRevObs}{83}
\providecommand{\bmixGsmWMaxPct}{100}
\providecommand{\bmixGsmWMeanPct}{43}
\providecommand{\bmixGsmWMinPct}{0}
\providecommand{\bmixMathLongCot}{46.9}
\providecommand{\bmixMathLongFd}{17.7}
\providecommand{\bmixMathLongReq}{38.0}
\providecommand{\bmixMathPairs}{406}
\providecommand{\bmixMathRevCorr}{24}
\providecommand{\bmixMathRevObs}{84}
\providecommand{\bmixMathShortFd}{27.9}
\providecommand{\bmixMathShortReq}{24.3}
\providecommand{\bmixMathWMeanPct}{43}
\providecommand{\capCells}{18}
\providecommand{\capCot}{3{,}072}
\providecommand{\capDiffExclCells}{5}
\providecommand{\capDirect}{1{,}024}
\providecommand{\capShallowerCells}{18}
\providecommand{\capSlopeBbhEciDiff}{$-$0.468}
\providecommand{\capSlopeBbhEciDiffFamHi}{$-$0.036}
\providecommand{\capSlopeBbhEciDiffFamLo}{$-$0.547}
\providecommand{\capSlopeBbhEciDiffHi}{$-$0.311}
\providecommand{\capSlopeBbhEciDiffLo}{$-$0.642}
\providecommand{\capSlopeBbhEciFd}{1.038}
\providecommand{\capSlopeBbhEciFdHi}{1.222}
\providecommand{\capSlopeBbhEciFdLo}{0.881}
\providecommand{\capSlopeBbhEciReq}{0.570}
\providecommand{\capSlopeBbhEciReqHi}{0.722}
\providecommand{\capSlopeBbhEciReqLo}{0.403}
\providecommand{\capSlopeGsmEciDiff}{$-$0.237}
\providecommand{\capSlopeGsmEciDiffFamHi}{0.053}
\providecommand{\capSlopeGsmEciDiffFamLo}{$-$0.593}
\providecommand{\capSlopeGsmEciDiffHi}{0.154}
\providecommand{\capSlopeGsmEciDiffLo}{$-$0.691}
\providecommand{\capSlopeGsmEciFd}{0.683}
\providecommand{\capSlopeGsmEciFdHi}{1.211}
\providecommand{\capSlopeGsmEciFdLo}{$-$0.102}
\providecommand{\capSlopeGsmEciReq}{0.446}
\providecommand{\capSlopeGsmEciReqHi}{1.061}
\providecommand{\capSlopeGsmEciReqLo}{$-$0.473}
\providecommand{\capSlopeMathEciDiff}{$-$0.277}
\providecommand{\capSlopeMathEciDiffFamHi}{0.156}
\providecommand{\capSlopeMathEciDiffFamLo}{$-$0.947}
\providecommand{\capSlopeMathEciDiffHi}{0.164}
\providecommand{\capSlopeMathEciDiffLo}{$-$0.770}
\providecommand{\capSlopeMathEciFd}{1.160}
\providecommand{\capSlopeMathEciFdHi}{1.475}
\providecommand{\capSlopeMathEciFdLo}{0.781}
\providecommand{\capSlopeMathEciReq}{0.884}
\providecommand{\capSlopeMathEciReqHi}{1.345}
\providecommand{\capSlopeMathEciReqLo}{0.343}
\providecommand{\capSlopeModelsBbh}{28}
\providecommand{\capSlopeModelsGsm}{29}
\providecommand{\claimsConstrainedN}{\nm{four}}
\providecommand{\claimsDirectN}{\nm{20}}
\providecommand{\claimsUnconstrainedN}{\nm{16}}
\providecommand{\combGapFivePairsBbh}{296}
\providecommand{\combGapFivePairsGsm}{429}
\providecommand{\combGapFivePairsHq}{133}
\providecommand{\combGapFivePairsMath}{442}
\providecommand{\combGapFivePairsPooled}{1{,}167}
\providecommand{\combGapFiveRevBbh}{12}
\providecommand{\combGapFiveRevGsm}{58}
\providecommand{\combGapFiveRevHq}{4}
\providecommand{\combGapFiveRevMath}{65}
\providecommand{\combGapFiveRevPooled}{135}
\providecommand{\combGapOnePairsBbh}{484}
\providecommand{\combGapOnePairsGsm}{565}
\providecommand{\combGapOnePairsHq}{422}
\providecommand{\combGapOnePairsMath}{569}
\providecommand{\combGapOnePairsPooled}{1{,}618}
\providecommand{\combGapOneRevBbh}{83}
\providecommand{\combGapOneRevGsm}{113}
\providecommand{\combGapOneRevHq}{84}
\providecommand{\combGapOneRevMath}{112}
\providecommand{\combGapOneRevPooled}{308}
\providecommand{\combGapTwoPairsBbh}{421}
\providecommand{\combGapTwoPairsGsm}{529}
\providecommand{\combGapTwoPairsHq}{324}
\providecommand{\combGapTwoPairsMath}{535}
\providecommand{\combGapTwoPairsPooled}{1{,}485}
\providecommand{\combGapTwoRevBbh}{50}
\providecommand{\combGapTwoRevGsm}{95}
\providecommand{\combGapTwoRevHq}{47}
\providecommand{\combGapTwoRevMath}{97}
\providecommand{\combGapTwoRevPooled}{242}
\providecommand{\combPanelRevBootHiBbh}{0.39}
\providecommand{\combPanelRevBootHiGsm}{0.33}
\providecommand{\combPanelRevBootHiHq}{0.33}
\providecommand{\combPanelRevBootHiMath}{0.33}
\providecommand{\combPanelRevBootHiPooled}{0.32}
\providecommand{\combPanelRevBootLoBbh}{0.09}
\providecommand{\combPanelRevBootLoGsm}{0.12}
\providecommand{\combPanelRevBootLoHq}{0.11}
\providecommand{\combPanelRevBootLoMath}{0.09}
\providecommand{\combPanelRevBootLoPooled}{0.12}
\providecommand{\combPanelRevPctBbh}{22}
\providecommand{\combPanelRevPctGsm}{21}
\providecommand{\combPanelRevPctHq}{21}
\providecommand{\combPanelRevPctMath}{20}
\providecommand{\combPanelRevPctPooled}{21}
\providecommand{\combRevBootHiBbh}{0.35}
\providecommand{\combRevBootHiBbhUncut}{0.35}
\providecommand{\combRevBootHiGsm}{0.31}
\providecommand{\combRevBootHiGsmUncut}{0.31}
\providecommand{\combRevBootHiHq}{0.41}
\providecommand{\combRevBootHiHqUncut}{0.41}
\providecommand{\combRevBootHiMath}{0.32}
\providecommand{\combRevBootHiMathUncut}{0.33}
\providecommand{\combRevBootHiPooled}{0.30}
\providecommand{\combRevBootHiPooledUncut}{0.30}
\providecommand{\combRevBootLoBbh}{0.10}
\providecommand{\combRevBootLoBbhUncut}{0.10}
\providecommand{\combRevBootLoGsm}{0.12}
\providecommand{\combRevBootLoGsmUncut}{0.13}
\providecommand{\combRevBootLoHq}{0.17}
\providecommand{\combRevBootLoHqUncut}{0.17}
\providecommand{\combRevBootLoMath}{0.11}
\providecommand{\combRevBootLoMathUncut}{0.11}
\providecommand{\combRevBootLoPooled}{0.13}
\providecommand{\combRevBootLoPooledUncut}{0.13}
\providecommand{\combRevPctBbh}{21}
\providecommand{\combRevPctBbhUncut}{21}
\providecommand{\combRevPctGsm}{21}
\providecommand{\combRevPctGsmUncut}{21}
\providecommand{\combRevPctHq}{28}
\providecommand{\combRevPctHqUncut}{28}
\providecommand{\combRevPctMath}{20}
\providecommand{\combRevPctMathUncut}{21}
\providecommand{\combRevPctPooled}{21}
\providecommand{\combRevPctPooledUncut}{21}
\providecommand{\crvBbhOwnDiff}{$-$0.499}
\providecommand{\crvBbhOwnDiffFamHi}{0.350}
\providecommand{\crvBbhOwnDiffFamLo}{$-$0.666}
\providecommand{\crvBbhOwnDiffHi}{$-$0.252}
\providecommand{\crvBbhOwnDiffLo}{$-$0.777}
\providecommand{\crvBbhOwnFd}{0.823}
\providecommand{\crvBbhOwnFdHi}{1.100}
\providecommand{\crvBbhOwnFdLo}{0.572}
\providecommand{\crvBbhOwnReq}{0.325}
\providecommand{\crvBbhOwnReqHi}{0.444}
\providecommand{\crvBbhOwnReqLo}{0.196}
\providecommand{\crvBbhOwnShare}{79}
\providecommand{\crvBbhOwnShareHi}{100}
\providecommand{\crvBbhOwnShareLo}{45}
\providecommand{\crvBbhReqAccDiff}{$-$0.349}
\providecommand{\crvBbhReqAccDiffFamHi}{$-$0.149}
\providecommand{\crvBbhReqAccDiffFamLo}{$-$0.395}
\providecommand{\crvBbhReqAccDiffHi}{$-$0.255}
\providecommand{\crvBbhReqAccDiffLo}{$-$0.452}
\providecommand{\crvBbhReqAccFd}{0.674}
\providecommand{\crvBbhReqAccFdHi}{0.789}
\providecommand{\crvBbhReqAccFdLo}{0.551}
\providecommand{\crvBbhReqAccReq}{0.325}
\providecommand{\crvBbhReqAccReqHi}{0.444}
\providecommand{\crvBbhReqAccReqLo}{0.196}
\providecommand{\crvGsmOwnDiff}{$-$0.663}
\providecommand{\crvGsmOwnDiffFamHi}{$-$0.494}
\providecommand{\crvGsmOwnDiffFamLo}{$-$1.205}
\providecommand{\crvGsmOwnDiffHi}{$-$0.347}
\providecommand{\crvGsmOwnDiffLo}{$-$1.160}
\providecommand{\crvGsmOwnFd}{0.343}
\providecommand{\crvGsmOwnFdHi}{1.065}
\providecommand{\crvGsmOwnFdLo}{$-$0.249}
\providecommand{\crvGsmOwnReq}{$-$0.320}
\providecommand{\crvGsmOwnReqHi}{0.210}
\providecommand{\crvGsmOwnReqLo}{$-$0.675}
\providecommand{\crvGsmOwnShare}{89}
\providecommand{\crvGsmOwnShareHi}{108}
\providecommand{\crvGsmOwnShareLo}{62}
\providecommand{\crvGsmReqAccDiff}{$-$0.491}
\providecommand{\crvGsmReqAccDiffFamHi}{$-$0.268}
\providecommand{\crvGsmReqAccDiffFamLo}{$-$0.753}
\providecommand{\crvGsmReqAccDiffHi}{$-$0.283}
\providecommand{\crvGsmReqAccDiffLo}{$-$0.681}
\providecommand{\crvGsmReqAccFd}{0.171}
\providecommand{\crvGsmReqAccFdHi}{0.611}
\providecommand{\crvGsmReqAccFdLo}{$-$0.174}
\providecommand{\crvGsmReqAccReq}{$-$0.320}
\providecommand{\crvGsmReqAccReqHi}{0.210}
\providecommand{\crvGsmReqAccReqLo}{$-$0.675}
\providecommand{\crvMathOwnDiff}{$-$1.033}
\providecommand{\crvMathOwnDiffFamHi}{$-$0.775}
\providecommand{\crvMathOwnDiffFamLo}{$-$1.889}
\providecommand{\crvMathOwnDiffHi}{$-$0.427}
\providecommand{\crvMathOwnDiffLo}{$-$1.501}
\providecommand{\crvMathOwnFd}{1.066}
\providecommand{\crvMathOwnFdHi}{1.591}
\providecommand{\crvMathOwnFdLo}{0.623}
\providecommand{\crvMathOwnReq}{0.033}
\providecommand{\crvMathOwnReqHi}{0.642}
\providecommand{\crvMathOwnReqLo}{$-$0.295}
\providecommand{\crvMathOwnShare}{102}
\providecommand{\crvMathOwnShareHi}{115}
\providecommand{\crvMathOwnShareLo}{94}
\providecommand{\crvMathReqAccDiff}{$-$0.552}
\providecommand{\crvMathReqAccDiffFamHi}{$-$0.422}
\providecommand{\crvMathReqAccDiffFamLo}{$-$0.751}
\providecommand{\crvMathReqAccDiffHi}{$-$0.300}
\providecommand{\crvMathReqAccDiffLo}{$-$0.680}
\providecommand{\crvMathReqAccFd}{0.585}
\providecommand{\crvMathReqAccFdHi}{0.974}
\providecommand{\crvMathReqAccFdLo}{0.350}
\providecommand{\crvMathReqAccReq}{0.033}
\providecommand{\crvMathReqAccReqHi}{0.642}
\providecommand{\crvMathReqAccReqLo}{$-$0.295}
\providecommand{\crvOwnShareStrictMax}{102}
\providecommand{\crvOwnShareStrictMin}{79}
\providecommand{\csBbhCotCompliantMean}{57.7}
\providecommand{\csBbhCotCorrectPct}{71.4}
\providecommand{\csBbhCotFormatGapMax}{77.1}
\providecommand{\csBbhCotFormatGapMean}{24.6}
\providecommand{\csBbhCotMalformedPct}{0.0}
\providecommand{\csBbhCotTruncatedPct}{0.3}
\providecommand{\csBbhCotUnreadablePct}{3.3}
\providecommand{\csBbhCotWrongPct}{24.9}
\providecommand{\csBbhForcedCompliantMean}{71.6}
\providecommand{\csBbhForcedCorrectPct}{52.7}
\providecommand{\csBbhForcedFormatGapMean}{12.7}
\providecommand{\csBbhForcedMalformedPct}{0.0}
\providecommand{\csBbhForcedTruncatedPct}{0.5}
\providecommand{\csBbhForcedUnreadablePct}{1.4}
\providecommand{\csBbhForcedWrongPct}{45.4}
\providecommand{\csBbhReqCompliantMean}{56.1}
\providecommand{\csBbhReqCorrectPct}{54.4}
\providecommand{\csBbhReqFormatGapMean}{20.5}
\providecommand{\csBbhReqMalformedPct}{0.0}
\providecommand{\csBbhReqTruncatedPct}{0.2}
\providecommand{\csBbhReqUnreadablePct}{2.2}
\providecommand{\csBbhReqWrongPct}{43.2}
\providecommand{\csGsmCotCompliantMean}{88.2}
\providecommand{\csGsmCotCorrectPct}{81.1}
\providecommand{\csGsmCotFormatGapMean}{4.1}
\providecommand{\csGsmCotMalformedPct}{0.0}
\providecommand{\csGsmCotTruncatedPct}{0.2}
\providecommand{\csGsmCotUnreadablePct}{0.1}
\providecommand{\csGsmCotWrongPct}{18.6}
\providecommand{\csGsmForcedCompliantMean}{94.1}
\providecommand{\csGsmForcedCorrectPct}{25.5}
\providecommand{\csGsmForcedFormatGapMean}{0.4}
\providecommand{\csGsmForcedMalformedPct}{0.0}
\providecommand{\csGsmForcedTruncatedPct}{1.0}
\providecommand{\csGsmForcedUnreadablePct}{0.1}
\providecommand{\csGsmForcedWrongPct}{73.4}
\providecommand{\csGsmReqCompliantMean}{85.2}
\providecommand{\csGsmReqCorrectPct}{35.1}
\providecommand{\csGsmReqFormatGapMean}{1.8}
\providecommand{\csGsmReqMalformedPct}{0.0}
\providecommand{\csGsmReqTruncatedPct}{0.1}
\providecommand{\csGsmReqUnreadablePct}{0.2}
\providecommand{\csGsmReqWrongPct}{64.7}
\providecommand{\cwCells}{131}
\providecommand{\cwCorrBbh}{9.9}
\providecommand{\cwCorrGsm}{1.5}
\providecommand{\cwCorrHq}{2.0}
\providecommand{\cwCorrMath}{3.6}
\providecommand{\cwCotAccBbh}{92.0}
\providecommand{\cwCotAccGsm}{96.3}
\providecommand{\cwCotAccHq}{66.0}
\providecommand{\cwCotAccMath}{98.9}
\providecommand{\cwFdAccBbh}{75.9}
\providecommand{\cwFdAccGsm}{90.2}
\providecommand{\cwFdAccHq}{63.9}
\providecommand{\cwFdAccMath}{83.9}
\providecommand{\cwFdGainBbh}{16.2}
\providecommand{\cwFdGainGsm}{6.1}
\providecommand{\cwFdGainHq}{2.0}
\providecommand{\cwFdGainMath}{15.0}
\providecommand{\cwFullCalls}{9{,}552}
\providecommand{\cwFullCorrMath}{3.4}
\providecommand{\cwFullFdGainMath}{16.0}
\providecommand{\cwFullFdGainMathHi}{19.5}
\providecommand{\cwFullFdGainMathLo}{12.5}
\providecommand{\cwFullItems}{398}
\providecommand{\cwFullReqGainMath}{2.7}
\providecommand{\cwFullReqGainMathHi}{3.9}
\providecommand{\cwFullReqGainMathLo}{1.5}
\providecommand{\cwFullTemplates}{8}
\providecommand{\cwFullWMathPct}{79}
\providecommand{\cwGainDiffBbh}{8.2}
\providecommand{\cwGainDiffBbhHi}{10.8}
\providecommand{\cwGainDiffBbhLo}{5.6}
\providecommand{\cwItems}{100}
\providecommand{\cwModels}{36}
\providecommand{\cwRepeatPairs}{20}
\providecommand{\cwRepeatSameSide}{17}
\providecommand{\cwReqAccBbh}{84.0}
\providecommand{\cwReqAccGsm}{94.9}
\providecommand{\cwReqAccHq}{63.5}
\providecommand{\cwReqAccMath}{95.5}
\providecommand{\cwReqGainBbh}{8.0}
\providecommand{\cwReqGainGsm}{1.4}
\providecommand{\cwReqGainHq}{2.4}
\providecommand{\cwReqGainMath}{3.3}
\providecommand{\cwRequests}{39{,}300}
\providecommand{\cwRequestsArm}{13{,}100}
\providecommand{\cwResidBbh}{$-$1.9}
\providecommand{\cwResidBbhContent}{$-$0.3}
\providecommand{\cwResidGsm}{$-$0.0}
\providecommand{\cwResidHq}{0.4}
\providecommand{\cwResidMath}{$-$0.3}
\providecommand{\cwWBbhPct}{39}
\providecommand{\cwWGsmPct}{76}
\providecommand{\cwWHqPct}{1}
\providecommand{\cwWMathPct}{76}
\providecommand{\ddirBbhFdToReq}{2.98}
\providecommand{\ddirBbhFdToReqContent}{1.24}
\providecommand{\ddirBbhFdToReqLofoMax}{2.67}
\providecommand{\ddirBbhFdToReqLofoMin}{$-$0.09}
\providecommand{\ddirBbhPilotFdToReq}{2.29}
\providecommand{\ddirBbhPilotReqToFd}{2.65}
\providecommand{\ddirBbhReqToFd}{2.42}
\providecommand{\ddirBbhReqToFdContent}{1.43}
\providecommand{\ddirBbhRevFamBootHi}{0.25}
\providecommand{\ddirBbhRevFamBootLo}{0.12}
\providecommand{\ddirBbhRevWithinFam}{17}
\providecommand{\ddirBbhRevWithinFamPairs}{81}
\providecommand{\ddirGsmFdToReq}{8.80}
\providecommand{\ddirGsmFdToReqContent}{8.34}
\providecommand{\ddirGsmFdToReqLofoMax}{11.92}
\providecommand{\ddirGsmFdToReqLofoMin}{4.40}
\providecommand{\ddirGsmPilotFdToReq}{8.43}
\providecommand{\ddirGsmPilotReqToFd}{2.06}
\providecommand{\ddirGsmReqToFdContent}{2.70}
\providecommand{\ddirGsmRevFamBootHi}{0.39}
\providecommand{\ddirGsmRevFamBootLo}{0.10}
\providecommand{\ddirGsmRevWithinFam}{23}
\providecommand{\ddirGsmRevWithinFamPairs}{88}
\providecommand{\ddirMathFdToReq}{13.08}
\providecommand{\ddirMathFdToReqContent}{11.48}
\providecommand{\ddirMathFdToReqLofoMax}{15.47}
\providecommand{\ddirMathFdToReqLofoMin}{8.24}
\providecommand{\ddirMathPilotFdToReq}{12.75}
\providecommand{\ddirMathPilotReqToFd}{2.57}
\providecommand{\ddirMathReqToFd}{3.00}
\providecommand{\ddirMathReqToFdContent}{2.98}
\providecommand{\ddirMathRevFamBootHi}{0.35}
\providecommand{\ddirMathRevFamBootLo}{0.07}
\providecommand{\ddirMathRevWithinFam}{25}
\providecommand{\ddirMathRevWithinFamPairs}{88}
\providecommand{\decBbhReqBestThird}{26.47}
\providecommand{\decGsmReqBestThird}{68.93}
\providecommand{\decMathReqBestThird}{56.01}
\providecommand{\diffModelsBbhSub}{27}
\providecommand{\diffModelsGsm}{28}
\providecommand{\diffModelsMath}{29}
\providecommand{\diffOrBbhSub}{1.07}
\providecommand{\diffOrBbhSubHi}{1.09}
\providecommand{\diffOrBbhSubLo}{1.05}
\providecommand{\diffOrGsm}{1.11}
\providecommand{\diffOrGsmHi}{1.15}
\providecommand{\diffOrGsmLo}{1.06}
\providecommand{\diffOrMath}{1.11}
\providecommand{\diffOrMathHi}{1.17}
\providecommand{\diffOrMathLo}{1.06}
\providecommand{\diffPosBbhSub}{24}
\providecommand{\diffPosGsm}{23}
\providecommand{\diffPosMath}{22}
\providecommand{\dtEligible}{30}
\providecommand{\fmtHqReqPlaceholderPct}{5.5}
\providecommand{\fmtHqReqUnboxedPct}{20.1}
\providecommand{\fsCheckBoxes}{15{,}474}
\providecommand{\fsCheckDiffer}{2}
\providecommand{\fsWBbhCompFsPct}{99.0}
\providecommand{\fsWBbhCompZsPct}{56.0}
\providecommand{\fsWBbhValGainReqContent}{1.42}
\providecommand{\fsWGsmM}{29}
\providecommand{\fsWGsmRho}{0.44}
\providecommand{\fsWGsmRhoHi}{0.75}
\providecommand{\fsWGsmRhoLo}{0.06}
\providecommand{\fsWGsmWDownN}{28}
\providecommand{\fsWGsmWFsPct}{4.0}
\providecommand{\fsWGsmWZsPct}{43.7}
\providecommand{\fsrBbhFdToReq}{3.42}
\providecommand{\fsrBbhGainFd}{5.65}
\providecommand{\fsrBbhGainReq}{21.72}
\providecommand{\fsrBbhRev}{99}
\providecommand{\fsrBbhRevBootHi}{37.4}
\providecommand{\fsrBbhRevBootLo}{16.0}
\providecommand{\fsrBbhRevPairs}{378}
\providecommand{\fsrBbhTplReq}{0.069}
\providecommand{\fsrGsmFdToReq}{7.91}
\providecommand{\fsrGsmGainFd}{2.87}
\providecommand{\fsrGsmGainReq}{$-$5.67}
\providecommand{\fsrGsmRev}{181}
\providecommand{\fsrGsmRevBootHi}{58.9}
\providecommand{\fsrGsmRevBootLo}{32.0}
\providecommand{\fsrGsmRevPairs}{406}
\providecommand{\fsrGsmTplReq}{0.100}
\providecommand{\genThiGsmArmM}{8}
\providecommand{\genThiGsmArmQwen}{7}
\providecommand{\genThiGsmArmRev}{11}
\providecommand{\genThiGsmArmRevPairs}{28}
\providecommand{\gsmBankConfigs}{56}
\providecommand{\gtKdisc}{24}
\providecommand{\gtKval}{24}
\providecommand{\gtM}{29}
\providecommand{\gtRben}{0.98}
\providecommand{\hqArmAccFd}{51.5}
\providecommand{\hqArmAccFdContent}{51.5}
\providecommand{\hqArmAccReq}{41.8}
\providecommand{\hqArmAccReqContent}{49.3}
\providecommand{\hqDdirFdToReq}{5.50}
\providecommand{\hqDdirFdToReqContentPostHoc}{0.45}
\providecommand{\hqDdirFdToReqHiContentPostHoc}{0.63}
\providecommand{\hqDdirFdToReqLoContentPostHoc}{0.28}
\providecommand{\hqDdirPilotFdToReqContentPostHoc}{0.27}
\providecommand{\hqDdirPilotReqToFdContentPostHoc}{1.10}
\providecommand{\hqDdirReqToFd}{3.63}
\providecommand{\hqDdirReqToFdContentPostHoc}{1.68}
\providecommand{\hqDdirReqToFdHiContentPostHoc}{1.77}
\providecommand{\hqDdirReqToFdLoContentPostHoc}{1.60}
\providecommand{\hqDecRegretCorr}{6.3}
\providecommand{\hqDecRegretCorrContent}{0.7}
\providecommand{\hqDecRegretFd}{6.3}
\providecommand{\hqDecRegretFdContent}{0.7}
\providecommand{\hqDecVarLongPct}{1}
\providecommand{\hqDecVarLongPctContent}{2}
\providecommand{\hqDecVarLongPctSixteen}{21}
\providecommand{\hqDecVarLongPctSixteenContent}{60}
\providecommand{\hqDecVarThresholdPct}{50}
\providecommand{\hqFdMeanGainContent}{$-$1.3}
\providecommand{\hqFdTrObsC}{0.908}
\providecommand{\hqFdTrObsCContent}{0.903}
\providecommand{\hqFdTrPredC}{0.917}
\providecommand{\hqFreshConfigs}{13}
\providecommand{\hqFreshDraws}{15}
\providecommand{\hqGapCotFdMedianContent}{0.7}
\providecommand{\hqGapReqFdMedian}{$-$7.0}
\providecommand{\hqGapReqFdMedianContent}{$-$1.1}
\providecommand{\hqItems}{1{,}000}
\providecommand{\hqM}{29}
\providecommand{\hqReqKdisc}{24}
\providecommand{\hqReqKval}{24}
\providecommand{\hqReqMeanGainContent}{0.9}
\providecommand{\hqReqTrObsC}{0.884}
\providecommand{\hqReqTrObsCContent}{0.866}
\providecommand{\hqReqTrPredC}{0.853}
\providecommand{\hqSplitBestSingleWorst}{4.14}
\providecommand{\hqSplitBestSingleWorstContentPostHoc}{1.92}
\providecommand{\hqSplitSplitWorst}{3.49}
\providecommand{\hqSplitSplitWorstContentPostHoc}{1.74}
\providecommand{\hqTrNewDrawReversed}{0.099}
\providecommand{\hqTrNewDrawReversedContent}{0.109}
\providecommand{\hqTrSwitchReversed}{0.320}
\providecommand{\hqTrSwitchReversedContent}{0.194}
\providecommand{\hraAnswers}{384}
\providecommand{\hraRaterAAnswers}{242}
\providecommand{\hraRaterACtrlWrongNAll}{23}
\providecommand{\hraRaterACtrlWrongReadAll}{1}
\providecommand{\hraRaterBCtrlWrongNAll}{27}
\providecommand{\hraRaterBCtrlWrongReadAll}{12}
\providecommand{\hraShared}{100}
\providecommand{\hrtABDisagree}{16}
\providecommand{\hrtABSideA}{14}
\providecommand{\hrtABSideB}{1}
\providecommand{\hrtAcOneAB}{0.82}
\providecommand{\hrtAcOneAC}{0.96}
\providecommand{\hrtAgreePctAB}{84.0}
\providecommand{\hrtAgreePctAC}{97.0}
\providecommand{\hrtAgreePctBC}{88.8}
\providecommand{\hrtAnswersC}{242}
\providecommand{\hrtDisagreeContentAgreePct}{98.1}
\providecommand{\hrtDisagreeContentAgreePctHi}{99.3}
\providecommand{\hrtDisagreeContentAgreePctLo}{95.3}
\providecommand{\hrtDisagreeN}{215}
\providecommand{\hrtKappaAB}{0.38}
\providecommand{\hrtKappaAC}{0.90}
\providecommand{\hrtKappaBC}{0.46}
\providecommand{\hrtMajTwoOfThree}{16}
\providecommand{\hrtMajUnanimous}{83}
\providecommand{\hrtRaterCContentAgreePct}{99.5}
\providecommand{\hrtRaterCCtrlWrongNAll}{26}
\providecommand{\hrtRaterCCtrlWrongReadAll}{0}
\providecommand{\hrtSharedThree}{100}
\providecommand{\hrtWithBDisagreeContentAgreePct}{95.3}
\providecommand{\labAgreeJudgeRaterOnePct}{84}
\providecommand{\labAgreeJudgeRaterTwoPct}{85}
\providecommand{\labAgreeRaterOneRaterTwoPct}{82}
\providecommand{\labCostUsd}{12.64}
\providecommand{\labGapReqCotRGsm}{$-$5.9}
\providecommand{\labGapReqCotRMath}{$-$7.5}
\providecommand{\labGapReqForcedNotRGsm}{$-$5.5}
\providecommand{\labGapReqForcedNotRMath}{$-$4.2}
\providecommand{\labHumanN}{150}
\providecommand{\labJudge}{Claude Opus 5}
\providecommand{\labKappaJudgeRaterOne}{0.71}
\providecommand{\labKappaJudgeRaterTwo}{0.72}
\providecommand{\labKappaRaterOneRaterTwo}{0.67}
\providecommand{\labN}{2{,}525}
\providecommand{\labNQwen}{110}
\providecommand{\labNUnread}{4}
\providecommand{\labQwenNLong}{80}
\providecommand{\labRLongBbhPct}{44}
\providecommand{\labRLongGsmPct}{63}
\providecommand{\labRLongMathPct}{68}
\providecommand{\labRShortGsmPct}{1}
\providecommand{\labRShortHqSixtyFourPct}{6}
\providecommand{\labRecallHqSixtyFourPct}{13}
\providecommand{\labRsqRGsm}{0.66}
\providecommand{\labRsqWGsm}{0.41}
\providecommand{\labSelfAgreeUnwPct}{99}
\providecommand{\labSelfKappaUnw}{0.98}
\providecommand{\labSelfN}{150}
\providecommand{\ladBigGapFivePairsBbh}{1}
\providecommand{\ladBigGapFivePairsGsm}{5}
\providecommand{\ladBigGapFivePairsHq}{0}
\providecommand{\ladBigGapFivePairsMath}{3}
\providecommand{\ladBigGapFivePairsPooled}{9}
\providecommand{\ladBigGapFiveRevBbh}{0}
\providecommand{\ladBigGapFiveRevGsm}{1}
\providecommand{\ladBigGapFiveRevHq}{0}
\providecommand{\ladBigGapFiveRevMath}{2}
\providecommand{\ladBigGapFiveRevPooled}{3}
\providecommand{\ladBigGapOnePairsBbh}{2}
\providecommand{\ladBigGapOnePairsGsm}{6}
\providecommand{\ladBigGapOnePairsHq}{3}
\providecommand{\ladBigGapOnePairsMath}{5}
\providecommand{\ladBigGapOnePairsPooled}{13}
\providecommand{\ladBigGapOneRevBbh}{0}
\providecommand{\ladBigGapOneRevGsm}{2}
\providecommand{\ladBigGapOneRevHq}{3}
\providecommand{\ladBigGapOneRevMath}{3}
\providecommand{\ladBigGapOneRevPooled}{5}
\providecommand{\ladBigGapTwoPairsBbh}{2}
\providecommand{\ladBigGapTwoPairsGsm}{5}
\providecommand{\ladBigGapTwoPairsHq}{1}
\providecommand{\ladBigGapTwoPairsMath}{5}
\providecommand{\ladBigGapTwoPairsPooled}{12}
\providecommand{\ladBigGapTwoRevBbh}{0}
\providecommand{\ladBigGapTwoRevGsm}{1}
\providecommand{\ladBigGapTwoRevHq}{1}
\providecommand{\ladBigGapTwoRevMath}{3}
\providecommand{\ladBigGapTwoRevPooled}{4}
\providecommand{\ladBigPairsBbh}{6}
\providecommand{\ladBigPairsGsm}{6}
\providecommand{\ladBigPairsHq}{6}
\providecommand{\ladBigPairsMath}{6}
\providecommand{\ladBigPairsPooled}{18}
\providecommand{\ladBigRevBbh}{2}
\providecommand{\ladBigRevBbhStrict}{2}
\providecommand{\ladBigRevBbhUncut}{2}
\providecommand{\ladBigRevGsm}{2}
\providecommand{\ladBigRevGsmStrict}{1}
\providecommand{\ladBigRevGsmUncut}{2}
\providecommand{\ladBigRevHq}{6}
\providecommand{\ladBigRevHqStrict}{6}
\providecommand{\ladBigRevHqUncut}{6}
\providecommand{\ladBigRevMath}{3}
\providecommand{\ladBigRevMathStrict}{4}
\providecommand{\ladBigRevMathUncut}{2}
\providecommand{\ladBigRevPooled}{7}
\providecommand{\ladBigRevPooledStrict}{7}
\providecommand{\ladBigRevPooledUncut}{6}
\providecommand{\ladCells}{24}
\providecommand{\ladDsFsNoAnswerPct}{20.0}
\providecommand{\ladEquivLlamaSame}{50}
\providecommand{\ladEquivQwenSame}{1}
\providecommand{\ladEquivQwenVerdict}{41}
\providecommand{\ladEquivRows}{50}
\providecommand{\ladGapFivePairsBbh}{6}
\providecommand{\ladGapFivePairsGsm}{13}
\providecommand{\ladGapFivePairsHq}{0}
\providecommand{\ladGapFivePairsMath}{9}
\providecommand{\ladGapFivePairsPooled}{28}
\providecommand{\ladGapFiveRevBbh}{0}
\providecommand{\ladGapFiveRevGsm}{1}
\providecommand{\ladGapFiveRevHq}{0}
\providecommand{\ladGapFiveRevMath}{2}
\providecommand{\ladGapFiveRevPooled}{3}
\providecommand{\ladGapOnePairsBbh}{10}
\providecommand{\ladGapOnePairsGsm}{15}
\providecommand{\ladGapOnePairsHq}{7}
\providecommand{\ladGapOnePairsMath}{13}
\providecommand{\ladGapOnePairsPooled}{38}
\providecommand{\ladGapOneRevBbh}{1}
\providecommand{\ladGapOneRevGsm}{2}
\providecommand{\ladGapOneRevHq}{6}
\providecommand{\ladGapOneRevMath}{4}
\providecommand{\ladGapOneRevPooled}{7}
\providecommand{\ladGapTwoPairsBbh}{9}
\providecommand{\ladGapTwoPairsGsm}{14}
\providecommand{\ladGapTwoPairsHq}{1}
\providecommand{\ladGapTwoPairsMath}{13}
\providecommand{\ladGapTwoPairsPooled}{36}
\providecommand{\ladGapTwoRevBbh}{0}
\providecommand{\ladGapTwoRevGsm}{1}
\providecommand{\ladGapTwoRevHq}{1}
\providecommand{\ladGapTwoRevMath}{4}
\providecommand{\ladGapTwoRevPooled}{5}
\providecommand{\ladInRangeCells}{22}
\providecommand{\ladNebiusPairs}{35}
\providecommand{\ladNebiusSameAnswer}{35}
\providecommand{\ladPairsBbh}{15}
\providecommand{\ladPairsGsm}{15}
\providecommand{\ladPairsHq}{15}
\providecommand{\ladPairsMath}{15}
\providecommand{\ladPairsPooled}{45}
\providecommand{\ladQwenReqBareBbh}{10{,}396}
\providecommand{\ladQwenReqBareGsm}{31{,}129}
\providecommand{\ladQwenReqBareHq}{24{,}706}
\providecommand{\ladQwenReqBareMath}{10{,}318}
\providecommand{\ladQwenReqBareTokensBbh}{3.0}
\providecommand{\ladQwenReqBareTokensGsm}{3.0}
\providecommand{\ladQwenReqBareTokensHq}{5.0}
\providecommand{\ladQwenReqBareTokensMath}{3.0}
\providecommand{\ladQwenReqRowsBbh}{47{,}250}
\providecommand{\ladQwenReqRowsGsm}{38{,}251}
\providecommand{\ladQwenReqRowsHq}{36{,}000}
\providecommand{\ladQwenReqRowsMath}{12{,}338}
\providecommand{\ladQwenReqUnreadPctBbh}{24.1}
\providecommand{\ladQwenReqUnreadPctGsm}{81.4}
\providecommand{\ladQwenReqUnreadPctHq}{68.6}
\providecommand{\ladQwenReqUnreadPctMath}{84.6}
\providecommand{\ladRequests}{556{,}708}
\providecommand{\ladRevBbh}{3}
\providecommand{\ladRevBbhStrict}{7}
\providecommand{\ladRevBbhStrictUncut}{7}
\providecommand{\ladRevBbhUncut}{3}
\providecommand{\ladRevGsm}{2}
\providecommand{\ladRevGsmStrict}{3}
\providecommand{\ladRevGsmStrictUncut}{3}
\providecommand{\ladRevGsmUncut}{2}
\providecommand{\ladRevHq}{12}
\providecommand{\ladRevHqStrict}{12}
\providecommand{\ladRevHqStrictUncut}{12}
\providecommand{\ladRevHqUncut}{12}
\providecommand{\ladRevMath}{4}
\providecommand{\ladRevMathStrict}{7}
\providecommand{\ladRevMathStrictUncut}{6}
\providecommand{\ladRevMathUncut}{3}
\providecommand{\ladRevPooled}{9}
\providecommand{\ladRevPooledStrict}{17}
\providecommand{\ladRevPooledStrictUncut}{16}
\providecommand{\ladRevPooledUncut}{8}
\providecommand{\ladShiftDeepSeekBbh}{5.9}
\providecommand{\ladShiftDeepSeekGsm}{2.0}
\providecommand{\ladShiftDeepSeekHq}{7.6}
\providecommand{\ladShiftDeepSeekMath}{15.0}
\providecommand{\ladShiftLlamaSeventyBbh}{0.0}
\providecommand{\ladShiftLlamaSeventyGsm}{21.8}
\providecommand{\ladShiftLlamaSeventyHq}{$-$0.5}
\providecommand{\ladShiftLlamaSeventyMath}{2.0}
\providecommand{\ladShiftLongDeepSeekBbh}{3.0}
\providecommand{\ladShiftLongDeepSeekGsm}{2.9}
\providecommand{\ladShiftLongDeepSeekHq}{1.0}
\providecommand{\ladShiftLongDeepSeekMath}{15.4}
\providecommand{\ladShiftLongLlamaSeventyBbh}{1.4}
\providecommand{\ladShiftLongLlamaSeventyGsm}{24.4}
\providecommand{\ladShiftLongLlamaSeventyHq}{$-$0.4}
\providecommand{\ladShiftLongLlamaSeventyMath}{1.8}
\providecommand{\ladShiftLongMinistralEightBbh}{0.0}
\providecommand{\ladShiftLongMinistralEightGsm}{0.0}
\providecommand{\ladShiftLongMinistralEightHq}{0.9}
\providecommand{\ladShiftLongMinistralEightMath}{10.8}
\providecommand{\ladShiftLongMinistralFourteenBbh}{0.0}
\providecommand{\ladShiftLongMinistralFourteenGsm}{0.2}
\providecommand{\ladShiftLongMinistralFourteenHq}{$-$0.3}
\providecommand{\ladShiftLongMinistralFourteenMath}{5.6}
\providecommand{\ladShiftLongMistralLargeBbh}{0.1}
\providecommand{\ladShiftLongMistralLargeGsm}{0.0}
\providecommand{\ladShiftLongMistralLargeHq}{$-$0.1}
\providecommand{\ladShiftLongMistralLargeMath}{5.1}
\providecommand{\ladShiftLongQwenOneTwentyTwoBbh}{0.0}
\providecommand{\ladShiftLongQwenOneTwentyTwoGsm}{0.0}
\providecommand{\ladShiftLongQwenOneTwentyTwoHq}{$-$0.6}
\providecommand{\ladShiftLongQwenOneTwentyTwoMath}{0.0}
\providecommand{\ladShiftMinistralEightBbh}{3.9}
\providecommand{\ladShiftMinistralEightGsm}{$-$0.3}
\providecommand{\ladShiftMinistralEightHq}{2.8}
\providecommand{\ladShiftMinistralEightMath}{11.5}
\providecommand{\ladShiftMinistralFourteenBbh}{7.0}
\providecommand{\ladShiftMinistralFourteenGsm}{0.1}
\providecommand{\ladShiftMinistralFourteenHq}{4.0}
\providecommand{\ladShiftMinistralFourteenMath}{5.5}
\providecommand{\ladShiftMistralLargeBbh}{2.2}
\providecommand{\ladShiftMistralLargeGsm}{0.2}
\providecommand{\ladShiftMistralLargeHq}{3.4}
\providecommand{\ladShiftMistralLargeMath}{5.6}
\providecommand{\ladShiftPanelMaxBbh}{16.3}
\providecommand{\ladShiftPanelMaxGsm}{60.7}
\providecommand{\ladShiftPanelMaxHq}{6.4}
\providecommand{\ladShiftPanelMaxMath}{49.7}
\providecommand{\ladShiftPanelMinBbh}{$-$4.9}
\providecommand{\ladShiftPanelMinGsm}{$-$13.1}
\providecommand{\ladShiftPanelMinHq}{$-$20.4}
\providecommand{\ladShiftPanelMinMath}{$-$14.4}
\providecommand{\ladShiftQwenOneTwentyTwoBbh}{$-$1.1}
\providecommand{\ladShiftQwenOneTwentyTwoGsm}{$-$11.0}
\providecommand{\ladShiftQwenOneTwentyTwoHq}{$-$1.6}
\providecommand{\ladShiftQwenOneTwentyTwoMath}{$-$9.6}
\providecommand{\ladShiftShortDeepSeekBbh}{2.9}
\providecommand{\ladShiftShortDeepSeekGsm}{$-$0.8}
\providecommand{\ladShiftShortDeepSeekHq}{6.6}
\providecommand{\ladShiftShortDeepSeekMath}{$-$0.5}
\providecommand{\ladShiftShortLlamaSeventyBbh}{$-$1.4}
\providecommand{\ladShiftShortLlamaSeventyGsm}{$-$2.5}
\providecommand{\ladShiftShortLlamaSeventyHq}{$-$0.1}
\providecommand{\ladShiftShortLlamaSeventyMath}{0.2}
\providecommand{\ladShiftShortMinistralEightBbh}{3.8}
\providecommand{\ladShiftShortMinistralEightGsm}{$-$0.3}
\providecommand{\ladShiftShortMinistralEightHq}{1.9}
\providecommand{\ladShiftShortMinistralEightMath}{0.8}
\providecommand{\ladShiftShortMinistralFourteenBbh}{7.0}
\providecommand{\ladShiftShortMinistralFourteenGsm}{$-$0.1}
\providecommand{\ladShiftShortMinistralFourteenHq}{4.2}
\providecommand{\ladShiftShortMinistralFourteenMath}{$-$0.1}
\providecommand{\ladShiftShortMistralLargeBbh}{2.1}
\providecommand{\ladShiftShortMistralLargeGsm}{0.2}
\providecommand{\ladShiftShortMistralLargeHq}{3.5}
\providecommand{\ladShiftShortMistralLargeMath}{0.5}
\providecommand{\ladShiftShortQwenOneTwentyTwoBbh}{$-$1.1}
\providecommand{\ladShiftShortQwenOneTwentyTwoGsm}{$-$10.9}
\providecommand{\ladShiftShortQwenOneTwentyTwoHq}{$-$1.1}
\providecommand{\ladShiftShortQwenOneTwentyTwoMath}{$-$9.6}
\providecommand{\ladSizeDeepSeekActiveInB}{8}
\providecommand{\ladSizeDeepSeekActiveOutB}{16}
\providecommand{\ladSizeDeepSeekTotalB}{552}
\providecommand{\ladSizeMaxB}{675}
\providecommand{\ladSizeMinistralEightB}{8.9}
\providecommand{\ladSizeMinistralFourteenB}{13.9}
\providecommand{\ladSizeMistralLargeActiveB}{41}
\providecommand{\ladSizeQwenActiveB}{10}
\providecommand{\ladSizeQwenB}{125}
\providecommand{\ladWDeepSeekBbh}{0.109}
\providecommand{\ladWDeepSeekGsm}{0.097}
\providecommand{\ladWDeepSeekHq}{0.003}
\providecommand{\ladWDeepSeekMath}{0.385}
\providecommand{\ladWLlamaEightBbh}{0.029}
\providecommand{\ladWLlamaEightGsm}{0.274}
\providecommand{\ladWLlamaEightHq}{0.013}
\providecommand{\ladWLlamaEightMath}{0.108}
\providecommand{\ladWLlamaSeventyBbh}{0.197}
\providecommand{\ladWLlamaSeventyGsm}{0.414}
\providecommand{\ladWLlamaSeventyHq}{0.006}
\providecommand{\ladWLlamaSeventyMath}{0.079}
\providecommand{\ladWMinistralEightBbh}{0.002}
\providecommand{\ladWMinistralEightGsm}{0.005}
\providecommand{\ladWMinistralEightHq}{0.004}
\providecommand{\ladWMinistralEightMath}{0.310}
\providecommand{\ladWMinistralFourteenBbh}{0.001}
\providecommand{\ladWMinistralFourteenGsm}{0.024}
\providecommand{\ladWMinistralFourteenHq}{0.004}
\providecommand{\ladWMinistralFourteenMath}{0.341}
\providecommand{\ladWMistralLargeBbh}{0.006}
\providecommand{\ladWMistralLargeGsm}{0.003}
\providecommand{\ladWMistralLargeHq}{0.003}
\providecommand{\ladWMistralLargeMath}{0.112}
\providecommand{\ladWPanelMaxBbh}{1.000}
\providecommand{\ladWPanelMaxGsm}{1.000}
\providecommand{\ladWPanelMaxHq}{0.109}
\providecommand{\ladWPanelMaxMath}{1.000}
\providecommand{\ladWPanelMedianBbh}{0.213}
\providecommand{\ladWPanelMedianGsm}{0.310}
\providecommand{\ladWPanelMedianHq}{0.007}
\providecommand{\ladWPanelMedianMath}{0.286}
\providecommand{\ladWPanelMinBbh}{0.008}
\providecommand{\ladWPanelMinGsm}{0.001}
\providecommand{\ladWPanelMinHq}{0.000}
\providecommand{\ladWPanelMinMath}{0.002}
\providecommand{\ladWQwenNineBbh}{0.335}
\providecommand{\ladWQwenNineGsm}{0.043}
\providecommand{\ladWQwenNineHq}{0.003}
\providecommand{\ladWQwenNineMath}{0.127}
\providecommand{\ladWQwenOneTwentyTwoBbh}{0.015}
\providecommand{\ladWQwenOneTwentyTwoGsm}{0.003}
\providecommand{\ladWQwenOneTwentyTwoHq}{0.004}
\providecommand{\ladWQwenOneTwentyTwoMath}{0.007}
\providecommand{\ladWQwenThirtyFiveBbh}{0.015}
\providecommand{\ladWQwenThirtyFiveGsm}{0.001}
\providecommand{\ladWQwenThirtyFiveHq}{0.007}
\providecommand{\ladWQwenThirtyFiveMath}{0.006}
\providecommand{\mtItems}{398}
\providecommand{\mtRben}{0.99}
\providecommand{\patSpragueBaseSymMinusLang}{14.5}
\providecommand{\patSpragueContentSymMinusLang}{16.8}
\providecommand{\patSpragueForcedSymMinusLang}{18.2}
\providecommand{\patSpragueLangFamilies}{12}
\providecommand{\patSpragueSymFamilies}{9}
\providecommand{\pceilBbhArmContentLimitPlug}{0.775}
\providecommand{\pceilBbhArmContentLimitPlugHi}{0.788}
\providecommand{\pceilBbhArmContentLimitPlugLo}{0.770}
\providecommand{\pceilBbhArmLimitPlug}{0.871}
\providecommand{\pceilBbhWithinOne}{0.917}
\providecommand{\pceilGsmArmContentLimitPlug}{0.783}
\providecommand{\pceilGsmArmContentLimitPlugHi}{0.793}
\providecommand{\pceilGsmArmContentLimitPlugLo}{0.776}
\providecommand{\pceilGsmArmLimitPlug}{0.809}
\providecommand{\pceilGsmDisjMax}{0.796}
\providecommand{\pceilGsmDisjMaxK}{14}
\providecommand{\pceilHqArmLimitPlug}{0.695}
\providecommand{\pceilHqArmLimitPlugContent}{0.733}
\providecommand{\pceilHqWithinOneContent}{0.866}
\providecommand{\pceilMathArmContentLimitPlug}{0.803}
\providecommand{\pceilMathArmContentLimitPlugHi}{0.808}
\providecommand{\pceilMathArmContentLimitPlugLo}{0.798}
\providecommand{\pceilMathArmLimitPlug}{0.834}
\providecommand{\pceilMathWithinOne}{0.959}
\providecommand{\rewordModelsGsm}{13}
\providecommand{\rewordModelsMath}{15}
\providecommand{\rswBbhForcedPrimaryLadders}{8}
\providecommand{\rswBbhForcedPrimaryRise}{8}
\providecommand{\rswBbhForcedPrimarySmallLoses}{3}
\providecommand{\rswBbhReqPrimaryLadders}{8}
\providecommand{\rswBbhReqPrimaryRise}{6}
\providecommand{\rswGsmForcedPrimaryLadders}{8}
\providecommand{\rswGsmForcedPrimaryRise}{3}
\providecommand{\rswGsmReqPrimaryLadders}{8}
\providecommand{\rswGsmReqPrimaryRise}{4}
\providecommand{\rswGsmReqPrimarySmallLoses}{0}
\providecommand{\rswMathForcedPrimaryLadders}{8}
\providecommand{\rswMathForcedPrimaryRise}{4}
\providecommand{\rswMathReqPrimaryLadders}{8}
\providecommand{\rswMathReqPrimaryRise}{7}
\providecommand{\siAppContent}{S16}
\providecommand{\siAppData}{S15}
\providecommand{\siSiCapslope}{S8}
\providecommand{\siSiClosed}{S14}
\providecommand{\siSiFiveshotansweringw}{S10}
\providecommand{\siSiHotpot}{S6}
\providecommand{\siSiHowlonganswerswere}{S2}
\providecommand{\siSiHumanaudit}{S17}
\providecommand{\siSiLarger}{S13}
\providecommand{\siSiReadingaudit}{S3}
\providecommand{\siSiRobustnessofthearm}{S5}
\providecommand{\siSiTemperature}{S9}
\providecommand{\siSiThemodelthatanswer}{S11}
\providecommand{\siSiThesplitpilot}{S18}
\providecommand{\siSiThreshold}{S12}
\providecommand{\siTabClaims}{S1}
\providecommand{\siTabHqstrict}{S6}
\providecommand{\siTabLaduncut}{S20}
\providecommand{\siTabOwnslope}{S9}
\providecommand{\siTabPublished}{S8}
\providecommand{\siTabWording}{S5}
\providecommand{\splitpGsmBestSingleWorst}{2.78}
\providecommand{\splitpGsmSplitWorst}{1.83}
\providecommand{\splitpSeeds}{20}
\providecommand{\tempCheckFdGainDiffCiZero}{4}
\providecommand{\tempCheckFdGainDiffMathPooled}{$-$0.7}
\providecommand{\tempCheckFdGainDiffMathPooledHi}{0.4}
\providecommand{\tempCheckFdGainDiffMathPooledLo}{$-$1.8}
\providecommand{\tempCheckModels}{5}
\providecommand{\tempCheckReqGainDiffCiZero}{5}
\providecommand{\tempCheckReqGainDiffMathPooled}{$-$0.2}
\providecommand{\tempCheckReqGainDiffMathPooledHi}{1.0}
\providecommand{\tempCheckReqGainDiffMathPooledLo}{$-$1.4}
\providecommand{\tempCheckTemperature}{0.7}
\providecommand{\tempCheckTemplates}{4}
\providecommand{\tempCheckWDiffMathPooledPct}{0.0}
\providecommand{\tempCheckWDiffMathPooledPctHi}{0.8}
\providecommand{\tempCheckWDiffMathPooledPctLo}{$-$0.8}
\providecommand{\thiGsmCotCap}{3,072}
\providecommand{\thiGsmCutPct}{11.4}
\providecommand{\thiGsmCutPctMax}{42.2}
\providecommand{\thiGsmM}{8}
\providecommand{\thiGsmMeanCot}{95.5}
\providecommand{\thiGsmMeanForced}{43.1}
\providecommand{\thiGsmMeanNative}{84.6}
\providecommand{\thiGsmMeanNativeCap}{79.9}
\providecommand{\thiGsmMeanReq}{56.6}
\providecommand{\thiMathCutPct}{12.8}
\providecommand{\thiMathCutPctMax}{38.5}
\providecommand{\thiMathMeanCot}{90.2}
\providecommand{\thiMathMeanForced}{39.6}
\providecommand{\thiMathMeanNative}{85.5}
\providecommand{\thiMathMeanNativeCap}{51.7}
\providecommand{\thiMathMeanReq}{53.3}
\providecommand{\thpCapSharePctMinExcluded}{83.4}
\providecommand{\thpGsmCap}{30{,}720}
\providecommand{\thpIncludedN}{8}
\providecommand{\thpModels}{9}
\providecommand{\thpThresholdPct}{50}
\providecommand{\thrDecBbhFiveDiffPct}{12}
\providecommand{\thrDecBbhFiveFdToReq}{0.41}
\providecommand{\thrDecBbhFiveFdToReqHi}{0.48}
\providecommand{\thrDecBbhFiveFdToReqLo}{0.34}
\providecommand{\thrDecBbhFiveReqToFd}{1.57}
\providecommand{\thrDecBbhFiveReqToFdHi}{1.61}
\providecommand{\thrDecBbhFiveReqToFdLo}{1.53}
\providecommand{\thrDecBbhTenDiffPct}{16}
\providecommand{\thrDecBbhTenFdToReq}{0.62}
\providecommand{\thrDecBbhTenFdToReqHi}{0.72}
\providecommand{\thrDecBbhTenFdToReqLo}{0.52}
\providecommand{\thrDecBbhTenReqToFd}{1.47}
\providecommand{\thrDecBbhTenReqToFdHi}{1.52}
\providecommand{\thrDecBbhTenReqToFdLo}{1.42}
\providecommand{\thrDecBbhTwoDiffPct}{15}
\providecommand{\thrDecBbhTwoFdToReq}{0.59}
\providecommand{\thrDecBbhTwoFdToReqHi}{0.66}
\providecommand{\thrDecBbhTwoFdToReqLo}{0.53}
\providecommand{\thrDecBbhTwoReqToFd}{1.91}
\providecommand{\thrDecBbhTwoReqToFdHi}{1.95}
\providecommand{\thrDecBbhTwoReqToFdLo}{1.88}
\providecommand{\thrDecGsmFiveDiffPct}{8}
\providecommand{\thrDecGsmFiveFdToReq}{0.19}
\providecommand{\thrDecGsmFiveFdToReqHi}{0.30}
\providecommand{\thrDecGsmFiveFdToReqLo}{0.07}
\providecommand{\thrDecGsmFiveReqToFd}{1.39}
\providecommand{\thrDecGsmFiveReqToFdHi}{1.43}
\providecommand{\thrDecGsmFiveReqToFdLo}{1.35}
\providecommand{\thrDecGsmTenDiffPct}{10}
\providecommand{\thrDecGsmTenFdToReq}{0.56}
\providecommand{\thrDecGsmTenFdToReqHi}{0.73}
\providecommand{\thrDecGsmTenFdToReqLo}{0.39}
\providecommand{\thrDecGsmTenReqToFd}{2.90}
\providecommand{\thrDecGsmTenReqToFdHi}{2.95}
\providecommand{\thrDecGsmTenReqToFdLo}{2.86}
\providecommand{\thrDecGsmTwoDiffPct}{8}
\providecommand{\thrDecGsmTwoFdToReq}{0.03}
\providecommand{\thrDecGsmTwoFdToReqHi}{0.13}
\providecommand{\thrDecGsmTwoFdToReqLo}{$-$0.06}
\providecommand{\thrDecGsmTwoReqToFd}{1.24}
\providecommand{\thrDecGsmTwoReqToFdHi}{1.28}
\providecommand{\thrDecGsmTwoReqToFdLo}{1.20}
\providecommand{\thrDecMathFiveDiffPct}{7}
\providecommand{\thrDecMathFiveFdToReq}{0.11}
\providecommand{\thrDecMathFiveFdToReqHi}{0.14}
\providecommand{\thrDecMathFiveFdToReqLo}{0.09}
\providecommand{\thrDecMathFiveReqToFd}{1.44}
\providecommand{\thrDecMathFiveReqToFdHi}{1.47}
\providecommand{\thrDecMathFiveReqToFdLo}{1.42}
\providecommand{\thrDecMathTenDiffPct}{6}
\providecommand{\thrDecMathTenFdToReq}{0.22}
\providecommand{\thrDecMathTenFdToReqHi}{0.25}
\providecommand{\thrDecMathTenFdToReqLo}{0.20}
\providecommand{\thrDecMathTenReqToFd}{2.63}
\providecommand{\thrDecMathTenReqToFdHi}{2.67}
\providecommand{\thrDecMathTenReqToFdLo}{2.59}
\providecommand{\thrDecMathTwoDiffPct}{3}
\providecommand{\thrDecMathTwoFdToReq}{0.01}
\providecommand{\thrDecMathTwoFdToReqHi}{0.03}
\providecommand{\thrDecMathTwoFdToReqLo}{$-$0.01}
\providecommand{\thrDecMathTwoReqToFd}{0.09}
\providecommand{\thrDecMathTwoReqToFdHi}{0.11}
\providecommand{\thrDecMathTwoReqToFdLo}{0.06}
\providecommand{\wckBbhConPairs}{378}
\providecommand{\wckBbhConRev}{81}
\providecommand{\wckBbhConRevContent}{78}
\providecommand{\wckBbhConRevContentPct}{21}
\providecommand{\wckBbhConRevPct}{21}
\providecommand{\wckBbhFdToReqPanel}{0.96}
\providecommand{\wckBbhLongFirstPct}{28.5}
\providecommand{\wckBbhLongPct}{31.6}
\providecommand{\wckBbhRevCorrFirst}{71}
\providecommand{\wckBbhVarLongPctFirst}{50}
\providecommand{\wckGsmFdToReqPanel}{3.04}
\providecommand{\wckGsmLongFirstPct}{39.3}
\providecommand{\wckGsmLongPct}{43.5}
\providecommand{\wckGsmRevCorrFirst}{36}
\providecommand{\wckGsmVarLongPctFirst}{92}
\providecommand{\wckMathFdToReqPanel}{4.51}
\providecommand{\wckMathLongFirstPct}{40.6}
\providecommand{\wckMathLongPct}{42.4}
\providecommand{\wckMathRevCorrFirst}{24}
\providecommand{\wckMathVarLongPctFirst}{83}
\providecommand{\wckNoQBbhFdToReq}{3.07}
\providecommand{\wckNoQBbhRev}{61}
\providecommand{\wckNoQBbhRevPairs}{351}
\providecommand{\wckNoQGsmFdToReq}{8.92}
\providecommand{\wckNoQGsmRev}{67}
\providecommand{\wckNoQGsmRevPairs}{378}
\providecommand{\wckNoQMathFdToReq}{9.58}
\providecommand{\wckNoQMathRev}{69}
\providecommand{\wckNoQMathRevPairs}{378}
\providecommand{\wckQwenSixMoeMathPooledFdContent}{50.7}
\providecommand{\wckQwenSixMoeMathPooledReqContent}{36.2}
\providecommand{\wckQwenSixMoeWGsm}{0.01}
\providecommand{\wckQwenSixMoeWMath}{0.04}
\title{Spontaneous reasoning reorders language models’ gains from chain-of-thought}
\author{\parbox{\textwidth}{\centering Dean Ariel\textsuperscript{1,2}, Eyal Hadad\textsuperscript{1}, Harel Navon\textsuperscript{1}, Tomer Hamuel\textsuperscript{1}, Ran Balicer\textsuperscript{3,4,5}, Seffi Cohen\textsuperscript{1}\\[6pt]
\small\textsuperscript{1}The Stein Faculty of Computer and Information Science, Ben-Gurion University of the Negev, Beer-Sheva, Israel\\
\small\textsuperscript{2}Department of Family Medicine, Clalit Health Services, Israel\\
\small\textsuperscript{3}Clalit Research Institute, Innovation Division, Clalit Health Services, Ramat-Gan, Israel\\
\small\textsuperscript{4}The Ivan and Francesca Berkowitz Family Living Laboratory Collaboration at Harvard Medical School and Clalit Research Institute, Ramat-Gan, Israel\\
\small\textsuperscript{5}School of Public Health, Faculty of Health Sciences, Ben-Gurion University of the Negev, Beer-Sheva, Israel\\[4pt]
\small Correspondence: Dean Ariel, deana@post.bgu.ac.il}}
\date{}
\begin{document}
\maketitle
\begin{abstract}
Chain-of-thought prompting asks a language model to reason in writing before answering. Its benefit is measured against a direct answer, for which the model is asked for the answer alone. In \claimsUnconstrainedN{} of \claimsDirectN{} published comparisons, nothing stopped the model from writing out its reasoning anyway. Many models do: told to show no working, \awxModels{} open-weight models answered at length a third of the time. We compared this requested answer with a forced one, which we began for the model and cut short after a few words. The most accurate of \cwModels{} models, a commercial model with reasoning off, gained \cwFullReqGainMath{} points from chain-of-thought against its requested answers and \cwFullFdGainMath{} against its forced ones. Measuring gains against the forced answer instead of the requested one reversed which of two models gained more in about one pair in five, shrank the measured advantage of stronger models and reversed the measured effect of five-shot prompting in arithmetic. Flagship models no longer let reasoning be switched off, so their measured gains depend on how the direct answer is built.
\end{abstract}

\section*{Introduction}
Language models answer reasoning questions more accurately when they are prompted to work through the problem in writing first, a technique called chain-of-thought prompting \citep{wei2022chain,kojima2022large}. Written reasoning costs time and computing \citep{sprague2025cot,meincke2025prompting2}, so whether it pays, and for which models, decides how models are run and is a recurring research question. The answers conflict. Chain-of-thought was reported to help large models and hurt small ones \citep{wei2022chain}, and later to help most the models that do not reason unprompted, often older or smaller ones \citep{meincke2025prompting2}.

Every claim about who benefits rests on a difference: a model's accuracy with chain-of-thought minus its accuracy when it answers directly, the baseline. Of \claimsDirectN{} published comparisons with such a baseline, only \claimsConstrainedN{} constrained the direct answer; the others asked for the final answer, showed examples with answers alone or posed the question with no instruction, and let the model write as it chose \citep{sprague2025cot,meincke2025prompting2} (Supplementary Table~\siTabClaims). We call a baseline obtained this way the requested baseline. Many models reason in writing when nobody asks them to \citep{meincke2025prompting2}, and many keep doing so when asked for the answer alone, even with reasoning switched off \citep{lei2026thinking}. When a model does (Fig.~1a), its baseline already contains part of the effect being measured, much as an A/B test understates an effect when the control group also gets the change. A direct answer can instead be forced: the evaluator writes the opening of the answer into the model's reply and stops the model a few words later, leaving no room for reasoning. We call this the forced baseline. Studies of whether models rely on their written reasoning use a similar device \citep{lanham2023measuring}. We found no study that measured whether the choice between the two baselines changes which models appear to benefit.

\begin{figure}[tbp]\centering\includegraphics[width=\textwidth]{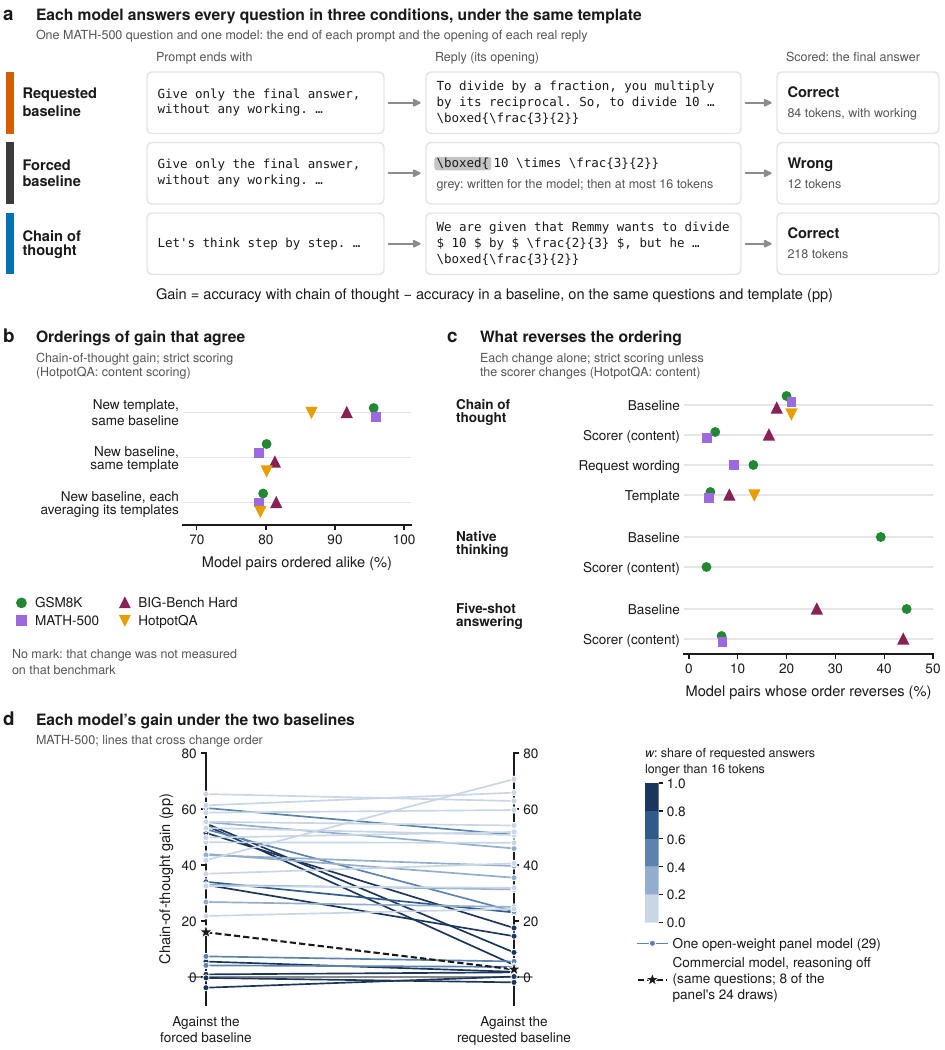}
\caption{\textbf{Changing only the baseline flipped about one ordering of gains in five.} \textbf{a}, The chain-of-thought condition and the two baselines. Chain-of-thought gain is chain-of-thought accuracy minus baseline accuracy on the same template. The requested baseline asks for the answer alone; the forced baseline writes the opening of the box for the model and allows at most 16 more tokens. Replies are the opening of real outputs of one model (Qwen3-30B-A3B-Instruct-2507) on one MATH-500 prealgebra question under one template; the requested reply writes its working although asked not to. Grey highlight, text written for the model. \textbf{b}, Share of model pairs ordered alike by chain-of-thought gain, one row each: two templates under the same baseline, the two baselines on the same template, and the two baselines when each averages its own 14 to 18 distinct templates. \textbf{c}, Share of model pairs whose order of gain flips when only the baseline changes, only the scorer changes, only the wording of the request for the final answer changes (GSM8K and MATH-500) or only the template changes (chain-of-thought against the requested baseline), one row each, for prompted chain-of-thought (\gtM{} models), native thinking (\genThiGsmArmM{} checkpoints, GSM8K) and five-shot answering (baseline change: gain measured within each baseline). Colours and symbols mark benchmarks; a benchmark without a symbol in a row was not measured for that change. \textbf{d}, Chain-of-thought gain of each of the \gtM{} open-weight models on MATH-500 against the forced baseline (left) and the requested baseline (right), one line per model, coloured by $w$, the share of its requested answers longer than 16 tokens; lines that cross mark the \aeMathReversed{} pairs whose order the baseline switch reverses. Dashed line with stars, the commercial model with its reasoning mode off, on the same questions and 8 of the 24 templates. Strict scoring unless stated, gains in points.}\label{fig:1}
\end{figure}

The question is becoming harder to avoid. The current flagship models of three leading developers reason by default and no longer accept a request with reasoning switched off \citep{openai2026reasoning,anthropic2026thinking,google2026thinking}; for them, a direct answer exists only if the evaluator asks for it or starts it. We therefore measured how often models reason when asked for the final answer, how this changes which models appear to gain most, and what choosing models from such gains costs. A panel of \gtM{} open-weight models answered every question of three reasoning benchmarks with chain-of-thought and under both baselines, a fourth benchmark was held back to test the findings, and six larger models and a commercial flagship model repeated the experiment. We find that this spontaneous reasoning varies widely between models and reorders them by their gain from chain-of-thought, and that a count of answer lengths corrects much of the distortion.

\section*{Results}

\subsection*{Study design}

Each model answered every question in three conditions (Fig.~1a). With chain-of-thought, it was asked to reason step by step before answering. In the requested baseline, it was asked to give only its final answer, without any working. In the forced baseline, the opening of the answer was prefilled in its reply and the model could add at most 16 tokens (word pieces, together a few words) before it was stopped. A model's gain is its accuracy with chain-of-thought minus its accuracy in a baseline, in percentage points (points).

The panel held \gtM{} open-weight models of up to 35 billion parameters from five families (Gemma, Llama, Mistral, Phi and Qwen), \bhM{} of them on BIG-Bench Hard after the exclusion rules (Methods). The questions came from three benchmarks: GSM8K, grade-school mathematics \citep{cobbe2021training}; MATH-500, competition mathematics \citep{hendrycks2021math,lightman2024lets}; and BIG-Bench Hard, tasks of logic and language (25 of its 27 subtasks, all but two with free-form answers) \citep{suzgun2023challenging}. A fourth benchmark, HotpotQA, asks questions that combine facts from several Wikipedia passages \citep{yang2018hotpotqa}. HotpotQA was run only after all other analyses were complete, and because some of its answers are long names, its forced baseline allowed 64 tokens.

Small changes in a prompt's format or wording change accuracy \citep{sclar2024quantifying,mizrahi2024state}, so every condition ran under \gtKdisc{} prompt templates drawn at random, and each model's gain was averaged over them. A further \gtKval{} templates, the held-out templates, were kept apart to test the choice of models described below. Strict scoring counted an answer as correct only in the required format. Content scoring also accepted a correct answer in the wrong format; human raters confirmed \hrtDisagreeContentAgreePct\% of the answers that content scoring credited and strict scoring rejected. Strict scoring is used unless stated, and content scoring throughout on HotpotQA (Methods).

\subsection*{Models reasoned when asked for the answer alone}

We call a requested answer long when it runs past 16 tokens, the forced baseline's limit, and call a model's share of long requested answers its $w$. Averaged over models, $w$ was \bmixGsmWMeanPct\% on GSM8K, \bmixMathWMeanPct\% on MATH-500 and \bmixBbhWMeanPct\% on BIG-Bench Hard, and models differed widely: on GSM8K, $w$ ran from \bmixGsmWMinPct\% to \bmixGsmWMaxPct\%. Because $w$ is counted from answer lengths, it needs no answer key.

Long answers often contained reasoning. A language model used as a judge read \labN{} requested answers, sampled at random within each benchmark, model and length band, under a written rubric, blind to the reference answer and to the model that wrote each one, and it agreed with two human raters as well as they agreed with each other (Methods). At most \labRShortGsmPct\% of short answers on any benchmark contained a step toward the answer that the question does not state, such as a calculation or a deduction. Among long answers, \labRLongBbhPct\% on BIG-Bench Hard, \labRLongGsmPct\% on GSM8K and \labRLongMathPct\% on MATH-500 did; most of the others added only text, such as a restatement of the question.

The most accurate model reasoned too. Qwen4-Max, a commercial flagship model run with its reasoning mode switched off (Methods), was the most accurate of all \cwModels{} models (the panel, six larger models and itself) under every condition on GSM8K and MATH-500, on the \cwItems{} questions per benchmark that it shared with them. On those questions, asked for the final answer, it gave long answers to \cwWGsmPct\% of GSM8K and \cwWMathPct\% of MATH-500 questions, and all \labQwenNLong{} of its long answers in the judge's sample contained a reasoning step.

\subsection*{Reasoning in the baseline shrank the measured gain}

A model that reasons in its requested baseline leaves less for chain-of-thought to add. Long requested answers scored much like chain-of-thought answers (Fig.~2a). On MATH-500 they were correct \bmixMathLongReq\% of the time, against \bmixMathLongFd\% for forced answers and \bmixMathLongCot\% for chain-of-thought answers to the same questions. Short requested answers scored close to forced ones (\bmixMathShortReq\% and \bmixMathShortFd\%). Models also reasoned more on harder questions. We scored a question's difficulty for a model by how often the other models' forced answers to it were wrong. Within each model, every 10 points of difficulty multiplied the odds of a long requested answer by \diffOrGsm{} on GSM8K (95\% interval \diffOrGsmLo{} to \diffOrGsmHi) and by \diffOrMath{} on MATH-500 (\diffOrMathLo{} to \diffOrMathHi), and the effect was positive in \diffPosGsm{} of \diffModelsGsm{} and \diffPosMath{} of \diffModelsMath{} models.

\begin{figure}[tbp]\centering\includegraphics[width=\textwidth]{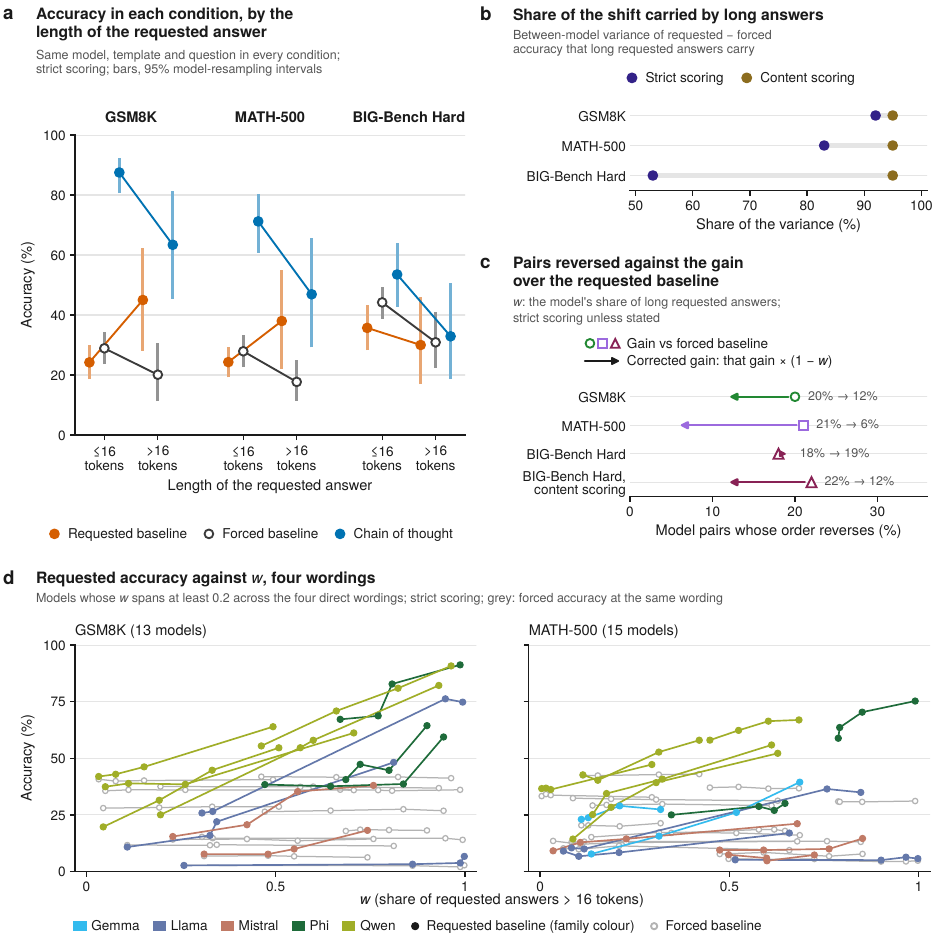}
\caption{\textbf{On mathematics, long requested answers accounted for most of the disagreement, and rewording the request moved accuracy with them.} \textbf{a}, Accuracy with the forced baseline, the requested baseline and chain-of-thought on the same model, template and question, for questions whose requested answer was at most 16 tokens long (left of each pair) or longer (right); the forced baseline is drawn open. Bars, 95\% model-resampling intervals. \textbf{b}, Share of the between-model variance of requested-minus-forced accuracy contributed by long requested answers, under strict and content scoring (two colours, joined by a grey bar). \textbf{c}, Share of model pairs whose order by the gain over the requested baseline is flipped by the gain over the forced baseline (open symbols) and by the corrected gain, that gain multiplied by $1-w$, where $w$ is the model's share of long requested answers (arrow heads). BIG-Bench Hard is shown under strict and, below it, content scoring. HotpotQA is reported in the text. \textbf{d}, Requested-baseline accuracy against $w$ across the four wordings of the request for the final answer, one line per model, for the models whose $w$ spanned at least 0.2 across wordings (\rewordModelsGsm{} on GSM8K, \rewordModelsMath{} on MATH-500); grey, forced-baseline accuracy under the same wording. Colours mark model families; strict scoring.}\label{fig:2}
\end{figure}

The difference between a model's two gains equals its requested minus its forced accuracy. We split this difference into a part from long and a part from short requested answers. Across models, the long part accounted for \bmixDecGsmVarLongPct\% of the variance of this difference on GSM8K and \bmixDecMathVarLongPct\% on MATH-500 (Fig.~2b), and for \bmixDecBbhVarLongPct\% on BIG-Bench Hard, \bmixDecBbhVarLongPctContent\% with content scoring. Reasoning, more than length, carried the effect: the share of a model's requested answers that the judge found to contain a reasoning step explained more of the variation of the difference across models than $w$ did ($R^2$ \labRsqRGsm{} against \labRsqWGsm{} on GSM8K; Supplementary Section~\siSiReadingaudit).

Rewording one sentence showed the effect within single models. We wrote four wordings of the sentence that asks for the final answer and reran GSM8K and MATH-500 with each, keeping the rest of the prompt. The wordings changed how often a model reasoned, and its requested accuracy moved with it. In every one of the \rewordModelsGsm{} GSM8K and \rewordModelsMath{} MATH-500 models whose $w$ spanned at least 0.2 across the wordings, requested accuracy rose with $w$, while forced accuracy stayed flat (Fig.~2d). The requested baseline was therefore fragile: rewording that one sentence flipped which of two models gained more in \pctof{\awGsmClauseSwitchReqReversed}\% of GSM8K pairs, against \pctof{\awGsmClauseSwitchFdReversed}\% when the same wordings were compared under the forced baseline (Supplementary Table~\siTabWording). We also tried two more requests, one with no instruction about working and one that forbade it outright. With no instruction about working, $w$ was \awxGsmDefW{} on GSM8K and the baseline switch reversed \awxGsmDefSwitchPct\% of pairs; told ``\awxFirmText'', models still answered at length a third of the time ($w$ \awxGsmFirmW), and the switch reversed \awxGsmFirmSwitchPct\%. Forced accuracy stayed between \awxGsmDefFd\% and \awxGsmCurFd\% under all six requests.

If long requested answers score like chain-of-thought answers and short ones like forced answers, a model keeps only the share $1-w$ of its forced-baseline gain when measured against the requested baseline. Multiplying the forced-baseline gain by $1-w$ therefore estimates the requested-baseline gain; we call this product the corrected gain. The flagship model showed how large the effect can be. Run on all \cwFullItems{} numeric MATH-500 questions under the first \cwFullTemplates{} of its templates, it gave long requested answers to \cwFullWMathPct\% of them. Its gain from chain-of-thought was \cwFullReqGainMath{} points against the requested baseline (95\% interval from resampling questions, \cwFullReqGainMathLo{} to \cwFullReqGainMathHi) and \cwFullFdGainMath{} against the forced baseline (\cwFullFdGainMathLo{} to \cwFullFdGainMathHi). Its corrected gain, \cwFullCorrMath{} points, was close to the requested-baseline gain, so its long answers accounted for most of the difference. On the questions it shared with the panel, the gap held on BIG-Bench Hard, where it gave long answers to \cwWBbhPct\% of questions and gained \cwReqGainBbh{} points against the requested baseline and \cwFdGainBbh{} against the forced baseline (difference \cwGainDiffBbh{} points, \cwGainDiffBbhLo{} to \cwGainDiffBbhHi), and was smaller on GSM8K (\cwReqGainGsm{} against \cwFdGainGsm{} points; Supplementary Section~\siSiClosed).

\subsection*{The baselines disagreed on which models gained most}

Because models reasoned spontaneously to different degrees, the two baselines ranked them differently. For every pair of models, we asked which gained more, then asked again under the other baseline. The answer flipped for \aeGsmReversedPct\% of pairs on GSM8K, \aeMathReversedPct\% on MATH-500 and \aeBbhReversedPct\% on BIG-Bench Hard (Fig.~1c,d). On MATH-500, \aeMathGapFivePairs{} pairs had gains that differed by more than 5 points under both baselines, and \aeMathGapFiveReversed{} of these flipped. Changing the template alone flipped far fewer pairs: two random templates under one baseline ordered \pctof{\pceilBbhWithinOne}--\pctof{\pceilMathWithinOne}\% of pairs alike (Fig.~1b), and averaging each baseline over its \pceilGsmDisjMaxK{} distinct templates (the 24 random draws repeated some) still left the two baselines agreeing on only \pctof{\pceilGsmDisjMax}\% of GSM8K pairs. Each model's gain was highly repeatable across templates: its reliability, the share of the variation of gains across models that is not template noise (Methods), was \gtRben{} on GSM8K, \mtRben{} on MATH-500 and \bhRben{} on BIG-Bench Hard. The flips came from the long answers: recomputing each model's baseline difference from its long requested answers alone reproduced \attrGsmLongRevShared{} of \attrGsmObsRev{} flips on GSM8K and \attrMathLongRevShared{} of \attrMathObsRev{} on MATH-500, and its short answers alone, \attrGsmShortRevShared{} and \attrMathShortRevShared; under content scoring, the long answers reproduced \attrContentShareMin--\attrContentShareMax\% of the flips on all three benchmarks (Supplementary Section~\siSiHowlonganswerswere).

The flips were robust. They occurred between models of the same family and among the \aeRecentM{} checkpoints released in 2025 or later, and they remained when whole families were resampled. They held under content scoring and under each of the four wordings of the request, and on BIG-Bench Hard they held when forced answers were restricted to the item's options so that every answer was parseable (Supplementary Section~\siSiRobustnessofthearm). Sampling at temperature 0.7 in place of greedy decoding (always taking the most likely next token) changed $w$ and both gains of five models on MATH-500 by less than a point (Supplementary Section~\siSiTemperature).

The corrected gain removed most of the flips on MATH-500 and fewer on GSM8K. Taking the ranking by requested-baseline gain as the target, the corrected gain misordered \bmixMathRevCorr{} of the \bmixMathPairs{} MATH-500 pairs and the forced-baseline gain \bmixMathRevObs{}; on GSM8K, \bmixGsmRevCorr{} against \bmixGsmRevObs{} (Fig.~2c). On BIG-Bench Hard it helped only with content scoring (Supplementary Section~\siSiHowlonganswerswere).

On HotpotQA, the benchmark held back to test the findings, the flips remained but the correction failed. There, chain-of-thought added little (median gain \hqGapCotFdMedianContent{} points), so small differences between the baselines were enough to reorder models: the baselines flipped \aeHqReversedPctContent\% of pairs, while two templates under one baseline ordered \pctof{\pceilHqWithinOneContent}\% alike. Length did not mark reasoning on this benchmark, whose answers are often long names, and correcting for $w$ did not help (Supplementary Section~\siSiHotpot).

\subsection*{The baseline changed published conclusions}

The claim that chain-of-thought helps larger models more \citep{kojima2022large} was tested in series of models released by one developer in several sizes. On MATH-500 it held in \rswMathReqPrimaryRise{} of \rswMathReqPrimaryLadders{} series under the requested baseline and in \rswMathForcedPrimaryRise{} under the forced baseline; on BIG-Bench Hard, in \rswBbhReqPrimaryRise{} and \rswBbhForcedPrimaryRise{} of \rswBbhReqPrimaryLadders{}, respectively (Supplementary Table~\siTabPublished).

The baseline also changed how gain related to capability, the question behind the published claims about large and small models. We regressed each gain across models on the accuracy of its own baseline, with gain and accuracy measured on different templates, so that chance errors shared by the two could not create a slope. Gain rose far less steeply with capability under the requested baseline: on MATH-500 by \crvMathOwnReq{} points of gain per point of accuracy, against \crvMathOwnFd{} under the forced baseline, and on BIG-Bench Hard by \crvBbhOwnReq{} against \crvBbhOwnFd. On GSM8K the two slopes had opposite signs, \crvGsmOwnReq{} against \crvGsmOwnFd. The difference excluded zero on all three benchmarks (on MATH-500, \crvMathOwnDiff{}; 95\% interval \crvMathOwnDiffLo{} to \crvMathOwnDiffHi), and on both mathematics benchmarks also when whole families were resampled (Supplementary Table~\siTabOwnslope). The corrected gain accounted for \crvOwnShareStrictMin--\crvOwnShareStrictMax\% of this difference. On an external capability index built from many benchmarks \citep{epoch2025eci}, gains rose under both baselines, again less steeply under the requested one (Supplementary Section~\siSiCapslope). The requested baseline thus shrinks the measured advantage of stronger models.

Beyond chain-of-thought, the baseline also reversed the measured effect of five-shot prompting, which shows the model five questions with their answers alone before the real one. On GSM8K, five-shot prompting changed accuracy by \fsrGsmGainReq{} points on average when measured within the requested baseline and by \fsrGsmGainFd{} when measured within the forced baseline, and the two measurements ordered \fsrGsmRev{} of \fsrGsmRevPairs{} pairs differently. The solved examples nearly stopped spontaneous reasoning, cutting $w$ from \fsWGsmWZsPct\% to \fsWGsmWFsPct\%, and the models whose $w$ fell most were the ones whose five-shot effect changed sign between the baselines (Spearman's $\rho$ \fsWGsmRho; 95\% interval \fsWGsmRhoLo{} to \fsWGsmRhoHi; Supplementary Section~\siSiFiveshotansweringw). Some models have a built-in reasoning mode that can be switched on or off \citep{qwen2025qwen3}. Among \genThiGsmArmM{} such models, \genThiGsmArmQwen{} of them from Qwen, changing the baseline flipped which of two models gained more from switching the mode on in \genThiGsmArmRev{} of \genThiGsmArmRevPairs{} pairs on GSM8K (Fig.~1c).

\subsection*{Choosing models under the wrong baseline was costly}

Gains are used to choose models. Consider a team that can afford chain-of-thought for a third of its models and gives it to the third that gains most. It ranks the models by gains from a small trial, a pilot, or by published gains, whose baseline need not match the way the team will run its models. We measured the loss, the gain the chosen third gives up against the best possible third, on the held-out templates, which played no part in the choice. For a team that deploys its models with a request for the final answer, choosing by forced-baseline gains lost \ddirGsmFdToReq{} points per chosen model more than choosing by requested-baseline gains (the added loss) on GSM8K, \ddirMathFdToReq{} on MATH-500 and \ddirBbhFdToReq{} on BIG-Bench Hard (Fig.~3a); for scale, the best possible third gained \decGsmReqBestThird{}, \decMathReqBestThird{} and \decBbhReqBestThird{} points. For a team that deploys with the forced baseline, choosing by requested-baseline gains cost \ddirBbhReqToFd{} to \ddirMathReqToFd{} points more. This reverse mismatch cost more than the first only on HotpotQA, at \hqDdirReqToFdContentPostHoc{} against \hqDdirFdToReqContentPostHoc{} points. One model's answers changed unusually when we began its reply for it; leaving it out, the added losses were \wckNoQGsmFdToReq{}, \wckNoQMathFdToReq{} and \wckNoQBbhFdToReq{} points (Supplementary Section~\siSiThemodelthatanswer).

\begin{figure}[tbp]\centering\includegraphics[width=0.82\textwidth]{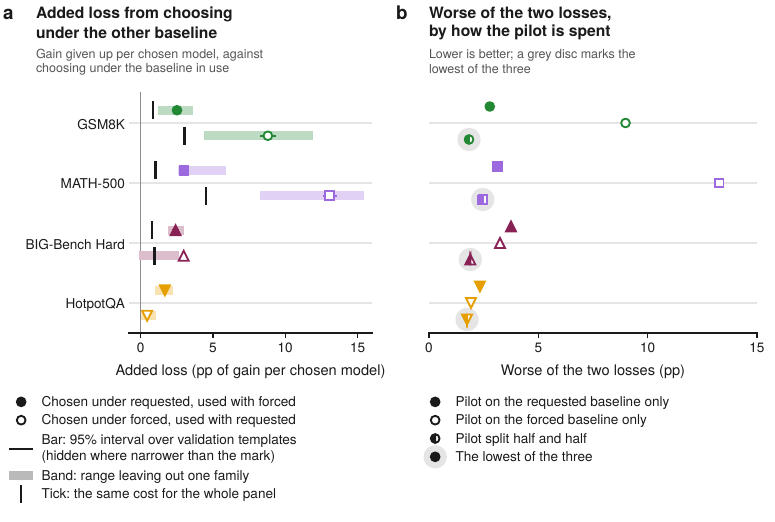}
\caption{\textbf{Choosing under the wrong baseline cost gain, and a split pilot limited the loss.} \textbf{a}, Added loss, the gain given up compared with the best possible choice, when models are chosen by their gain on one template under one baseline and the choice is valued under the other, against models chosen under the baseline in use, in points of gain per chosen model: chosen under the forced baseline and valued under the requested one (open symbols) and the opposite (filled symbols). Thin bars, 95\% $t$ intervals over held-out templates; light bands, range when one family is left out; black ticks, the same cost per model of the whole panel. \textbf{b}, The worse of the two deployment losses for a 100-question pilot spent wholly on the requested baseline (filled), wholly on the forced baseline (open), or half on each (half-filled; a grey disc marks the lowest of the three), averaged over 200 pilots and valued on held-out questions and held-out templates. One third of the panel receives chain-of-thought.}\label{fig:3}
\end{figure}

Two steps limited this loss. First, a team holding forced-baseline gains can multiply each by its model's $1-w$, which needs no answer key and no judge model; on mathematics, $w$ counted on 50 questions correlated at \awMechGsmScreenCollCorr{} with $w$ from the full run. Choosing by these corrected gains cut the loss of a forced-baseline choice from \bmixDecMathChoiceValFd{} to \bmixDecMathChoiceValCorr{} points per chosen model on MATH-500 and from \bmixDecGsmChoiceValFd{} to \bmixDecGsmChoiceValCorr{} on GSM8K. These losses are counted against the best possible third, not against a requested-baseline choice as above. The correction did not help on HotpotQA. Second, a team that does not know which baseline it will deploy can split its pilot, half under each baseline, with no extra model calls. We scored each pilot by its loss under whichever baseline turned out worse for it, and the split pilot beat the better single-baseline pilot on all four benchmarks, for example \splitpGsmSplitWorst{} against \splitpGsmBestSingleWorst{} points on GSM8K (Fig.~3b).

\subsection*{The baselines also disagreed among larger models}

Six larger models repeated the whole experiment (Methods; Supplementary Section~\siSiLarger). Five extend families already in the panel: Llama, to 70 billion parameters; Qwen, to \ladSizeQwenB{} billion; and Mistral, with three models up to \ladSizeMaxB{} billion. The sixth, from DeepSeek, has no smaller relative in the panel. Their gains fell inside the panel's range in \ladInRangeCells{} of \ladCells{} benchmark cases (Fig.~4a,b). Among the larger models, how often they reasoned did not follow size: on GSM8K, $w$ was \pctof{\ladWLlamaSeventyGsm}\% for the 70-billion-parameter Llama model, against \pctof{\ladWLlamaEightGsm}\% for its 8-billion-parameter relative, but \pctof{\ladWMistralLargeGsm}\% for the largest Mistral model. With the larger models added, the baselines flipped \combRevPctMath--\combRevPctGsm\% of pairs under content scoring (Fig.~4c). Among the six larger models themselves, they flipped \ladRevPooled{} of \ladPairsPooled{} pairs across the three main benchmarks, and among the four with 70 billion parameters or more, \ladBigRevPooled{} of \ladBigPairsPooled.

\begin{figure}[tbp]\centering\includegraphics[width=\textwidth]{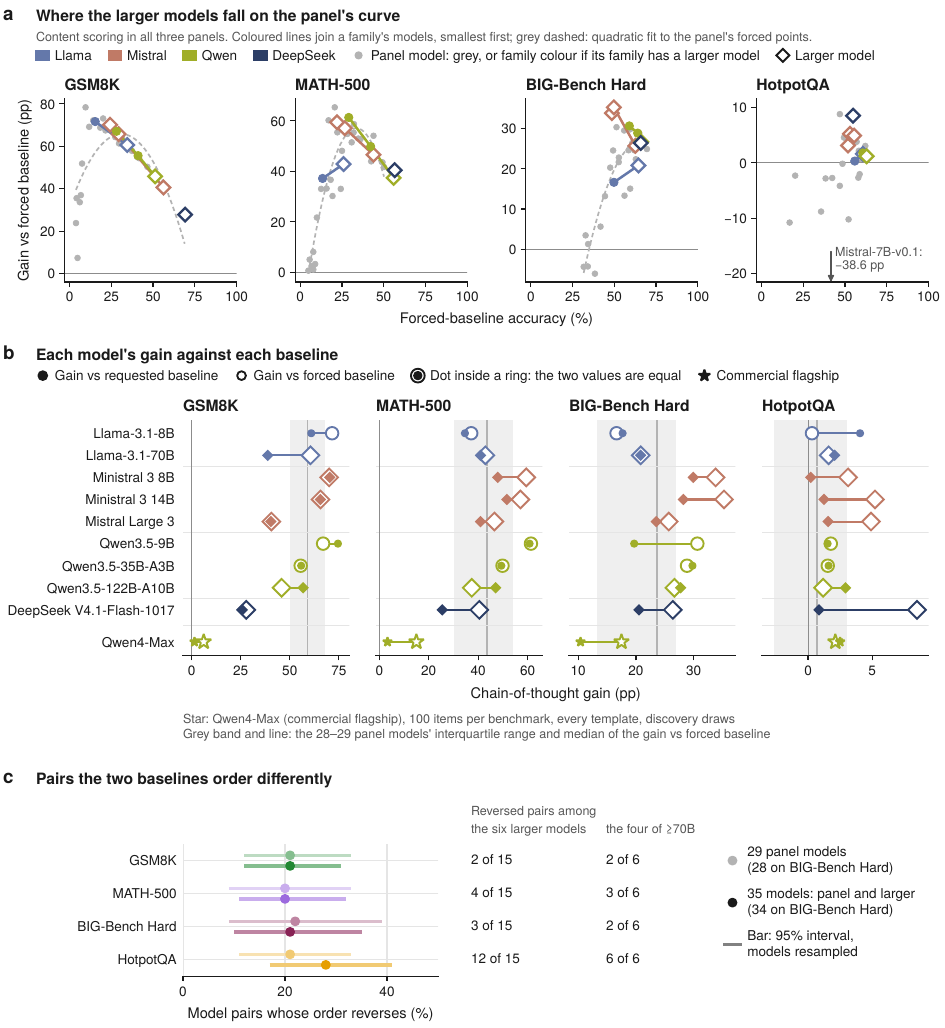}
\caption{\textbf{Larger models fell within the panel's range, and the two baselines flipped their pairs too.} \textbf{a}, Chain-of-thought gain over the forced baseline against forced-baseline accuracy. Grey circles, panel models of other families; coloured symbols, the Llama, Mistral and Qwen families (circles, panel models; diamonds, larger models), joined from the smallest model of each family; olive diamond, the DeepSeek model, which has no smaller relative in the panel. Dashed curve, quadratic fit to the panel. \textbf{b}, Each model's chain-of-thought gain against the requested baseline (filled) and the forced baseline (open); a dot inside a ring marks equal values. The last row (stars) is the commercial flagship model, on the \cwItems{} questions per benchmark it shared with the other models. Grey band and line, interquartile range and median of the panel's gains over the forced baseline. \textbf{c}, Share of model pairs whose order of gain flips between the two baselines, among the panel (grey) and among the panel with the six larger models (black). Bars, 95\% intervals with models resampled. Counts, flipped pairs among the six larger models and among the four with 70 billion parameters or more. Content scoring; gains in points.}\label{fig:4}
\end{figure}
\section*{Discussion}

Language models often reason when asked for the final answer alone, and that reasoning enters the baseline against which their gain from chain-of-thought is measured. Across \gtM{} open-weight models, six larger models and a commercial flagship model, it shrank measured gains unevenly, so the choice of baseline alone flipped which of two models gained more for about one pair in five and changed published conclusions.

These results offer a measurement explanation for conflicting claims about who benefits from reasoning. Apparent jumps in ability can come from how a result is measured \citep{schaeffer2023are}, and so can apparent differences in who gains from chain-of-thought. The report that chain-of-thought helps most the models that do not reason unprompted \citep{meincke2025prompting2} rests on requested baselines. In our panel, that baseline shrank the measured gain of the models that reason unprompted and flattened the rise of gain with capability on every benchmark. The distortion also reaches the most capable models: the most accurate model we tested reasoned in four of five requested mathematics answers. For flagship models that no longer allow reasoning to be switched off, the only direct answer an evaluator can obtain without starting the reply is one that is asked for, and models with a switchable reasoning mode often keep reasoning when it is off \citep{wang2025hybrid}. Some recent models, called decision models, never write reasoning at all: they score a fixed set of options in a single pass and return no text \citep{ariel2026decision,google2026decisions}.

Neither baseline is correct for every purpose. A team that will deploy models with a request for the final answer should measure gains under that request. The forced baseline removes written reasoning, but starting the reply \citep{lei2026thinking} and cutting it short can also change how some models answer (Supplementary Section~\siSiThemodelthatanswer), so neither baseline isolates the effect of reasoning alone. The requested baseline also depends on the exact wording of the request, because the wording changes how often models reason; the forced baseline barely does.

A report of who benefits from chain-of-thought should say how the direct answer was obtained and give each model's $w$, which needs only the lengths of its requested answers. When the deployment baseline is uncertain, a pilot split between the two baselines protects against the larger loss, at a small price when the baseline is known. We release code that counts $w$, measures gains under both baselines and recomputes a claim of who benefits.

Our findings have limits. First, the panel covers open-weight models of up to 35 billion parameters from five families, on four benchmarks in English, with one answer format, a final answer in a box, and greedy decoding. The larger models add one model per size, so they show where each size falls and not why. Second, length marks reasoning loosely: many long answers added text without a reasoning step, so $w$ is an upper bound on how often a model reasons. Third, where chain-of-thought adds little, as on HotpotQA, the baselines still reorder models but length no longer marks reasoning, and correcting for $w$ did not help; the split pilot helped on all four benchmarks. Fourth, the rewording test covered mathematics only. Last, content scoring uses rule-based answer extraction, checked by human raters on a sample.

To conclude, spontaneous reasoning in the requested baseline reorders language models' gains from chain-of-thought, and choices made from those gains can be wrong. As flagship models stop accepting requests with reasoning switched off, the direct answer has become something the evaluator builds; stating how it was built, and how often it came back long, makes claims about who benefits from reasoning comparable.
\section*{Methods}

\subsection*{Models and panels}
The panel holds every eligible open instruction-tuned checkpoint of five families (Gemma, Llama, Mistral, Phi and Qwen), \dtEligible{} in all; written exclusion rules, checked on items from each benchmark's training split, left \gtM{} models on GSM8K and MATH-500 and \bhM{} on BIG-Bench Hard. All models ran with greedy decoding and their own chat formats, with any switchable reasoning mode turned off; model identities, revisions and decoding settings are listed in Supplementary Section~\siAppData.

\subsection*{Larger and commercial models}
Six larger models ran every template, condition and question of the four benchmarks (\ladRequests{} requests per model). Five extend three of the panel's families: two Mistral models of \ladSizeMinistralEightB{} and \ladSizeMinistralFourteenB{} billion parameters and the largest model of their release, a mixture-of-experts model of \ladSizeMaxB{} billion total and \ladSizeMistralLargeActiveB{} billion active parameters \citep{mistral2025mistral3,mistral2025large3}; a mixture-of-experts Qwen model of \ladSizeQwenB{} billion total and \ladSizeQwenActiveB{} billion active parameters, above the panel's two models of its release; and a 70-billion-parameter Llama model, above the panel's 8-billion-parameter one. The sixth, a mixture-of-experts DeepSeek model of \ladSizeDeepSeekTotalB{} billion total parameters, with \ladSizeDeepSeekActiveInB{} billion active per input token and \ladSizeDeepSeekActiveOutB{} billion per output token \citep{deepseek2026v41}, has no smaller relative in the panel. Four ran on local GPUs and two through providers' APIs that allow the reply to be prefilled, which the forced baseline needs. Generation limits were raised for these runs, and every output was cut, token by token, at the panel's limits for the primary analysis; results from the uncut outputs are in Supplementary Table~\siTabLaduncut. Numerical precision differed between models (16-bit, 8-bit or 4-bit floating point, or not stated by the provider), and one provider's endpoint did not repeat greedy outputs exactly, so each of that model's outputs is a single draw (Supplementary Section~\siSiLarger). Asked for a direct answer, the Qwen model usually wrote the bare answer without the box the prompt asks for (\ladQwenReqBareGsm{} of \ladQwenReqRowsGsm{} zero-shot requested outputs on GSM8K), which strict scoring cannot parse, so results for the larger models use content scoring. Each larger model was placed against the panel on the same items and draws, and flipped pairs were counted among the six, among the four of 70 billion parameters or more, and among all models together, with intervals from resampling models \citep{field2007bootstrapping}. With one model per size, no trend across sizes was estimated. The commercial flagship model, Qwen4-Max, with its built-in reasoning mode switched off, ran the three conditions on every template of the four benchmarks and the same \cwItems{} questions per benchmark (\cwRequestsArm{} requests per condition), through an interface that lets the start of its reply be written; its results are reported beside the panel's and do not enter the comparisons of pairs. It was then run on all \cwFullItems{} numeric MATH-500 questions under the first \cwFullTemplates{} discovery templates in the three conditions (\cwFullCalls{} requests), with intervals from resampling questions.

\subsection*{Benchmarks and templates}
We used all 1{,}319 GSM8K test items \citep{cobbe2021training}, the 25 BIG-Bench Hard subtasks with letter, word-pair or integer answers, all but two of its 27 \citep{suzgun2023challenging}, analysed singly and pooled with equal task weights, and the \mtItems{} MATH-500 items with numeric answers \citep{hendrycks2021math,lightman2024lets}. HotpotQA used \hqItems{} questions drawn at random from its distractor validation set, each with its ten given passages, excluding yes/no answers and items too long for the shortest context window \citep{yang2018hotpotqa}; answers were scored by exact match after the official normalization, and its forced baseline allowed 64 tokens, enough for the longest reference answer. For each benchmark, a bank of templates was generated with PromptSuite \citep{habba2025promptsuite}, varying instruction wording, separators and other surface features while keeping the item text and the answer format fixed. \gtKdisc{} discovery draws and \gtKval{} validation draws (the held-out templates of the main text) were sampled from the bank with replacement; each sampled template, called a draw, is applied to every model in every condition. Validation draws that reuse a discovery template are identified, and decisions are valued only on validation templates that no discovery draw used.

\subsection*{Conditions}
Every condition asks for the final answer inside \texttt{\textbackslash boxed\{\}}. The chain-of-thought condition adds the zero-shot trigger of \citet{kojima2022large}, ``Let's think step by step.'' The requested baseline asks for the answer alone. In the forced baseline, the assistant turn is prefilled with \texttt{\textbackslash boxed\{} and at most 16 new tokens are generated; on GSM8K and BIG-Bench Hard, generation also stops at the first closing brace. A forced answer is parseable when the box closes and its content can be parsed as an answer to the item. On BIG-Bench Hard, a constrained forced baseline used the forced baseline's prompts with decoding restricted to the item's answer set, so every answer was parseable. The five-shot condition placed five questions, each followed by its boxed reference answer alone, before the forced request; its baseline is the zero-shot forced baseline, and a further run on GSM8K and BIG-Bench Hard placed the same demonstrations before the requested baseline, so that the five-shot gain could also be measured within the requested baseline. Runs with the built-in reasoning mode switched on, in the \thiGsmM{} checkpoints that have one, are described in Supplementary Section~\siAppData.

\subsection*{Scoring}
Strict scoring parses the answer inside a closed box, with rules for each benchmark: on GSM8K the box must stand on the last non-empty line and hold a single answer; on BIG-Bench Hard the last complete box is parsed, and tasks with integer answers accept only a numeric expression; the rules for MATH-500 are in Supplementary Section~\siAppContent. Content scoring parses each output of the requested, forced and chain-of-thought conditions on the full panels with task-specific rule-based extraction and classifies it as correct, wrong, unparseable, a malformed box or truncated, so that format and correctness can be told apart. Its extraction rules for each benchmark are listed in Supplementary Section~\siAppContent. On HotpotQA, whose forced baseline had no such stop, content scoring parses the last box or, without one, the answer written outside the box; HotpotQA results use it throughout, and its strict scores are in Supplementary Table~\siTabHqstrict. Three of the authors, who saw no scorer's verdict, checked \hraAnswers{} outputs of GSM8K, of BIG-Bench Hard and of a GSM8K run with eight worked examples before the chain-of-thought prompt against the reference answer. Three in four sampled outputs were ones on which strict and content scoring disagreed, so the rates describe this sample. Of the \hrtDisagreeN{} GSM8K and BIG-Bench Hard outputs that strict scoring marked wrong and content scoring marked correct, the raters read \hrtDisagreeContentAgreePct\% as correct (95\% Wilson interval \citep{wilson1927probable}, \hrtDisagreeContentAgreePctLo{} to \hrtDisagreeContentAgreePctHi). Rater assignments and agreement between raters are in Supplementary Section~\siSiHumanaudit.

\subsection*{Agreement between templates and baselines}
Agreement is the share of model pairs whose gains two evaluations order the same way, counting a tie as one half. Template agreement compares two draws under the same baseline; the comparison between baselines uses the same draw under both, or each baseline averaged over its own $K$ distinct templates. Intervals for the share of flipped pairs resample models, or whole families, with the draws held fixed. The reliability of the gain is the covariance across models between the gains measured on two different templates, averaged over all pairs of templates, divided by the variance across models of the gain on one template; it is the share of that variance that reflects stable differences between models and not template noise \citep{brennan2001generalizability}.

\subsection*{Long answers and the corrected gain}
A requested output is long when it runs past 16 generated tokens, the forced baseline's limit; on HotpotQA, whose forced baseline allowed 64 tokens for long names, the same 16 tokens were used, and the tables of the larger and commercial models also give $w$ at 64 tokens. Requested length counts the whole reply and forced length counts the tokens after the opening of the box; counting requested tokens only up to the first complete box lowered the long share from \wckGsmLongPct\% to \wckGsmLongFirstPct\% on GSM8K, from \wckMathLongPct\% to \wckMathLongFirstPct\% on MATH-500 and from \wckBbhLongPct\% to \wckBbhLongFirstPct\% on BIG-Bench Hard and left the results below nearly unchanged (Supplementary Section~\siSiHowlonganswerswere). For model $i$, $w_i$ is its share of long requested outputs on the discovery draws.

Accuracy was compared on the same model, template and question in the requested, forced and chain-of-thought conditions, pooled over all such model--template--question rows, with intervals from resampling models. The decomposition splits each model's requested-minus-forced accuracy into the part from short and the part from long requested outputs, each weighted by its share of rows, and reports the long part's share of the variance of that difference across models. The corrected gain $(1-w_i)(C_i-F_i)$, with $C_i$ and $F_i$ chain-of-thought and forced accuracy, assumes that short requested answers score like forced ones and long ones like chain-of-thought answers; a form that uses only the short rows' chain-of-thought gain gave similar choices (Supplementary Section~\siSiHowlonganswerswere). Flipped pairs were counted between the observed gain over the requested baseline and either the forced gain or the corrected gain.

A language model judge \citep{zheng2023judging} labelled a random sample of \labN{} requested answers, stratified by benchmark, length bin (1--16, 17--64 and more than 64 tokens) and model, with the commercial model's answers added; it saw the prompt and the answer, not the reference answer, the model's name or any score. Each answer was labelled as containing intermediate reasoning (at least one step toward the answer that the question does not state), extra text without such a step, or the answer alone, under a written rubric (Supplementary Section~\siSiReadingaudit). Shares were weighted to the discovery draws, with intervals from bootstrap resampling within strata; \labSelfAgreeUnwPct\% of \labSelfN{} answers judged twice received the same label. Two human raters, blind to the judge's labels and to each other, labelled \labHumanN{} of the same answers; the judge agreed with each rater (Cohen's $\kappa$ \citep{cohen1960coefficient} \labKappaJudgeRaterOne{} and \labKappaJudgeRaterTwo) about as well as the raters agreed with each other ($\kappa$ \labKappaRaterOneRaterTwo).

HotpotQA used the decomposition rule, its threshold and the $(1-w)$ choice rule of the other benchmarks unchanged; nine of its 24 validation draws reuse a discovery template, and outcomes on the other draws are reported apart.

The rewording test reran GSM8K and MATH-500 for all \awMechGsmModels{} models on six templates with four wordings of the request for the final answer and four of the chain-of-thought trigger, crossed, and with two further requests: one with no instruction about working and one that forbade showing any working. A screen of 50 training items on two templates gave each model's $w$ under each wording from token counts alone, and from these and the main collection's discovery draws a prediction stated the share of pairs flipped by a wording change, the share flipped by the baseline switch under each wording, the within-model rise of requested accuracy with $w$, and a smaller spread of forced than of requested accuracy across wordings; each statement was then checked on the rerun. A question's difficulty for a model was 100 minus the percentage of the other models' forced answers to it that were correct. A logistic regression of whether each requested answer was long on difficulty, with one intercept per model, gave the odds ratio per 10 points of difficulty, with 95\% intervals from resampling models; on BIG-Bench Hard the fit added one intercept per task (odds ratio \diffOrBbhSub, \diffOrBbhSubLo{} to \diffOrBbhSubHi; positive in \diffPosBbhSub{} of \diffModelsBbhSub{} models). Five panel models, one per family, were rerun on MATH-500 under the first \tempCheckTemplates{} discovery templates in all three conditions with sampling at temperature \tempCheckTemperature{} (Supplementary Section~\siSiTemperature).

\subsection*{Five-shot answering and built-in reasoning}
The same counts were made for five-shot answering against the zero-shot forced baseline on all three benchmarks, and for the built-in reasoning mode in the \genThiGsmArmM{} checkpoints that have one, on GSM8K, against both the requested and the forced baseline. For these comparisons, the five-shot and reasoning-mode outputs were content-scored with the rules used for chain-of-thought.

\subsection*{Capability slopes}
Capability slopes were fitted across models with the cross-draw estimator, so that a model's capability and its gain come from different templates and their shared noise cannot couple them \citep{oldham1962note,blomqvist1977relation}. Each gain was regressed on its own baseline's accuracy, and both gains on requested-baseline accuracy and on the Epoch Capabilities Index \citep{epoch2025eci}; intervals resample models and, separately, families. The share of the difference between the two own-accuracy slopes reproduced by the corrected gain is the slope of the corrected gain on requested accuracy minus the slope of the forced-baseline gain on forced accuracy, divided by that difference (Supplementary Section~\siSiCapslope).

\subsection*{Decision analysis}
A team gives chain-of-thought to the third of the panel's $M$ models predicted to gain most, $n=\mathrm{round}(M/3)$ models, and ranks the models by their own gains in a pilot. One-draw pilots are the distinct discovery templates, and pilots of two and four draws are 200 random sets. A choice is valued on validation templates that no discovery draw uses. The loss (also called regret) is the mean gain of the best possible third minus that of the chosen third, in points per chosen model, averaged over pilot sets; intervals are $t$ intervals over validation templates. Chosen models receive chain-of-thought and the others answer as the baseline does; a loss per chosen model multiplied by $n/M$ gives the loss per model of the whole panel when every model serves equal traffic. For the cost of the wrong baseline, the saved choices made under each baseline were valued under the other and compared with choices made under that baseline on the same pilot sets, for one-draw pilots on all items and for 100-item pilots in which items are split in halves by a hash of the item identity (within each BIG-Bench Hard task), pilots draw from one half and choices are valued on the other. In the split pilot, the 100 pilot baseline calls per model were divided equally between the requested and the forced baseline (Supplementary Section~\siSiThesplitpilot); its loss under each deployment baseline was compared with those of the two single-baseline pilots of the same size, over 200 pilots, \splitpSeeds{} further seeds and with each family left out. The analyses were repeated without Qwen3.6-35B-A3B, the model whose forced and short requested answers differed most, and with each family left out. For HotpotQA, every rule, the split pilot and the predictions came from its discovery draws alone and were checked once on its validation draws. A rule that gives chain-of-thought to every model whose pilot gain exceeds 2, 5 or 10 points is analysed in Supplementary Section~\siSiThreshold.

\subsection*{Data availability}
The benchmarks are public: GSM8K \citep{cobbe2021training}, MATH-500 \citep{lightman2024lets}, BIG-Bench Hard \citep{suzgun2023challenging} and HotpotQA \citep{yang2018hotpotqa}. Every model output, score and per-draw accuracy generated in this study, with the template banks, draws and item lists, is available from the corresponding author on request.

\subsection*{Code availability}
Code for generation, scoring and every analysis, including a tool that counts $w$ from a pilot, estimates reliability and recomputes a claim of who benefits under both baselines, is available at https://github.com/deanariel-md/spontaneous-reasoning under an MIT licence; version 1.0.0, which with the data reproduces every result reported here, is archived at Zenodo (https://doi.org/10.5281/zenodo.24591807). The counting of $w$ and the correction are also released as a standalone package with no data, cot-baseline, at https://github.com/deanariel-md/cot-baseline (version 1.0.0, https://doi.org/10.5281/zenodo.24591830).
\section*{Acknowledgements}
This research received no external funding. During the preparation of this work, the authors used Claude Opus 5.5 (Anthropic) to assist with coding, copyediting, and language refinement. After using this tool, the authors reviewed, verified, and edited all resulting code and text and take full responsibility for the content of the publication. All other models referenced in this manuscript were evaluated strictly as part of the experimental study and were not used to draft or prepare the work.

\section*{Author contributions}
D.A. conceived and designed the study, carried out the experiments and analyses, and wrote the manuscript. E.H. advised on the study and contributed to its design. H.N. and T.H. contributed to the study design and rated model outputs. R.B. provided scientific advice and guidance throughout the work. S.C. supervised the project. All authors reviewed and approved the final manuscript.

\section*{Competing interests}
The authors declare no competing interests.

\bibliographystyle{unsrtnat}
\bibliography{refs}

\clearpage
\begin{center}{\LARGE\bfseries Supplementary Information}\end{center}
\bigskip

\bigskip\noindent{\Large\bfseries Supplementary results}\par\medskip

\section*{S1. Published claims of who benefits}

Supplementary Table~S1 lists published claims that the benefit of a prompting or inference strategy differs between models, with the place in each paper where the claim is made, how the answer without the strategy was obtained, and the number of prompt templates or exemplar sets behind each condition, as each paper reports them.
{\footnotesize\setlength{\tabcolsep}{3pt}
\begin{longtable}{P{2.5cm}P{2.0cm}P{3.6cm}P{1.5cm}P{2.5cm}P{2.0cm}}
\caption*{\textbf{Supplementary Table S1.} Published claims about which models benefit most from a strategy.}\\
\toprule Study & Strategy & Claim & Where & Answer without the strategy & Prompts per condition \\ \midrule \endfirsthead
\toprule Study & Strategy & Claim & Where & Answer without the strategy & Prompts per condition \\ \midrule \endhead
\bottomrule \endfoot
Brown et al. 2020, NeurIPS (arXiv:2005.14165) & few-shot prompting & Few-shot gain over zero-shot often grows with model capacity, suggesting larger models are better meta-learners & \S{}1; Fig. 1.3 & likelihood over options, or free generation & one prompt format per task \\
Wei et al. 2022, NeurIPS (\url{10.52202/068431-1800}) & few-shot chain-of-thought & CoT helps only at about 100B parameters; hurts most models under 10B & \S{}3.2; App. A.1 & examples with answers only & one exemplar set per benchmark in the main results \\
Kojima et al. 2022, NeurIPS (\url{10.52202/068431-1613}) & zero-shot chain-of-thought & Zero-shot CoT gains grow with model size; ineffective for smaller models & \S{}4.1, Fig. 3 & answer cue such as `The answer is' & single fixed prompt \\
Suzgun et al. 2023, Findings of ACL (\url{10.18653/v1/2023.findings-acl.824}) & chain-of-thought prompting & CoT gain grows with scale; zero or negative for smaller InstructGPT models & \S{}4.2 & examples with answers only & one 3-shot exemplar set per task \\
Chung et al. 2024, JMLR (arXiv:2210.11416) & instruction finetuning; CoT prompting & Instruction finetuning gain larger for 8B in absolute terms, 540B in relative error reduction & App. A.3; App. A.2 & examples with answers only & fixed prompts per benchmark \\
Wang et al. 2023, ICLR (arXiv:2203.11171) & self-consistency decoding & Self-consistency gains grow with model scale (+3-6\% at 20B, +9-23\% at 137B and 175B, arithmetic) & \S{}3.2; \S{}3.5, Fig. 4 & chain-of-thought (the strategy is self-consistency) & one exemplar set, same as Wei et al. \\
Mallen et al. 2023, ACL (\url{10.18653/v1/2023.acl-long.546}) & retrieval; GenRead prompting & Retrieval helps smaller models most; GenRead helps GPT-3 but not smaller models & \S{}5.2; Fig. 7 & answer cue `A:' (15-shot for GPT-Neo and OPT, zero-shot for GPT-3) & one Q:/A: template \\
Madaan et al. 2023, NeurIPS (\url{10.52202/075280-2019}) & iterative self-refinement feedback & Self-Refine suits stronger models like GPT-4; Vicuna-13B struggles with refinement & \S{}3.3; \S{}4 & first answer from the task prompt (the strategy is refinement) & one few-shot prompt set per task \\
Deng et al. 2023, arXiv preprint (arXiv:2311.04205) & Rephrase and Respond prompting & Rephrase-and-Respond gains largest for GPT-4, modest for Vicuna & \S{}4.3 & question only, free reply & two-step RaR (rephrase, then respond) \\
Li et al. 2023, arXiv preprint (arXiv:2307.11760) & emotional stimuli prompts & Larger models may gain more from emotional stimuli; data show mixed ordering & \S{}3.4.1; Table 6 & original task prompt, free reply & 11 emotional stimuli averaged \\
Bsharat et al. 2023, arXiv preprint (arXiv:2312.16171) & principled prompt instructions & Improvements grow with model size; gains exceed 20\% from LLaMA-2-7B to GPT-4 & \S{}1; \S{}4, Fig. 6 & question only, free reply & one principled prompt per question \\
Hebenstreit et al. 2024, PeerJ Computer Science (\url{10.7717/peerj-cs.1999}) & zero-shot reasoning strategies & GPT-4 benefits most from reasoning strategies; CoT gains robust across models and datasets & abstract & question only, free reply & 10 zero-shot prompts, one wording per strategy \\
Xu et al. 2024, EMNLP (\url{10.18653/v1/2024.emnlp-main.871}) & re-reading (Re2) prompting & Re-reading helps consistently except in a few scenarios on vanilla ChatGPT & abstract & answer cue `A:' after an answer-format instruction & one Re2 template \\
Zheng et al. 2024, Findings of EMNLP (\url{10.18653/v1/2024.findings-emnlp.888}) & persona system prompts & Personas have no significant effect in most cases, consistently across six models and sizes & \S{}4, Fig. 4--5 & asked for the answer only & 4 prompt templates \\
Liu et al. 2025, ICML (arXiv:2410.21333) & chain-of-thought prompting & On artificial grammar, CoT lowers performance more for strong models; weaker Llama 3.1 8B drops less & \S{}4.1 & asked for the answer only & one zero-shot and one CoT prompt per task \\
Meincke et al. 2025, arXiv preprint (arXiv:2506.07142) & chain-of-thought prompting & CoT helps non-reasoning models, especially older or smaller; negligible for dedicated reasoning models & Summary & asked for the answer only; also question only, free reply & one CoT wording; 25 trials per question \\
Cheng et al. 2025, Findings of EMNLP (\url{10.18653/v1/2025.findings-emnlp.729}) & few-shot CoT exemplars & Traditional CoT exemplars add nothing over zero-shot CoT for strong models like Qwen2.5 & abstract & zero-shot chain-of-thought (the strategy is exemplars) & not stated \\
Lin et al. 2026, EMNLP (arXiv:2605.30621) & updated agent harness & Harness benefit is non-monotonic in capability: mid-tier models benefit most & abstract & agent with the original harness & not stated \\
Eskandari Miandoab and Sarathy 2024, CustomNLP4U (\url{10.18653/v1/2024.customnlp4u-1.20}) & argument generation prompting & Argument generation helps models up to 8B most (4.7-4.9\%), the largest marginally (0.45\%) & \S{}6.3; \S{}7 & asked for the answer only & one prompt per method \\
Khan 2025, arXiv preprint (arXiv:2510.22251) & constrained chain-of-thought prompting & Constrained CoT helps gpt-4o but not gpt-5, where it trails standard CoT; simpler prompts for capable models & abstract; Fig. 1 & question only, free reply & one prompt per strategy \\
Wang et al. 2024, arXiv preprint (arXiv:2411.02093) & prompt engineering techniques & Earlier prompt techniques may give diminished or negative benefit on advanced models & abstract & task instruction only, free reply & not stated \\
Rudyk et al. 2026, ICSME (arXiv:2608.24641) & structured prompting techniques & Newer GPT models gain less or negatively; Qwen models keep benefiting from few-shot & abstract & asked for the answer only (a Python function in one code block) & one template per technique \\
Preet et al. 2026, arXiv preprint (arXiv:2607.14109) & prompting techniques; enabling reasoning & Technique gains are small; benefit of enabling reasoning varies by model, +2.10 to +12.74 pp & abstract, \S{}6 & asked for the answer only & one public template per technique \\
DeepSeek-AI 2025, arXiv preprint (arXiv:2501.12948v1) & few-shot prompting & Few-shot prompting consistently degrades DeepSeek-R1 performance; zero-shot recommended & v1 \S{}5, `Prompting Engineering' & not stated & not stated \\
Ruan et al. 2024, NeurIPS (\url{10.52202/079017-0506}) & chain-of-thought; self-consistency & CoT benefit is predictable from capability; Phi benefits earlier but scales less rapidly & \S{}4.3; \S{}6 & examples with answers only & one prompt per arm \\
\end{longtable}}

\section*{S2. Long answers and the decomposition}

Requested length counts the whole reply. Counting only the tokens up to the first complete box, the long share fell from \wckGsmLongPct\% to \wckGsmLongFirstPct\% on GSM8K, \wckMathLongPct\% to \wckMathLongFirstPct\% on MATH-500 and \wckBbhLongPct\% to \wckBbhLongFirstPct\% on BIG-Bench Hard; long answers' share of the between-model variance became \wckGsmVarLongPctFirst\%, \wckMathVarLongPctFirst\% and \wckBbhVarLongPctFirst\%, and the corrected gains reversed \wckGsmRevCorrFirst, \wckMathRevCorrFirst{} and \wckBbhRevCorrFirst{} pairs.

Each model's difference between requested and forced accuracy was split into the part from questions it answered at length and the part from questions it answered briefly, and the pairs reversed by the baseline switch were recounted with the difference replaced by each part alone (discovery draws).

\begin{table}[h]\centering\small
\begin{tabular}{@{}lccc@{}}
\toprule
 & Pairs reversed & Reproduced by the long part & Reproduced by the short part \\
\midrule
GSM8K, strict & \attrGsmObsRev & \attrGsmLongRevShared & \attrGsmShortRevShared \\
MATH-500, strict & \attrMathObsRev & \attrMathLongRevShared & \attrMathShortRevShared \\
BIG-Bench Hard, strict & \attrBbhObsRev & \attrBbhLongRevShared & \attrBbhShortRevShared \\
GSM8K, content & \attrGsmObsRevContent & \attrGsmLongRevSharedContent & \attrGsmShortRevSharedContent \\
MATH-500, content & \attrMathObsRevContent & \attrMathLongRevSharedContent & \attrMathShortRevSharedContent \\
BIG-Bench Hard, content & \attrBbhObsRevContent & \attrBbhLongRevSharedContent & \attrBbhShortRevSharedContent \\
\bottomrule
\end{tabular}
\caption*{\textbf{Supplementary Table S2.} Pairs whose order of chain-of-thought gain is reversed by the baseline switch, and how many of them each part of the baseline difference reverses on its own. On BIG-Bench Hard under strict scoring, short requested answers often left out the answer box, which content scoring parses.}

\end{table}

In the median model, short requested answers scored \bmixDecMathShortGap\pp{} from forced answers to the same questions on MATH-500 and \bmixDecGsmShortGap\pp{} on GSM8K, and long requested answers closed \bmixDecMathRecov{} of the gap between forced and chain-of-thought accuracy on MATH-500.

The corrected gain multiplies each model's gain over the forced baseline on all rows by $1-w$. A form that uses only the chain-of-thought gain on the rows whose requested answer was short chose models with a total loss under the requested baseline of \bmixDecGsmChoiceValShort\pp{} on GSM8K, \bmixDecMathChoiceValShort{} on MATH-500 and \bmixDecBbhChoiceValShort{} on BIG-Bench Hard, against \bmixDecGsmChoiceValCorr, \bmixDecMathChoiceValCorr{} and \bmixDecBbhChoiceValCorr\pp{} for the corrected gain.

Table~S3 repeats the decomposition and the corrected-gain count at cutoffs of 8, 32 and 64 tokens. The 16-token cutoff is the forced baseline's own limit: at 8 tokens, long answers include some the forced baseline could have produced, and at 32 or 64, short answers include some it could not.

\begin{table}[ht]
\centering\small
\caption*{\textbf{Supplementary Table S3.} The cutoff for a long answer. Long: share (\%) of the between-model variance of requested-minus-forced accuracy from long requested answers. Rev.: pairs ordered differently by the gain over the requested baseline and by the corrected gain $(1-w)(C-F)$, discovery means. $\bar w$: mean $w$ over models (\%). HotpotQA was examined at 16 and 64 tokens only.}
\adjustbox{max width=\textwidth}{%
\begin{tabular}{@{}llrrrrrrrrrrrrr@{}}
\toprule
 & & & \multicolumn{3}{c}{8 tokens} & \multicolumn{3}{c}{16 tokens} & \multicolumn{3}{c}{32 tokens} & \multicolumn{3}{c}{64 tokens}\\
\cmidrule(lr){4-6}\cmidrule(lr){7-9}\cmidrule(lr){10-12}\cmidrule(l){13-15}
Benchmark & Scoring & Pairs & Long (\%) & Rev. & $\bar w$ (\%) & Long (\%) & Rev. & $\bar w$ (\%) & Long (\%) & Rev. & $\bar w$ (\%) & Long (\%) & Rev. & $\bar w$ (\%)\\
\midrule
GSM8K & strict & 406 & 96 & 115 & 73 & 92 & 48 & 43 & 91 & 21 & 30 & 83 & 19 & 21\\
 & content & 406 & 99 & 124 & 73 & 95 & 53 & 43 & 94 & 27 & 30 & 86 & 24 & 21\\
MATH-500 & strict & 406 & 84 & 51 & 68 & 83 & 24 & 43 & 83 & 27 & 33 & 82 & 28 & 26\\
 & content & 406 & 95 & 62 & 68 & 95 & 25 & 43 & 95 & 28 & 33 & 95 & 25 & 26\\
BIG-Bench Hard & strict & 378 & 92 & 78 & 58 & 53 & 70 & 32 & 39 & 66 & 19 & 29 & 63 & 10\\
 & content & 378 & 99 & 80 & 58 & 95 & 47 & 32 & 89 & 34 & 19 & 78 & 41 & 10\\
HotpotQA & strict & 406 & -- & -- & -- & 20 & 152 & 37 & -- & -- & -- & 0 & 142 & 2\\
 & content & 406 & -- & -- & -- & 59 & 88 & 37 & -- & -- & -- & 1 & 84 & 2\\
\bottomrule
\end{tabular}
}
\end{table}

\section*{S3. Reading the requested answers}

The judge, \labJudge{} with its reasoning mode off and its default sampling, read each sampled requested answer with its full prompt, and saw neither the reference answer, the model's name nor any score. The rubric defines three labels. Intermediate reasoning: at least one step toward the answer that the question does not state, such as a calculation, a deduction or facts combined toward the answer, wherever it appears in the reply and whether or not it is right. Extra text: words beside the answer with no such step, such as a restatement of the question, a preamble, a format remark or an explanation that adds no step. Answer only: the answer and nothing else. Repeating numbers, facts or options that the prompt states is not a step. The sample of \labN{} answers was stratified by benchmark, length bin (1--16, 17--64 and more than 64 tokens) and model, with a fixed seed, and \labNQwen{} answers of the commercial model were added; \labNUnread{} answers could not be read and were left out. Shares are weighted to the discovery draws, each template counted as often as it was drawn, with 95\% intervals from 2,000 bootstrap draws within strata. Of \labSelfN{} answers judged twice, \labSelfAgreeUnwPct\% received the same label (Cohen's $\kappa$ \labSelfKappaUnw). Two human raters, blind to the judge's labels and to each other, labelled \labHumanN{} of the same answers with the same rubric. Weighted to the sample design, the judge agreed with rater 1 on \labAgreeJudgeRaterOnePct\% of answers ($\kappa$ \labKappaJudgeRaterOne) and with rater 2 on \labAgreeJudgeRaterTwoPct\% ($\kappa$ \labKappaJudgeRaterTwo), and the raters agreed with each other on \labAgreeRaterOneRaterTwoPct\% ($\kappa$ \labKappaRaterOneRaterTwo); most disagreements in every pair were between reasoning and extra text. The judging cost \labCostUsd{} US dollars.

Table~S4 gives the results. Short answers almost never reasoned, so the 16-token cutoff found nearly every reasoning answer; among long answers, the share that reasoned rose from the 17--64-token bin to the longest, where nearly all did. On HotpotQA, whose answers are names, \labRShortHqSixtyFourPct\% of answers of 64 tokens or less reasoned, so a 64-token cutoff found only \labRecallHqSixtyFourPct\% of the reasoning answers. The mixture behind the corrected gain held roughly on the same questions: the accuracy of answers labelled as reasoning differed from chain-of-thought accuracy by \labGapReqCotRGsm{} points on GSM8K and \labGapReqCotRMath{} on MATH-500, and that of the other answers differed from forced accuracy by \labGapReqForcedNotRGsm{} and \labGapReqForcedNotRMath{} points.

\begin{table}[ht]
\centering\small
\caption*{\textbf{Supplementary Table S4.} Reading the requested answers. R: share (\%) of answers labelled as containing intermediate reasoning, among short and long answers at the cutoff shown, with 95\% intervals; Recall: share of reasoning answers that are long; $r$: correlation across models of the judged share with $w$; $R^2$: share of the variance across models of requested-minus-forced accuracy (strict) explained by the judged share and by $w$; the last two columns compare accuracy on the same questions (strict, points).}
\adjustbox{max width=\textwidth}{%
\begin{tabular}{@{}lrrrrrrrrrr@{}}
\toprule
Benchmark & $n$ & Cut & R, short (\%) & R, long (\%) & Recall (\%) & $r(\text{R}, w)$ & $R^2$ R & $R^2$ $w$ & Req.$-$CoT, R & Req.$-$forced, not R \\
\midrule
GSM8K & 607 & 16 & 1 [0, 3] & 63 [60, 67] & 98 & 0.84 & 0.66 & 0.41 & $-$5.9 [$-$13.1, 0.8] & $-$5.5 [$-$9.1, $-$1.9] \\
MATH-500 & 607 & 16 & 0 [0, 0] & 68 [64, 72] & 100 & 0.87 & 0.61 & 0.32 & $-$7.5 [$-$13.9, $-$1.6] & $-$4.2 [$-$8.4, $-$0.2] \\
BIG-Bench Hard & 588 & 16 & 0 [0, 0] & 44 [39, 50] & 100 & 0.78 & 0.16 & 0.01 & 2.0 [$-$4.4, 7.3] & $-$2.1 [$-$6.9, 2.2] \\
HotpotQA & 609 & 16 & 0 [0, 0] & 20 [14, 26] & 100 & 0.41 & 0.00 & 0.00 & $-$4.1 [$-$22.9, 14.0] & $-$12.2 [$-$15.7, $-$8.5] \\
 & & 64 & 6 [4, 9] & 60 [52, 69] & 13 & & & & & \\
Qwen4-Max, GSM8K & 55 & 16 & 0 [0, 0] & 100 [100, 100] & 100 & & & & & \\
Qwen4-Max, MATH-500 & 55 & 16 & 0 [0, 0] & 100 [100, 100] & 100 & & & & & \\
\bottomrule
\end{tabular}
}
\end{table}

\section*{S4. Rewording the request and the trigger}

The rewording run repeated GSM8K and MATH-500 for all \awMechGsmModels{} models on six templates, crossing four wordings of the request for a bare answer with four wordings of the chain-of-thought trigger. Table~S5 gives the share of model pairs reversed by each change, with benefits averaged over the six templates.

\begin{table}[h]\centering\small
\caption*{\textbf{Supplementary Table S5.} Share of model pairs whose order of chain-of-thought gain flips when one element of the prompt changes (strict scoring).}
\begin{tabular}{lcc}
\toprule
Change & GSM8K & MATH-500 \\
\midrule
Baseline switch, first wording of the request & \awGsmArmSwitchClauseOne & \awMathArmSwitchClauseOne \\
Baseline switch, second wording & \awGsmArmSwitchClauseTwo & \awMathArmSwitchClauseTwo \\
Baseline switch, third wording & \awGsmArmSwitchClauseThree & \awMathArmSwitchClauseThree \\
Baseline switch, fourth wording & \awGsmArmSwitchClauseFour & \awMathArmSwitchClauseFour \\
Wording of the request, requested baseline & \awGsmClauseSwitchReqReversed & \awMathClauseSwitchReqReversed \\
Wording of the request, forced baseline & \awGsmClauseSwitchFdReversed & \awMathClauseSwitchFdReversed \\
Wording of the trigger, requested baseline & \awGsmTriggerSwitchReqReversed & \awMathTriggerSwitchReqReversed \\
Template, requested baseline & \awGsmConfigSwitchReqReversed & \awMathConfigSwitchReqReversed \\
\bottomrule
\end{tabular}
\end{table}

Under the four wordings, mean requested-minus-forced accuracy on GSM8K was \awGsmGapClauseOneMean, \awGsmGapClauseTwoMean, \awGsmGapClauseThreeMean{} and \awGsmGapClauseFourMean\pp. Choosing models under the current wording and valuing the choice under another wording of the request cost \awGsmDecClauseExtra\pp{} per chosen model on GSM8K (\awGsmDecClauseExtraLo{} to \awGsmDecClauseExtraHi) and \awMathDecClauseExtra\pp{} on MATH-500, against \awGsmDecArmExtra{} and \awMathDecArmExtra\pp{} for the baseline switch; a new wording of the trigger cost \awGsmDecTriggerExtra\pp.

The prediction from the screen's token counts made five statements per benchmark, all of which held on both: the reversal share of a change of wording within the requested baseline (GSM8K \awMechGsmClauseObs\%, predicted \awMechGsmClausePredLo--\awMechGsmClausePredHi\%; MATH-500 \awMechMathClauseObs\%, predicted \awMechMathClausePredLo--\awMechMathClausePredHi\%); a larger reversal share for a change of wording than for a change of template; the baseline-switch share under each wording within its predicted range (\awMechGsmArmHoldsCount{} of \awMechGsmArmClauses{} and \awMechMathArmHoldsCount{} of \awMechMathArmClauses); a within-model rise of requested accuracy with $w$ (\awMechGsmSlopePositive{} of \awMechGsmSlopeModels{} and \awMechMathSlopePositive{} of \awMechMathSlopeModels{} models); and a smaller spread of forced than of requested accuracy across wordings (\awMechGsmFdLess{} and \awMechMathFdLess{} of \awMechGsmModels{} models). The screen's $w$ and the rerun's $w$ correlated at \awMechGsmScreenCollCorr{} and \awMechMathScreenCollCorr.

\section*{S5. Robustness of the flips and of the loss}

The limit of agreement is the share of model pairs that the two baselines would order alike if each averaged ever more templates, estimated from all distinct templates with a 95\% bootstrap interval over templates. Under strict scoring it was \pceilGsmArmLimitPlug{} on GSM8K, \pceilMathArmLimitPlug{} on MATH-500 and \pceilBbhArmLimitPlug{} on BIG-Bench Hard; on HotpotQA it was \pceilHqArmLimitPlugContent{} under content scoring (Table~S6). Under content scoring it was \pceilGsmArmContentLimitPlug{} on GSM8K (95\% bootstrap interval over templates, \pceilGsmArmContentLimitPlugLo{} to \pceilGsmArmContentLimitPlugHi), \pceilMathArmContentLimitPlug{} on MATH-500 (\pceilMathArmContentLimitPlugLo{} to \pceilMathArmContentLimitPlugHi) and \pceilBbhArmContentLimitPlug{} on BIG-Bench Hard (\pceilBbhArmContentLimitPlugLo{} to \pceilBbhArmContentLimitPlugHi). Resampling whole families gave intervals of \ddirGsmRevFamBootLo{} to \ddirGsmRevFamBootHi, \ddirMathRevFamBootLo{} to \ddirMathRevFamBootHi{} and \ddirBbhRevFamBootLo{} to \ddirBbhRevFamBootHi{} for the share of flipped pairs, and within single families \ddirGsmRevWithinFam{} of \ddirGsmRevWithinFamPairs, \ddirMathRevWithinFam{} of \ddirMathRevWithinFamPairs{} and \ddirBbhRevWithinFam{} of \ddirBbhRevWithinFamPairs{} pairs flipped. Leaving out one family at a time, the added loss of choosing under the forced baseline and deploying under the requested one ranged from \ddirGsmFdToReqLofoMin{} to \ddirGsmFdToReqLofoMax\pp{} on GSM8K, \ddirMathFdToReqLofoMin{} to \ddirMathFdToReqLofoMax\pp{} on MATH-500 and \ddirBbhFdToReqLofoMin{} to \ddirBbhFdToReqLofoMax\pp{} on BIG-Bench Hard. With 100-question pilots on split halves of the questions (Methods), it was \ddirGsmPilotFdToReq, \ddirMathPilotFdToReq{} and \ddirBbhPilotFdToReq\pp, and the reverse \ddirGsmPilotReqToFd, \ddirMathPilotReqToFd{} and \ddirBbhPilotReqToFd\pp. Valued on content-scored gains, the one-draw added losses were \ddirGsmFdToReqContent, \ddirMathFdToReqContent{} and \ddirBbhFdToReqContent\pp{} forced to requested and \ddirGsmReqToFdContent, \ddirMathReqToFdContent{} and \ddirBbhReqToFdContent\pp{} requested to forced.

Spread over every model of the panel, the larger added loss of choosing under the wrong baseline was \wckGsmFdToReqPanel\pp{} per model on GSM8K, \wckMathFdToReqPanel{} on MATH-500 and \wckBbhFdToReqPanel{} on BIG-Bench Hard.

On BIG-Bench Hard, a constrained forced baseline used the forced baseline's prompts with decoding restricted to the item's answer options, so that every forced answer was parseable. Switching from the requested baseline to it flipped \wckBbhConRevContent{} of \wckBbhConPairs{} pairs (\wckBbhConRevContentPct\%) under content scoring and \wckBbhConRev{} (\wckBbhConRevPct\%) under strict scoring.

\section*{S6. HotpotQA}

HotpotQA was run after the analyses of the other three benchmarks were complete, with the same \hqM{} models, \hqItems{} questions and \hqReqKdisc{} discovery and \hqReqKval{} validation draws. Every rule, choice and prediction came from the discovery draws alone and was then checked on the validation draws; \hqFreshDraws{} validation draws use \hqFreshConfigs{} templates that no discovery draw used, and the decision analyses are valued on these. HotpotQA is scored by exact match on the answer that content scoring extracts. Mean chain-of-thought gain was \hqReqMeanGainContent\pp{} against the requested baseline and \hqFdMeanGainContent\pp{} against the forced one.

In the requested baseline, \fmtHqReqPlaceholderPct\% of outputs copied the instruction's placeholder, \texttt{\textbackslash boxed\{<answer>\}}, usually after the answer, and \fmtHqReqUnboxedPct\% gave no usable box; strict exact match scores both as wrong, and the forced baseline, whose reply begins inside the box, produces neither. Table~S6 gives the main HotpotQA results under both scorings.

\begin{table}[h]\centering\small
\caption*{\textbf{Supplementary Table S6.} HotpotQA under strict and content exact match.}
\begin{tabular}{P{8.4cm}cc}
\toprule
 & Strict & Content \\
\midrule
Requested baseline accuracy (\%) & \hqArmAccReq & \hqArmAccReqContent \\
Forced baseline accuracy (\%) & \hqArmAccFd & \hqArmAccFdContent \\
Median requested minus forced accuracy across models (pp) & \hqGapReqFdMedian & \hqGapReqFdMedianContent \\
Pairs reversed by the baseline switch, of \aeHqPairs & \aeHqReversed & \aeHqReversedContent \\
Agreement between baselines, limit as templates grow & \pceilHqArmLimitPlug & \pceilHqArmLimitPlugContent \\
Long answers' share of the baseline-shift variance, 16 tokens (\%) & \hqDecVarLongPctSixteen & \hqDecVarLongPctSixteenContent \\
Cost, chosen under forced and used with requested (pp) & \hqDdirFdToReq & \hqDdirFdToReqContentPostHoc \\
Cost, chosen under requested and used with forced (pp) & \hqDdirReqToFd & \hqDdirReqToFdContentPostHoc \\
Worse deployment loss, better single-baseline pilot (pp) & \hqSplitBestSingleWorst & \hqSplitBestSingleWorstContentPostHoc \\
Worse deployment loss, split pilot (pp) & \hqSplitSplitWorst & \hqSplitSplitWorstContentPostHoc \\
\bottomrule
\end{tabular}
\end{table}

\begin{table}[h]\centering\small
\caption*{\textbf{Supplementary Table S7.} HotpotQA: predictions from the discovery draws and their outcomes on the validation draws. Long answers are those over 64 tokens, the cutoff of HotpotQA's forced baseline. The content-scored outcomes use choices made with the final content-scoring rules.}
\begin{tabular}{P{6.2cm}P{1.8cm}P{2.9cm}P{2.9cm}}
\toprule
Statement & Predicted & Outcome, strict & Outcome, content \\
\midrule
Switching the baseline reverses more pairs than a new template & yes & \hqTrSwitchReversed{} against \hqTrNewDrawReversed & \hqTrSwitchReversedContent{} against \hqTrNewDrawReversedContent \\
One-draw agreement within the requested baseline & \hqReqTrPredC & \hqReqTrObsC & \hqReqTrObsCContent \\
One-draw agreement within the forced baseline & \hqFdTrPredC & \hqFdTrObsC & \hqFdTrObsCContent \\
Long answers' share of the baseline-shift variance & at least \hqDecVarThresholdPct\% & \hqDecVarLongPct\%, not met & \hqDecVarLongPctContent\%, not met \\
Choosing by $(1-w)$ lowers the total loss under the requested baseline & yes & \hqDecRegretCorr{} against \hqDecRegretFd\pp, not met & \hqDecRegretCorrContent{} against \hqDecRegretFdContent\pp, not met \\
Split pilot lowers the worse deployment loss & yes & \hqSplitBestSingleWorst{} to \hqSplitSplitWorst\pp & \hqSplitBestSingleWorstContentPostHoc{} to \hqSplitSplitWorstContentPostHoc\pp \\
\bottomrule
\end{tabular}
\end{table}

Choosing under the forced baseline and valuing under the requested one cost \hqDdirFdToReqContentPostHoc\pp{} per chosen model (95\% interval \hqDdirFdToReqLoContentPostHoc{} to \hqDdirFdToReqHiContentPostHoc), and the reverse \hqDdirReqToFdContentPostHoc\pp{} (\hqDdirReqToFdLoContentPostHoc{} to \hqDdirReqToFdHiContentPostHoc); with the 100-question pilots, \hqDdirPilotFdToReqContentPostHoc{} and \hqDdirPilotReqToFdContentPostHoc\pp.

\section*{S7. Published claims re-tested}

Table~S8 re-tests published claims about who benefits under both baselines. Symbolic tasks are the \patSpragueSymFamilies{} BIG-Bench Hard task families marked algorithmic by \citet{suzgun2023challenging}, and language tasks the other \patSpragueLangFamilies. A size ladder is a set of models of one family and one generation with at least two parameter counts. Within each ladder, the slope of gain on the base-2 logarithm of the parameter count was fitted in each draw and averaged over draws, and the smallest model's gain was taken with a delete-one-draw jackknife interval.

\begin{table}[p]
\centering\small
\caption*{\textbf{Supplementary Table S8.} \textbf{Published claims about who benefits, re-tested on our data.} Requested baseline and strict scoring unless stated. Size ladders are models of one family and one generation that differ only in size; counts are ladders in which gain rose with size, or in which the smallest model lost accuracy from chain-of-thought.}

\begin{tabular}{@{}P{3.6cm}P{2.0cm}P{6.6cm}P{2.6cm}@{}}
\toprule
Claim & Benchmark & Our result & Verdict\\
\midrule
Chain-of-thought helps symbolic tasks more than language tasks \citep{sprague2025cot} & BIG-Bench Hard & symbolic tasks gained \patSpragueBaseSymMinusLang\pp{} more (forced baseline \patSpragueForcedSymMinusLang, content scores \patSpragueContentSymMinusLang) & holds under both baselines and content scoring\\
Gain from chain-of-thought rises with size \citep{kojima2022large} & GSM8K & \rswGsmReqPrimaryRise{} of \rswGsmReqPrimaryLadders{} ladders (requested), \rswGsmForcedPrimaryRise{} of \rswGsmForcedPrimaryLadders{} (forced) & depends on arm\\
 & MATH-500 & \rswMathReqPrimaryRise{} of \rswMathReqPrimaryLadders{} (requested), \rswMathForcedPrimaryRise{} of \rswMathForcedPrimaryLadders{} (forced) & depends on arm\\
 & BIG-Bench Hard & \rswBbhReqPrimaryRise{} of \rswBbhReqPrimaryLadders{} (requested), \rswBbhForcedPrimaryRise{} of \rswBbhForcedPrimaryLadders{} (forced) & mostly holds\\
Chain-of-thought hurts the smallest models \citep{wei2022chain} & all three & smallest model lost in \rswGsmReqPrimarySmallLoses{} to \rswBbhForcedPrimarySmallLoses{} of \rswGsmReqPrimaryLadders{} ladders per benchmark and arm & rarely holds\\
\bottomrule
\end{tabular}
\end{table}

\section*{S8. Capability slopes}

Each chain-of-thought gain of each model was regressed across models on the accuracy of its own baseline, with the cross-draw estimator (accuracy and gain from different templates, so that template noise cannot couple them). The gain over the requested baseline rose less steeply on all three benchmarks, and on GSM8K the two slopes had opposite signs; the paired difference excluded zero with models resampled on all three benchmarks and with families resampled on GSM8K and MATH-500 (Table~S9).

\begin{table}[h]
\centering\small
\begin{tabular}{@{}lccc@{}}
\toprule
 & Gain over requested baseline & Gain over forced baseline & Difference \\
\midrule
GSM8K & \crvGsmOwnReq{} (\crvGsmOwnReqLo, \crvGsmOwnReqHi) & \crvGsmOwnFd{} (\crvGsmOwnFdLo, \crvGsmOwnFdHi) & \crvGsmOwnDiff{} (\crvGsmOwnDiffLo, \crvGsmOwnDiffHi) \\
MATH-500 & \crvMathOwnReq{} (\crvMathOwnReqLo, \crvMathOwnReqHi) & \crvMathOwnFd{} (\crvMathOwnFdLo, \crvMathOwnFdHi) & \crvMathOwnDiff{} (\crvMathOwnDiffLo, \crvMathOwnDiffHi) \\
BIG-Bench Hard & \crvBbhOwnReq{} (\crvBbhOwnReqLo, \crvBbhOwnReqHi) & \crvBbhOwnFd{} (\crvBbhOwnFdLo, \crvBbhOwnFdHi) & \crvBbhOwnDiff{} (\crvBbhOwnDiffLo, \crvBbhOwnDiffHi) \\
\bottomrule
\end{tabular}
\caption*{\textbf{Supplementary Table S9.} Slope of each chain-of-thought gain on the accuracy of its own baseline, in points of gain per point of accuracy, strict scoring, with 95\% intervals from resampling models. With families resampled, the interval of the difference was \crvGsmOwnDiffFamLo{} to \crvGsmOwnDiffFamHi{} on GSM8K, \crvMathOwnDiffFamLo{} to \crvMathOwnDiffFamHi{} on MATH-500 and \crvBbhOwnDiffFamLo{} to \crvBbhOwnDiffFamHi{} on BIG-Bench Hard.}

\end{table}

With both gains regressed on requested-baseline accuracy, the gain over the requested baseline also rose less steeply on all three benchmarks (Table~S10).

\begin{table}[h]
\centering\small
\begin{tabular}{@{}lccc@{}}
\toprule
 & Gain over requested baseline & Gain over forced baseline & Difference \\
\midrule
GSM8K & \crvGsmReqAccReq{} (\crvGsmReqAccReqLo, \crvGsmReqAccReqHi) & \crvGsmReqAccFd{} (\crvGsmReqAccFdLo, \crvGsmReqAccFdHi) & \crvGsmReqAccDiff{} (\crvGsmReqAccDiffLo, \crvGsmReqAccDiffHi) \\
MATH-500 & \crvMathReqAccReq{} (\crvMathReqAccReqLo, \crvMathReqAccReqHi) & \crvMathReqAccFd{} (\crvMathReqAccFdLo, \crvMathReqAccFdHi) & \crvMathReqAccDiff{} (\crvMathReqAccDiffLo, \crvMathReqAccDiffHi) \\
BIG-Bench Hard & \crvBbhReqAccReq{} (\crvBbhReqAccReqLo, \crvBbhReqAccReqHi) & \crvBbhReqAccFd{} (\crvBbhReqAccFdLo, \crvBbhReqAccFdHi) & \crvBbhReqAccDiff{} (\crvBbhReqAccDiffLo, \crvBbhReqAccDiffHi) \\
\bottomrule
\end{tabular}
\caption*{\textbf{Supplementary Table S10.} Slope of each chain-of-thought gain on requested-baseline accuracy, in points of gain per point of accuracy, strict scoring, with 95\% intervals from resampling models. With families resampled, the interval of the difference was \crvGsmReqAccDiffFamLo{} to \crvGsmReqAccDiffFamHi{} on GSM8K, \crvMathReqAccDiffFamLo{} to \crvMathReqAccDiffFamHi{} on MATH-500 and \crvBbhReqAccDiffFamLo{} to \crvBbhReqAccDiffFamHi{} on BIG-Bench Hard.}

\end{table}

The corrected gain $(1-w)(C-F)$, regressed on requested accuracy, reproduced \crvGsmOwnShare\% of the difference between the own-accuracy slopes on GSM8K (95\% interval \crvGsmOwnShareLo{} to \crvGsmOwnShareHi), \crvMathOwnShare\% on MATH-500 (\crvMathOwnShareLo{} to \crvMathOwnShareHi) and \crvBbhOwnShare\% on BIG-Bench Hard (\crvBbhOwnShareLo{} to \crvBbhOwnShareHi), with models resampled.

On the Epoch Capabilities Index \citep{epoch2025eci}, a scale shared by both baselines, each model's chain-of-thought gain, averaged over the discovery templates, was regressed across models separately for the gain over the requested and over the forced baseline (\capSlopeModelsGsm{} models on GSM8K and MATH-500, \capSlopeModelsBbh{} on BIG-Bench Hard; every model has an index value). Gains rose with the index under both baselines, and less steeply under the requested baseline on all three benchmarks (Table~S11). Repeating the fit with content scoring, and with forced-baseline or chain-of-thought accuracy as the capability axis (slopes estimated with the axis and the gain taken from different templates), the requested slope was the shallower in all \capShallowerCells{} of \capCells{} combinations of benchmark, scorer and axis; the interval of the difference excluded zero in \capDiffExclCells{} of them, with models and with families resampled.

\begin{table}[h]
\centering\small
\begin{tabular}{@{}lccc@{}}
\toprule
 & Requested baseline & Forced baseline & Difference \\
\midrule
GSM8K & \capSlopeGsmEciReq{} (\capSlopeGsmEciReqLo, \capSlopeGsmEciReqHi) & \capSlopeGsmEciFd{} (\capSlopeGsmEciFdLo, \capSlopeGsmEciFdHi) & \capSlopeGsmEciDiff{} (\capSlopeGsmEciDiffLo, \capSlopeGsmEciDiffHi) \\
MATH-500 & \capSlopeMathEciReq{} (\capSlopeMathEciReqLo, \capSlopeMathEciReqHi) & \capSlopeMathEciFd{} (\capSlopeMathEciFdLo, \capSlopeMathEciFdHi) & \capSlopeMathEciDiff{} (\capSlopeMathEciDiffLo, \capSlopeMathEciDiffHi) \\
BIG-Bench Hard & \capSlopeBbhEciReq{} (\capSlopeBbhEciReqLo, \capSlopeBbhEciReqHi) & \capSlopeBbhEciFd{} (\capSlopeBbhEciFdLo, \capSlopeBbhEciFdHi) & \capSlopeBbhEciDiff{} (\capSlopeBbhEciDiffLo, \capSlopeBbhEciDiffHi) \\
\bottomrule
\end{tabular}
\caption*{\textbf{Supplementary Table S11.} Slope of chain-of-thought gain on the Epoch Capabilities Index, in points of gain per index point, strict scoring, with 95\% intervals from resampling models. With families resampled, the interval of the difference was \capSlopeGsmEciDiffFamLo{} to \capSlopeGsmEciDiffFamHi{} on GSM8K, \capSlopeMathEciDiffFamLo{} to \capSlopeMathEciDiffFamHi{} on MATH-500 and \capSlopeBbhEciDiffFamLo{} to \capSlopeBbhEciDiffFamHi{} on BIG-Bench Hard.}

\end{table}

\section*{S9. Decoding by sampling}

Five panel models, one per family and each the nearest its family's member to the panel's median size (Gemma 3 12B, Llama 2 13B, Mistral NeMo, Phi-3-medium and Qwen3-14B), answered every MATH-500 question under the first \tempCheckTemplates{} discovery templates in all three conditions, sampled at temperature \tempCheckTemperature{} with one sample per request and every other setting as in the greedy run. Pooled over the five models, sampling minus greedy decoding changed $w$ by \tempCheckWDiffMathPooledPct{} points (95\% interval from resampling questions, \tempCheckWDiffMathPooledPctLo{} to \tempCheckWDiffMathPooledPctHi), the gain over the requested baseline by \tempCheckReqGainDiffMathPooled{} points (\tempCheckReqGainDiffMathPooledLo{} to \tempCheckReqGainDiffMathPooledHi) and the gain over the forced baseline by \tempCheckFdGainDiffMathPooled{} points (\tempCheckFdGainDiffMathPooledLo{} to \tempCheckFdGainDiffMathPooledHi). The interval of the change contained zero for every model's requested-baseline gain (\tempCheckReqGainDiffCiZero{} of \tempCheckModels) and for \tempCheckFdGainDiffCiZero{} of \tempCheckModels{} models' forced-baseline gains. Content scoring gave the same result.

\section*{S10. Five-shot answering within each baseline}

Five demonstrations were also placed before the requested baseline on GSM8K and BIG-Bench Hard, so that the five-shot gain could be measured within each baseline. Mean gain was \fsrGsmGainReq\pp{} within the requested and \fsrGsmGainFd\pp{} within the forced baseline on GSM8K, and \fsrBbhGainReq{} and \fsrBbhGainFd\pp{} on BIG-Bench Hard, where the requested gain was \fsWBbhValGainReqContent\pp{} under content scoring because the demonstrations mostly taught the answer format (strict compliance rose from \fsWBbhCompZsPct\% to \fsWBbhCompFsPct\%). On GSM8K the demonstrations cut $w$ from \fsWGsmWZsPct\% to \fsWGsmWFsPct\%, in \fsWGsmWDownN{} of \fsWGsmM{} models, and the models whose $w$ fell most were those whose five-shot gain changed sign between the baselines (Spearman's $\rho$ \fsWGsmRho, 95\% interval \fsWGsmRhoLo{} to \fsWGsmRhoHi). The two gains ordered \fsrGsmRev{} of \fsrGsmRevPairs{} GSM8K pairs differently (model-resampling interval \fsrGsmRevBootLo--\fsrGsmRevBootHi\%) and \fsrBbhRev{} of \fsrBbhRevPairs{} BIG-Bench Hard pairs (\fsrBbhRevBootLo--\fsrBbhRevBootHi\%), against template-only reversals of \fsrGsmTplReq{} and \fsrBbhTplReq{} within the requested baseline. Choosing under the forced baseline and valuing under the requested one cost \fsrGsmFdToReq{} and \fsrBbhFdToReq\pp{} per chosen model.

\section*{S11. The model most affected by the forced baseline}

Qwen3.6-35B-A3B wrote working in few requested answers ($w$ \wckQwenSixMoeWGsm{} on GSM8K, \wckQwenSixMoeWMath{} on MATH-500) but scored far higher when its reply was begun for it (MATH-500, content scoring: \wckQwenSixMoeMathPooledReqContent\% requested, \wckQwenSixMoeMathPooledFdContent\% forced). Without it, the baseline switch reversed \wckNoQGsmRev{} of \wckNoQGsmRevPairs{} pairs on GSM8K, \wckNoQMathRev{} of \wckNoQMathRevPairs{} on MATH-500 and \wckNoQBbhRev{} of \wckNoQBbhRevPairs{} on BIG-Bench Hard, and the added loss of choosing under the forced baseline and deploying under the requested one was \wckNoQGsmFdToReq, \wckNoQMathFdToReq{} and \wckNoQBbhFdToReq\pp.

\section*{S12. A per-model threshold rule}

Under this rule each model receives chain-of-thought when its one-draw pilot gain exceeds $c$ points, decided once under each baseline from the same pilots as in the main decision analysis and valued on validation templates absent from the discovery draws. A model given chain-of-thought is valued at its gain minus $c$, so that $c$ is what chain-of-thought must earn to be worth giving. Most models gained more than any cutoff tried, so the two baselines decided differently for few models (Table~S12). Valued at the gain itself, without subtracting $c$, deciding under the forced baseline gives chain-of-thought to more models and lost nothing on mathematics when the requested baseline was deployed.

\begin{table}[h]
\centering\small
\begin{tabular}{@{}llccc@{}}
\toprule
 & $c$ & Decisions that differ & Forced to requested & Requested to forced \\
\midrule
GSM8K & 2 & \thrDecGsmTwoDiffPct\% & \thrDecGsmTwoFdToReq{} (\thrDecGsmTwoFdToReqLo, \thrDecGsmTwoFdToReqHi) & \thrDecGsmTwoReqToFd{} (\thrDecGsmTwoReqToFdLo, \thrDecGsmTwoReqToFdHi) \\
 & 5 & \thrDecGsmFiveDiffPct\% & \thrDecGsmFiveFdToReq{} (\thrDecGsmFiveFdToReqLo, \thrDecGsmFiveFdToReqHi) & \thrDecGsmFiveReqToFd{} (\thrDecGsmFiveReqToFdLo, \thrDecGsmFiveReqToFdHi) \\
 & 10 & \thrDecGsmTenDiffPct\% & \thrDecGsmTenFdToReq{} (\thrDecGsmTenFdToReqLo, \thrDecGsmTenFdToReqHi) & \thrDecGsmTenReqToFd{} (\thrDecGsmTenReqToFdLo, \thrDecGsmTenReqToFdHi) \\
MATH-500 & 2 & \thrDecMathTwoDiffPct\% & \thrDecMathTwoFdToReq{} (\thrDecMathTwoFdToReqLo, \thrDecMathTwoFdToReqHi) & \thrDecMathTwoReqToFd{} (\thrDecMathTwoReqToFdLo, \thrDecMathTwoReqToFdHi) \\
 & 5 & \thrDecMathFiveDiffPct\% & \thrDecMathFiveFdToReq{} (\thrDecMathFiveFdToReqLo, \thrDecMathFiveFdToReqHi) & \thrDecMathFiveReqToFd{} (\thrDecMathFiveReqToFdLo, \thrDecMathFiveReqToFdHi) \\
 & 10 & \thrDecMathTenDiffPct\% & \thrDecMathTenFdToReq{} (\thrDecMathTenFdToReqLo, \thrDecMathTenFdToReqHi) & \thrDecMathTenReqToFd{} (\thrDecMathTenReqToFdLo, \thrDecMathTenReqToFdHi) \\
BIG-Bench Hard & 2 & \thrDecBbhTwoDiffPct\% & \thrDecBbhTwoFdToReq{} (\thrDecBbhTwoFdToReqLo, \thrDecBbhTwoFdToReqHi) & \thrDecBbhTwoReqToFd{} (\thrDecBbhTwoReqToFdLo, \thrDecBbhTwoReqToFdHi) \\
 & 5 & \thrDecBbhFiveDiffPct\% & \thrDecBbhFiveFdToReq{} (\thrDecBbhFiveFdToReqLo, \thrDecBbhFiveFdToReqHi) & \thrDecBbhFiveReqToFd{} (\thrDecBbhFiveReqToFdLo, \thrDecBbhFiveReqToFdHi) \\
 & 10 & \thrDecBbhTenDiffPct\% & \thrDecBbhTenFdToReq{} (\thrDecBbhTenFdToReqLo, \thrDecBbhTenFdToReqHi) & \thrDecBbhTenReqToFd{} (\thrDecBbhTenReqToFdLo, \thrDecBbhTenReqToFdHi) \\
\bottomrule
\end{tabular}
\caption*{\textbf{Supplementary Table S12.} Threshold rule. Decisions that differ: share of panel models given chain-of-thought under one baseline only, averaged over pilot sets. Added loss in points of accuracy per panel model, deciding under one baseline and deploying under the other, with 95\% $t$ intervals over validation templates.}

\end{table}

\section*{S13. Larger models}

Six larger models ran the panel's complete design: every template, draw, condition and question of the four benchmarks, \ladRequests{} requests per model (Table~S13). Five extend three of the panel's families. No model of the Mistral~3 release was in the panel, so all three of its models are new; the Qwen3.5 and Llama~3.1 families add one model each above the panel's members of the same release (Qwen3.5-9B and Qwen3.5-35B-A3B; Llama-3.1-8B-Instruct). DeepSeek V4.1-Flash-1017 has no smaller relative in the panel and is placed against the panel on its own. Models are ordered by total parameters within each family; across these steps the active parameters, the architecture and the numerical precision change as well, so a step from one size to the next is not a change of size alone.

\begin{table}[h]\centering\small
\caption*{\textbf{Supplementary Table S13.} The six larger models. Size in billions of parameters. Serving: local GPUs or the provider's application programming interface (API).}
\begin{tabular}{@{}llllll@{}}
\toprule
Model & Family & Size & Architecture & Precision & Serving\\
\midrule
Ministral 3 8B Instruct 2512 & Mistral 3 & \ladSizeMinistralEightB & dense & bfloat16 & local\\
Ministral 3 14B Instruct 2512 & Mistral 3 & \ladSizeMinistralFourteenB & dense & bfloat16 & local\\
Mistral Large 3 (2512) & Mistral 3 & \ladSizeMaxB{} total, \ladSizeMistralLargeActiveB{} active & mixture of experts & not stated & Mistral API\\
Llama-3.1-70B-Instruct & Llama 3.1 & 70 & dense & FP8 & local\\
Qwen3.5-122B-A10B & Qwen3.5 & \ladSizeQwenB{} total, \ladSizeQwenActiveB{} active & mixture of experts & FP8 & local\\
DeepSeek V4.1-Flash-1017 & none & \ladSizeDeepSeekTotalB{} total, \ladSizeDeepSeekActiveInB--\ladSizeDeepSeekActiveOutB{} active & mixture of experts & NVFP4 & Nebius API\\
\bottomrule
\end{tabular}
\par\smallskip\footnotesize Sizes as published by each developer; active parameters for DeepSeek V4.1-Flash: \ladSizeDeepSeekActiveInB{} billion per input token and \ladSizeDeepSeekActiveOutB{} billion per output token.
\end{table}

\subsection*{Runs}
The forced baseline needs the start of the model's reply to be written for it. Locally, the assistant turn was prefilled as for the panel; the Mistral API accepts a prefix for the assistant turn, and the Nebius endpoint accepts a raw completion prompt to which the model's chat template had been applied. Mistral Large 3 requests carried the template's default system message, as the panel's Mistral models and the two smaller Mistral~3 models did. Generation limits were raised for these runs, and every output was then cut, token by token, at the panel's limits; the primary analyses read the cut outputs, so that every model is scored under the same limits, and the uncut reading is given below. Three of the four local models ran with batch-invariant kernels, under which a rerun of \ladEquivRows{} Llama-3.1-70B-Instruct rows reproduced \ladEquivLlamaSame{} of them token for token. Qwen3.5-122B-A10B ran with default kernels, as the panel's Qwen3.5 models did; on \ladEquivRows{} long rows that the run had cut at the panel's limit, a rerun repeated \ladEquivQwenSame{} token for token and gave the same verdict on \ladEquivQwenVerdict. The Nebius endpoint did not repeat greedy outputs token for token, even at temperature zero, with a single candidate token or with a fixed seed, but its repeated chain-of-thought replies gave the same answer in \ladNebiusSameAnswer{} of \ladNebiusPairs{} pairs; each of its outputs is one draw from the served model.

\subsection*{Answer formats}
Asked for a direct answer, Qwen3.5-122B-A10B usually wrote the answer alone, a few tokens long, without the box the prompt asks for (Table~S14). Strict scoring cannot read such replies, so its strict requested accuracy measures the format, and the analyses of the larger models use content scoring, which parses a bare answer.

\begin{table}[h]\centering\small
\caption*{\textbf{Supplementary Table S14.} Zero-shot requested replies of Qwen3.5-122B-A10B that strict scoring cannot read, and those among them that are a bare answer with no box; median length of the bare answers in tokens.}
\begin{tabular}{@{}lrrrr@{}}
\toprule
 & Replies & Unparseable (\%) & Bare answers & Median tokens\\
\midrule
GSM8K & \ladQwenReqRowsGsm & \ladQwenReqUnreadPctGsm & \ladQwenReqBareGsm & \ladQwenReqBareTokensGsm\\
MATH-500 & \ladQwenReqRowsMath & \ladQwenReqUnreadPctMath & \ladQwenReqBareMath & \ladQwenReqBareTokensMath\\
BIG-Bench Hard & \ladQwenReqRowsBbh & \ladQwenReqUnreadPctBbh & \ladQwenReqBareBbh & \ladQwenReqBareTokensBbh\\
HotpotQA & \ladQwenReqRowsHq & \ladQwenReqUnreadPctHq & \ladQwenReqBareHq & \ladQwenReqBareTokensHq\\
\bottomrule
\end{tabular}
\end{table}

On BIG-Bench Hard, the three Mistral~3 models and DeepSeek V4.1-Flash-1017 often wrote the forced answer in a form the strict rule does not accept: the option letter with a backslash or inside a text command, the option's text without its letter, the letter in angle brackets or in bold, or a copy of the instruction's placeholder. Their content-minus-strict gaps in the forced baseline lie above the panel's largest, and content scoring, which parses these forms, carries the comparison. In \ladDsFsNoAnswerPct\% of DeepSeek V4.1-Flash-1017's five-shot forced BIG-Bench Hard replies the box held no answer at all; that condition enters no comparison of the main text.

\subsection*{Flipped pairs}
Table~S15 counts the pairs that the two baselines order differently among the six larger models, among the four of 70 billion parameters or more, and among all models together, with benefits averaged over the discovery draws. Many reversals among the six rest on small differences in gain. Table~S16 counts only the pairs whose two models' gains differ by more than a threshold under both baselines.

\begin{table}[h]\centering\small
\caption*{\textbf{Supplementary Table S15.} Pairs of models ordered differently by chain-of-thought gain against the requested and the forced baseline. Panel: the 29 panel models (28 on BIG-Bench Hard); all: the panel and the six larger models. Share with its 95\% model-resampling interval. Pooled: GSM8K, MATH-500 and BIG-Bench Hard together.}
\setlength{\tabcolsep}{3pt}
\begin{tabular}{@{}lccccll@{}}
\toprule
 & \multicolumn{2}{c}{Six larger models} & \multicolumn{2}{c}{70B or more} & Panel, content & All, content\\
\cmidrule(lr){2-3}\cmidrule(lr){4-5}
 & Content & Strict & Content & Strict & \% (95\% interval) & \% (95\% interval)\\
\midrule
GSM8K & \ladRevGsm{} of \ladPairsGsm & \ladRevGsmStrict & \ladBigRevGsm{} of \ladBigPairsGsm & \ladBigRevGsmStrict & \combPanelRevPctGsm{} (\combPanelRevBootLoGsm--\combPanelRevBootHiGsm) & \combRevPctGsm{} (\combRevBootLoGsm--\combRevBootHiGsm)\\
MATH-500 & \ladRevMath{} of \ladPairsMath & \ladRevMathStrict & \ladBigRevMath{} of \ladBigPairsMath & \ladBigRevMathStrict & \combPanelRevPctMath{} (\combPanelRevBootLoMath--\combPanelRevBootHiMath) & \combRevPctMath{} (\combRevBootLoMath--\combRevBootHiMath)\\
BIG-Bench Hard & \ladRevBbh{} of \ladPairsBbh & \ladRevBbhStrict & \ladBigRevBbh{} of \ladBigPairsBbh & \ladBigRevBbhStrict & \combPanelRevPctBbh{} (\combPanelRevBootLoBbh--\combPanelRevBootHiBbh) & \combRevPctBbh{} (\combRevBootLoBbh--\combRevBootHiBbh)\\
HotpotQA & \ladRevHq{} of \ladPairsHq & \ladRevHqStrict & \ladBigRevHq{} of \ladBigPairsHq & \ladBigRevHqStrict & \combPanelRevPctHq{} (\combPanelRevBootLoHq--\combPanelRevBootHiHq) & \combRevPctHq{} (\combRevBootLoHq--\combRevBootHiHq)\\
Pooled & \ladRevPooled{} of \ladPairsPooled & \ladRevPooledStrict & \ladBigRevPooled{} of \ladBigPairsPooled & \ladBigRevPooledStrict & \combPanelRevPctPooled{} (\combPanelRevBootLoPooled--\combPanelRevBootHiPooled) & \combRevPctPooled{} (\combRevBootLoPooled--\combRevBootHiPooled)\\
\bottomrule
\end{tabular}
\end{table}

\begin{table}[h]\centering\small
\caption*{\textbf{Supplementary Table S16.} Reversed pairs among the pairs whose gains differ by more than 1, 2 or 5\pp{} under both baselines (content scoring): reversed of such pairs.}
\setlength{\tabcolsep}{4pt}
\begin{tabular}{@{}lccccccccc@{}}
\toprule
 & \multicolumn{3}{c}{Six larger models} & \multicolumn{3}{c}{70B or more} & \multicolumn{3}{c}{All models}\\
\cmidrule(lr){2-4}\cmidrule(lr){5-7}\cmidrule(lr){8-10}
 & 1\pp & 2\pp & 5\pp & 1\pp & 2\pp & 5\pp & 1\pp & 2\pp & 5\pp\\
\midrule
GSM8K & \ladGapOneRevGsm/\ladGapOnePairsGsm & \ladGapTwoRevGsm/\ladGapTwoPairsGsm & \ladGapFiveRevGsm/\ladGapFivePairsGsm & \ladBigGapOneRevGsm/\ladBigGapOnePairsGsm & \ladBigGapTwoRevGsm/\ladBigGapTwoPairsGsm & \ladBigGapFiveRevGsm/\ladBigGapFivePairsGsm & \combGapOneRevGsm/\combGapOnePairsGsm & \combGapTwoRevGsm/\combGapTwoPairsGsm & \combGapFiveRevGsm/\combGapFivePairsGsm\\
MATH-500 & \ladGapOneRevMath/\ladGapOnePairsMath & \ladGapTwoRevMath/\ladGapTwoPairsMath & \ladGapFiveRevMath/\ladGapFivePairsMath & \ladBigGapOneRevMath/\ladBigGapOnePairsMath & \ladBigGapTwoRevMath/\ladBigGapTwoPairsMath & \ladBigGapFiveRevMath/\ladBigGapFivePairsMath & \combGapOneRevMath/\combGapOnePairsMath & \combGapTwoRevMath/\combGapTwoPairsMath & \combGapFiveRevMath/\combGapFivePairsMath\\
BIG-Bench Hard & \ladGapOneRevBbh/\ladGapOnePairsBbh & \ladGapTwoRevBbh/\ladGapTwoPairsBbh & \ladGapFiveRevBbh/\ladGapFivePairsBbh & \ladBigGapOneRevBbh/\ladBigGapOnePairsBbh & \ladBigGapTwoRevBbh/\ladBigGapTwoPairsBbh & \ladBigGapFiveRevBbh/\ladBigGapFivePairsBbh & \combGapOneRevBbh/\combGapOnePairsBbh & \combGapTwoRevBbh/\combGapTwoPairsBbh & \combGapFiveRevBbh/\combGapFivePairsBbh\\
HotpotQA & \ladGapOneRevHq/\ladGapOnePairsHq & \ladGapTwoRevHq/\ladGapTwoPairsHq & \ladGapFiveRevHq/\ladGapFivePairsHq & \ladBigGapOneRevHq/\ladBigGapOnePairsHq & \ladBigGapTwoRevHq/\ladBigGapTwoPairsHq & \ladBigGapFiveRevHq/\ladBigGapFivePairsHq & \combGapOneRevHq/\combGapOnePairsHq & \combGapTwoRevHq/\combGapTwoPairsHq & \combGapFiveRevHq/\combGapFivePairsHq\\
Pooled & \ladGapOneRevPooled/\ladGapOnePairsPooled & \ladGapTwoRevPooled/\ladGapTwoPairsPooled & \ladGapFiveRevPooled/\ladGapFivePairsPooled & \ladBigGapOneRevPooled/\ladBigGapOnePairsPooled & \ladBigGapTwoRevPooled/\ladBigGapTwoPairsPooled & \ladBigGapFiveRevPooled/\ladBigGapFivePairsPooled & \combGapOneRevPooled/\combGapOnePairsPooled & \combGapTwoRevPooled/\combGapTwoPairsPooled & \combGapFiveRevPooled/\combGapFivePairsPooled\\
\bottomrule
\end{tabular}
\end{table}

\subsection*{Answers with written working}
Table~S17 gives each model's $w$, its share of requested answers longer than the forced baseline's limit (16 tokens; 64 on HotpotQA). How $w$ changed with size differed between families and benchmarks: in the Llama family it rose on GSM8K and BIG-Bench Hard and fell on MATH-500 and HotpotQA. Table~S18 splits each larger model's requested-minus-forced accuracy into the part from requested answers longer than 16 tokens and the part from the others. The largest positive shifts came from long answers: Llama-3.1-70B-Instruct on GSM8K (\ladShiftLlamaSeventyGsm\pp, of which \ladShiftLongLlamaSeventyGsm\pp{} from long answers) and DeepSeek V4.1-Flash-1017 on MATH-500 (\ladShiftDeepSeekMath\pp, \ladShiftLongDeepSeekMath\pp{} from long answers). The negative shifts of Qwen3.5-122B-A10B lie in its short answers (GSM8K \ladShiftQwenOneTwentyTwoGsm\pp, \ladShiftShortQwenOneTwentyTwoGsm\pp{} from short answers): its bare requested answers scored below its forced answers to the same questions.

\begin{table}[h]\centering\small
\caption*{\textbf{Supplementary Table S17.} $w$ by family and size, discovery draws. Models from the panel are marked (panel).}
\begin{tabular}{@{}llcccc@{}}
\toprule
Family & Model & GSM8K & MATH-500 & BIG-Bench Hard & HotpotQA\\
\midrule
Mistral 3 & Ministral 3 8B & \ladWMinistralEightGsm & \ladWMinistralEightMath & \ladWMinistralEightBbh & \ladWMinistralEightHq\\
 & Ministral 3 14B & \ladWMinistralFourteenGsm & \ladWMinistralFourteenMath & \ladWMinistralFourteenBbh & \ladWMinistralFourteenHq\\
 & Mistral Large 3 & \ladWMistralLargeGsm & \ladWMistralLargeMath & \ladWMistralLargeBbh & \ladWMistralLargeHq\\
Qwen3.5 & 9B (panel) & \ladWQwenNineGsm & \ladWQwenNineMath & \ladWQwenNineBbh & \ladWQwenNineHq\\
 & 35B-A3B (panel) & \ladWQwenThirtyFiveGsm & \ladWQwenThirtyFiveMath & \ladWQwenThirtyFiveBbh & \ladWQwenThirtyFiveHq\\
 & 122B-A10B & \ladWQwenOneTwentyTwoGsm & \ladWQwenOneTwentyTwoMath & \ladWQwenOneTwentyTwoBbh & \ladWQwenOneTwentyTwoHq\\
Llama 3.1 & 8B (panel) & \ladWLlamaEightGsm & \ladWLlamaEightMath & \ladWLlamaEightBbh & \ladWLlamaEightHq\\
 & 70B & \ladWLlamaSeventyGsm & \ladWLlamaSeventyMath & \ladWLlamaSeventyBbh & \ladWLlamaSeventyHq\\
DeepSeek & V4.1-Flash-1017 & \ladWDeepSeekGsm & \ladWDeepSeekMath & \ladWDeepSeekBbh & \ladWDeepSeekHq\\
\midrule
Panel & median & \ladWPanelMedianGsm & \ladWPanelMedianMath & \ladWPanelMedianBbh & \ladWPanelMedianHq\\
 & range & \ladWPanelMinGsm--\ladWPanelMaxGsm & \ladWPanelMinMath--\ladWPanelMaxMath & \ladWPanelMinBbh--\ladWPanelMaxBbh & \ladWPanelMinHq--\ladWPanelMaxHq\\
\bottomrule
\end{tabular}
\end{table}

\begin{table}[h]\centering\small
\caption*{\textbf{Supplementary Table S18.} Requested-minus-forced accuracy (pp, content scoring, discovery draws) with its parts from long and short requested answers, which add to it.}
\setlength{\tabcolsep}{3pt}
\begin{tabular}{@{}lcccc@{}}
\toprule
 & GSM8K & MATH-500 & BIG-Bench Hard & HotpotQA\\
 & total (long, short) & total (long, short) & total (long, short) & total (long, short)\\
\midrule
Ministral 3 8B & \ladShiftMinistralEightGsm{} (\ladShiftLongMinistralEightGsm, \ladShiftShortMinistralEightGsm) & \ladShiftMinistralEightMath{} (\ladShiftLongMinistralEightMath, \ladShiftShortMinistralEightMath) & \ladShiftMinistralEightBbh{} (\ladShiftLongMinistralEightBbh, \ladShiftShortMinistralEightBbh) & \ladShiftMinistralEightHq{} (\ladShiftLongMinistralEightHq, \ladShiftShortMinistralEightHq)\\
Ministral 3 14B & \ladShiftMinistralFourteenGsm{} (\ladShiftLongMinistralFourteenGsm, \ladShiftShortMinistralFourteenGsm) & \ladShiftMinistralFourteenMath{} (\ladShiftLongMinistralFourteenMath, \ladShiftShortMinistralFourteenMath) & \ladShiftMinistralFourteenBbh{} (\ladShiftLongMinistralFourteenBbh, \ladShiftShortMinistralFourteenBbh) & \ladShiftMinistralFourteenHq{} (\ladShiftLongMinistralFourteenHq, \ladShiftShortMinistralFourteenHq)\\
Mistral Large 3 & \ladShiftMistralLargeGsm{} (\ladShiftLongMistralLargeGsm, \ladShiftShortMistralLargeGsm) & \ladShiftMistralLargeMath{} (\ladShiftLongMistralLargeMath, \ladShiftShortMistralLargeMath) & \ladShiftMistralLargeBbh{} (\ladShiftLongMistralLargeBbh, \ladShiftShortMistralLargeBbh) & \ladShiftMistralLargeHq{} (\ladShiftLongMistralLargeHq, \ladShiftShortMistralLargeHq)\\
Llama-3.1-70B & \ladShiftLlamaSeventyGsm{} (\ladShiftLongLlamaSeventyGsm, \ladShiftShortLlamaSeventyGsm) & \ladShiftLlamaSeventyMath{} (\ladShiftLongLlamaSeventyMath, \ladShiftShortLlamaSeventyMath) & \ladShiftLlamaSeventyBbh{} (\ladShiftLongLlamaSeventyBbh, \ladShiftShortLlamaSeventyBbh) & \ladShiftLlamaSeventyHq{} (\ladShiftLongLlamaSeventyHq, \ladShiftShortLlamaSeventyHq)\\
Qwen3.5-122B-A10B & \ladShiftQwenOneTwentyTwoGsm{} (\ladShiftLongQwenOneTwentyTwoGsm, \ladShiftShortQwenOneTwentyTwoGsm) & \ladShiftQwenOneTwentyTwoMath{} (\ladShiftLongQwenOneTwentyTwoMath, \ladShiftShortQwenOneTwentyTwoMath) & \ladShiftQwenOneTwentyTwoBbh{} (\ladShiftLongQwenOneTwentyTwoBbh, \ladShiftShortQwenOneTwentyTwoBbh) & \ladShiftQwenOneTwentyTwoHq{} (\ladShiftLongQwenOneTwentyTwoHq, \ladShiftShortQwenOneTwentyTwoHq)\\
DeepSeek V4.1-Flash & \ladShiftDeepSeekGsm{} (\ladShiftLongDeepSeekGsm, \ladShiftShortDeepSeekGsm) & \ladShiftDeepSeekMath{} (\ladShiftLongDeepSeekMath, \ladShiftShortDeepSeekMath) & \ladShiftDeepSeekBbh{} (\ladShiftLongDeepSeekBbh, \ladShiftShortDeepSeekBbh) & \ladShiftDeepSeekHq{} (\ladShiftLongDeepSeekHq, \ladShiftShortDeepSeekHq)\\
\midrule
Panel range & \ladShiftPanelMinGsm{} to \ladShiftPanelMaxGsm & \ladShiftPanelMinMath{} to \ladShiftPanelMaxMath & \ladShiftPanelMinBbh{} to \ladShiftPanelMaxBbh & \ladShiftPanelMinHq{} to \ladShiftPanelMaxHq\\
\bottomrule
\end{tabular}
\end{table}

\subsection*{Accuracy in each condition}
\begin{table}[ht]
\centering\small
\caption*{\textbf{Supplementary Table S19.} Accuracy of each larger model in each condition and its two gains (points), outputs cut at the panel's limits, means over the discovery draws; the last row of each block is the panel's median.}
\adjustbox{max width=\textwidth}{%
\begin{tabular}{@{}lrrrrrrrrrr@{}}
\toprule
 & \multicolumn{5}{c}{Strict} & \multicolumn{5}{c}{Content}\\
\cmidrule(lr){2-6}\cmidrule(lr){7-11}
Model & Requested & Forced & CoT & CoT $-$ req. & CoT $-$ forced & Requested & Forced & CoT & CoT $-$ req. & CoT $-$ forced\\
\midrule
\multicolumn{11}{@{}l}{\textit{GSM8K}}\\
Ministral 3 8B Instruct 2512 & 23.7 & 23.9 & 93.2 & 69.5 & 69.3 & 24.0 & 24.3 & 94.4 & 70.4 & 70.1\\
Ministral 3 14B Instruct 2512 & 29.3 & 29.1 & 94.2 & 64.9 & 65.1 & 29.5 & 29.4 & 95.1 & 65.6 & 65.7\\
Mistral Large 3 (2512) & 55.7 & 55.5 & 96.4 & 40.6 & 40.9 & 56.5 & 56.2 & 96.8 & 40.4 & 40.6\\
Llama-3.1-70B-Instruct & 54.1 & 33.8 & 94.9 & 40.8 & 61.0 & 56.4 & 34.5 & 95.1 & 38.8 & 60.6\\
Qwen3.5-122B-A10B & 11.4 & 51.2 & 96.8 & 85.4 & 45.6 & 40.2 & 51.2 & 97.0 & 56.8 & 45.8\\
DeepSeek V4.1-Flash-1017 & 70.7 & 68.7 & 96.8 & 26.1 & 28.1 & 71.2 & 69.2 & 97.0 & 25.8 & 27.8\\
\textit{Panel median (29 models)} & \textit{27.9} & \textit{25.9} & \textit{91.9} & \textit{50.0} & \textit{58.9} & \textit{28.7} & \textit{25.9} & \textit{92.1} & \textit{50.6} & \textit{59.0}\\
\midrule
\multicolumn{11}{@{}l}{\textit{MATH-500}}\\
Ministral 3 8B Instruct 2512 & 33.6 & 22.1 & 81.5 & 47.9 & 59.4 & 33.6 & 22.1 & 81.5 & 47.9 & 59.4\\
Ministral 3 14B Instruct 2512 & 32.2 & 26.6 & 83.8 & 51.6 & 57.2 & 32.2 & 26.7 & 83.8 & 51.6 & 57.1\\
Mistral Large 3 (2512) & 49.5 & 43.9 & 90.4 & 40.9 & 46.5 & 49.5 & 43.9 & 90.4 & 40.9 & 46.5\\
Llama-3.1-70B-Instruct & 22.9 & 25.6 & 68.9 & 45.9 & 43.2 & 28.0 & 26.0 & 68.9 & 40.9 & 42.9\\
Qwen3.5-122B-A10B & 11.5 & 55.9 & 93.2 & 81.7 & 37.3 & 46.3 & 55.9 & 93.3 & 47.0 & 37.4\\
DeepSeek V4.1-Flash-1017 & 70.3 & 56.6 & 97.0 & 26.7 & 40.4 & 71.6 & 56.6 & 97.0 & 25.5 & 40.4\\
\textit{Panel median (29 models)} & \textit{24.6} & \textit{22.3} & \textit{77.2} & \textit{31.3} & \textit{43.6} & \textit{29.7} & \textit{22.3} & \textit{77.2} & \textit{31.0} & \textit{43.6}\\
\midrule
\multicolumn{11}{@{}l}{\textit{BIG-Bench Hard}}\\
Ministral 3 8B Instruct 2512 & 51.3 & 17.8 & 77.4 & 26.1 & 59.6 & 52.4 & 48.5 & 82.3 & 29.9 & 33.8\\
Ministral 3 14B Instruct 2512 & 55.3 & 18.1 & 78.5 & 23.2 & 60.4 & 56.7 & 49.7 & 84.9 & 28.2 & 35.2\\
Mistral Large 3 (2512) & 61.6 & 39.3 & 82.5 & 20.8 & 43.1 & 64.8 & 62.7 & 88.3 & 23.5 & 25.7\\
Llama-3.1-70B-Instruct & 49.7 & 57.0 & 76.3 & 26.6 & 19.3 & 64.4 & 64.4 & 85.3 & 20.8 & 20.8\\
Qwen3.5-122B-A10B & 50.0 & 64.4 & 91.4 & 41.5 & 27.1 & 65.1 & 66.2 & 92.8 & 27.7 & 26.7\\
DeepSeek V4.1-Flash-1017 & 46.6 & 19.8 & 88.9 & 42.3 & 69.1 & 71.7 & 65.8 & 92.2 & 20.5 & 26.4\\
\textit{Panel median (28 models)} & \textit{30.5} & \textit{40.8} & \textit{51.9} & \textit{11.8} & \textit{11.2} & \textit{58.8} & \textit{56.0} & \textit{78.6} & \textit{20.5} & \textit{23.7}\\
\midrule
\multicolumn{11}{@{}l}{\textit{HotpotQA}}\\
Ministral 3 8B Instruct 2512 & 54.5 & 51.8 & 54.8 & 0.3 & 3.0 & 54.7 & 51.9 & 55.0 & 0.3 & 3.2\\
Ministral 3 14B Instruct 2512 & 56.5 & 52.8 & 58.0 & 1.5 & 5.2 & 57.0 & 53.0 & 58.2 & 1.2 & 5.2\\
Mistral Large 3 (2512) & 58.8 & 55.4 & 60.4 & 1.6 & 4.9 & 58.8 & 55.5 & 60.4 & 1.6 & 4.9\\
Llama-3.1-70B-Instruct & 56.9 & 60.7 & 62.3 & 5.4 & 1.6 & 60.2 & 60.7 & 62.4 & 2.2 & 1.7\\
Qwen3.5-122B-A10B & 18.8 & 62.9 & 61.6 & 42.8 & $-$1.3 & 61.3 & 62.9 & 64.2 & 2.9 & 1.3\\
DeepSeek V4.1-Flash-1017 & 62.6 & 54.5 & 63.4 & 0.8 & 8.8 & 62.6 & 55.0 & 63.4 & 0.8 & 8.4\\
\textit{Panel median (29 models)} & \textit{50.2} & \textit{55.9} & \textit{58.0} & \textit{3.5} & \textit{0.5} & \textit{53.9} & \textit{55.9} & \textit{58.0} & \textit{1.6} & \textit{0.7}\\
\bottomrule
\end{tabular}
}
\end{table}

\subsection*{Uncut outputs}
Read without the cut at the panel's limits, the counts change little (Table~S20). The largest changes are on MATH-500, where the larger models' chain-of-thought outputs most often ran past the panel's limit.

\begin{table}[h]\centering\small
\caption*{\textbf{Supplementary Table S20.} Reversed pairs with the larger models' outputs read uncut (content scoring unless stated).}
\begin{tabular}{@{}lcccl@{}}
\toprule
 & Six, content & Six, strict & 70B or more & All models, \% (95\% interval)\\
\midrule
GSM8K & \ladRevGsmUncut{} of \ladPairsGsm & \ladRevGsmStrictUncut & \ladBigRevGsmUncut{} of \ladBigPairsGsm & \combRevPctGsmUncut{} (\combRevBootLoGsmUncut--\combRevBootHiGsmUncut)\\
MATH-500 & \ladRevMathUncut{} of \ladPairsMath & \ladRevMathStrictUncut & \ladBigRevMathUncut{} of \ladBigPairsMath & \combRevPctMathUncut{} (\combRevBootLoMathUncut--\combRevBootHiMathUncut)\\
BIG-Bench Hard & \ladRevBbhUncut{} of \ladPairsBbh & \ladRevBbhStrictUncut & \ladBigRevBbhUncut{} of \ladBigPairsBbh & \combRevPctBbhUncut{} (\combRevBootLoBbhUncut--\combRevBootHiBbhUncut)\\
HotpotQA & \ladRevHqUncut{} of \ladPairsHq & \ladRevHqStrictUncut & \ladBigRevHqUncut{} of \ladBigPairsHq & \combRevPctHqUncut{} (\combRevBootLoHqUncut--\combRevBootHiHqUncut)\\
Pooled & \ladRevPooledUncut{} of \ladPairsPooled & \ladRevPooledStrictUncut & \ladBigRevPooledUncut{} of \ladBigPairsPooled & \combRevPctPooledUncut{} (\combRevBootLoPooledUncut--\combRevBootHiPooledUncut)\\
\bottomrule
\end{tabular}
\end{table}

\subsection*{Summary}
Each size is one checkpoint, so the larger models show where each falls against the panel and how often the two baselines order them differently; they give no interval across sizes and no test of a trend.

\section*{S14. The commercial flagship model}

Qwen4-Max, the flagship model that Alibaba serves through its interface, was run with its reasoning mode switched off, through OpenRouter with the provider fixed to Alibaba, at temperature 0, top-$p$ 1 and presence penalty 0. It ran the three conditions on every template of the four benchmarks (\cwCells{} benchmark--template cells) and the same \cwItems{} questions per benchmark (four per task on BIG-Bench Hard), \cwRequests{} requests in all; every request was answered, none was refused and no reasoning tokens were billed. The requested condition sent the panel's request for the final answer as the only message, with the panel's limit. The forced condition sent the same message followed by a reply that begins with \texttt{\textbackslash boxed\{}, which the provider's partial mode lets the model continue, with the panel's limits and, on GSM8K and BIG-Bench Hard, its stop at the first closing brace. The chain-of-thought condition sent the panel's zero-shot request. Outputs were scored with the readers of the panel, and gains and the correction follow the rule of the main text on the discovery draws.

Table~S21 gives the results. The model had the highest requested, forced and chain-of-thought accuracy of all \cwModels{} models on GSM8K and MATH-500. On GSM8K and MATH-500, where $w$ was largest, its gain against the requested baseline was a fraction of its gain against the forced baseline, and multiplying its gain against the forced baseline by $1-w$ came within half a point of it on GSM8K, MATH-500 and HotpotQA. On BIG-Bench Hard the difference was \cwResidBbh{} points under strict scoring and \cwResidBbhContent{} under content scoring. Its outputs at temperature 0 were not fully repeatable: of \cwRepeatPairs{} requested prompts sent twice, \cwRepeatSameSide{} came back on the same side of 16 tokens both times. With one closed model and no access to its weights, these results show how its direct answers behave; they do not enter the comparisons of pairs.

\begin{table}[ht]
\centering
\small
\caption*{\textbf{Supplementary Table S21.} The closed model, discovery draws, strict scoring. Accuracy in \%; $w$: share of requested answers longer than 16 tokens (64 on HotpotQA), \%; gains and the corrected gain $(1-w)\times$(chain-of-thought $-$ forced) in points.}
\begin{tabular}{lrrrr}
\toprule
 & GSM8K & MATH-500 & BIG-Bench Hard & HotpotQA\\
\midrule
Requested accuracy & \cwReqAccGsm & \cwReqAccMath & \cwReqAccBbh & \cwReqAccHq\\
Forced accuracy & \cwFdAccGsm & \cwFdAccMath & \cwFdAccBbh & \cwFdAccHq\\
Chain-of-thought accuracy & \cwCotAccGsm & \cwCotAccMath & \cwCotAccBbh & \cwCotAccHq\\
$w$ & \cwWGsmPct & \cwWMathPct & \cwWBbhPct & \cwWHqPct\\
Gain against requested & \cwReqGainGsm & \cwReqGainMath & \cwReqGainBbh & \cwReqGainHq\\
Gain against forced & \cwFdGainGsm & \cwFdGainMath & \cwFdGainBbh & \cwFdGainHq\\
Corrected gain & \cwCorrGsm & \cwCorrMath & \cwCorrBbh & \cwCorrHq\\
Difference & \cwResidGsm & \cwResidMath & \cwResidBbh & \cwResidHq\\
\bottomrule
\end{tabular}
\end{table}

\bigskip\noindent{\Large\bfseries Supplementary methods}\par\medskip

\section*{S15. Models and prompts}

\paragraph{Panel and prompts.}
The panel holds every eligible checkpoint of five families (Gemma, Llama, Mistral, Phi and Qwen), \dtEligible{} in all, of which \gtM{} were retained on GSM8K and MATH-500 and \bhM{} on BIG-Bench Hard after written exclusion rules applied to training items. Native reasoning modes were switched off identically in every condition through each model's chat template. Prompt banks were generated with PromptSuite \citep{habba2025promptsuite}, varying instruction wording, separators and formatting, with paraphrases kept or dropped by a written rule; the GSM8K bank has \gsmBankConfigs{} configurations and the BIG-Bench Hard bank \bbhBankConfigs. Every configuration was screened for admissibility on training items, and none was excluded. Every condition requests the final answer in a box, the format chosen by a screening rule on training items, and the strict scorer accepts only a boxed answer. Decoding is greedy, with output caps of \capDirect{} tokens in the requested baseline and \capCot{} with chain-of-thought, and a truncated output counts as wrong. Kernels are batch-invariant except for models with Mamba or linear-attention layers, and each model ran on a single GPU type. Each condition received \gtKdisc{} discovery and \gtKval{} validation draws, sampled with replacement from the bank. BIG-Bench Hard uses \bbhItemsPerTask{} items from each of its 25 tasks. Table~S22 lists the panel. The forced baseline reuses the requested baseline's prompts and draws, prefills the assistant turn with the opening of the box, and allows at most 16 new tokens, stopping on GSM8K and BIG-Bench Hard at the first closing brace.

\begin{table*}[t]
\centering\small
\adjustbox{max width=\textwidth}{\begin{tabular}{@{}lllcc@{}}
\toprule
Family & Checkpoint & Revision & GSM8K & BIG-Bench Hard\\
\midrule
Gemma & gemma-2b-it & \texttt{9698841} & \checkmark & \checkmark\\
 & gemma-7b-it & \texttt{9c5798d} & \checkmark & \checkmark\\
 & gemma-3-4b-it & \texttt{093f9f3} & \checkmark & \checkmark\\
 & gemma-2-9b-it & \texttt{11c9b30} & \checkmark & \checkmark\\
 & gemma-2-27b-it & \texttt{aaf20e6} & \checkmark & \checkmark\\
 & gemma-3-12b-it & \texttt{96b6f1e} & \checkmark & \checkmark\\
 & gemma-3-27b-it & \texttt{005ad34} & \checkmark & \checkmark\\
 & gemma-4-26b-a4b-it & \texttt{4d7ae49} & excluded & excluded\\
 & gemma-4-31b-it & \texttt{842da37} & \checkmark & excluded\\
Llama & llama-2-7b-chat-hf & \texttt{f5db02d} & \checkmark & \checkmark\\
 & llama-2-13b-chat-hf & \texttt{a2cb7a7} & \checkmark & \checkmark\\
 & meta-llama-3-8b-instruct & \texttt{8afb486} & \checkmark & \checkmark\\
 & llama-3.1-8b-instruct & \texttt{0e9e39f} & \checkmark & \checkmark\\
Mistral & mistral-7b-instruct-v0.3 & \texttt{c170c70} & \checkmark & \checkmark\\
 & mistral-7b-instruct-v0.1 & \texttt{ec5deb6} & \checkmark & \checkmark\\
 & mistral-nemo-instruct-2407 & \texttt{04d8a90} & \checkmark & \checkmark\\
 & mistral-small-3.1-24b-instruct-2503 & \texttt{68faf51} & \checkmark & \checkmark\\
Phi & phi-3-mini-4k-instruct & \texttt{f39ac1d} & \checkmark & \checkmark\\
 & phi-3-medium-4k-instruct & \texttt{48d87cd} & \checkmark & \checkmark\\
 & phi-4 & \texttt{2db69c1} & \checkmark & \checkmark\\
Qwen & qwen2.5-7b-instruct & \texttt{a09a354} & \checkmark & \checkmark\\
 & qwen3-8b & \texttt{b968826} & \checkmark & \checkmark\\
 & qwen3-30b-a3b & \texttt{ad44e77} & \checkmark & \checkmark\\
 & qwen3-30b-a3b-instruct-2507 & \texttt{0d7cf23} & \checkmark & \checkmark\\
 & qwen3-14b & \texttt{40c0698} & \checkmark & \checkmark\\
 & qwen3-32b & \texttt{9216db5} & \checkmark & \checkmark\\
 & qwen3.5-9b & \texttt{c202236} & \checkmark & \checkmark\\
 & qwen3.5-35b-a3b & \texttt{59d61f3} & \checkmark & \checkmark\\
 & qwen3.6-35b-a3b & \texttt{995ad96} & \checkmark & \checkmark\\
 & qwen3.6-27b & \texttt{6a9e13b} & \checkmark & \checkmark\\
\bottomrule
\end{tabular}}
\caption*{\textbf{Supplementary Table S22.} The panel: checkpoint, pinned revision and retention on each benchmark (GSM8K and MATH-500 share one panel).}

\end{table*}

\paragraph{Native reasoning.}

Of the \thpModels{} panel checkpoints with a switch for their thinking mode, \thpIncludedN{} ran with it switched on; the one left out reached the cap of \thpGsmCap{} output tokens in \thpCapSharePctMinExcluded\% of its pilot outputs on training items, above the threshold of \thpThresholdPct\% (Table~S23). All but one of the \thpIncludedN{} are Qwen checkpoints, so the results are descriptive. The runs reuse the requested baseline's prompts and draws on GSM8K and MATH-500 with greedy decoding, an output cut at the cap counts as wrong, and native reasoning and prompted chain-of-thought are each compared with the requested and the forced baseline. Mean accuracy in the forced, requested, chain-of-thought and native-reasoning conditions was \thiGsmMeanForced, \thiGsmMeanReq, \thiGsmMeanCot{} and \thiGsmMeanNative\% on GSM8K, and \thiMathMeanForced, \thiMathMeanReq, \thiMathMeanCot{} and \thiMathMeanNative\% on MATH-500. The cap cut \thiGsmCutPct\% of native outputs on GSM8K and \thiMathCutPct\% on MATH-500, up to \thiGsmCutPctMax\% and \thiMathCutPctMax\% for single checkpoints, and native reasoning's shortfall is concentrated in the checkpoints with many cut outputs: the others came close to chain-of-thought on GSM8K and exceeded it on MATH-500 (Table~S24). Cut at chain-of-thought's cap of \thiGsmCotCap{} tokens, native reasoning averaged \thiGsmMeanNativeCap\% on GSM8K and \thiMathMeanNativeCap\% on MATH-500.

\begin{table}[t]
\centering\small
\resizebox{\columnwidth}{!}{%
\begin{tabular}{@{}lrrrlrr@{}}
\toprule
& \multicolumn{3}{c}{Pilot outputs at the cap (\%)} & & \multicolumn{2}{c}{Run outputs cut at the cap (\%)}\\
\cmidrule(lr){2-4}\cmidrule(lr){6-7}
Checkpoint & GSM8K & MATH & Both & Item I & GSM8K & MATH\\
\midrule
gemma-4-31b-it & 0.8 & 0.5 & 0.6 & run & 0.3 & 0.9\\
qwen3-14b & 0.0 & 0.8 & 0.4 & run & 0.1 & 0.7\\
qwen3-30b-a3b & 0.0 & 1.0 & 0.5 & run & 0.1 & 0.8\\
qwen3-32b & 0.0 & 1.5 & 0.8 & run & 0.1 & 0.9\\
qwen3-8b & 0.2 & 2.5 & 1.4 & run & 0.1 & 1.4\\
qwen3.5-35b-a3b & 41.5 & 32.8 & 37.1 & run & 42.2 & 34.0\\
qwen3.5-9b & 86.8 & 80.0 & 83.4 & excluded & -- & --\\
qwen3.6-27b & 16.5 & 29.0 & 22.8 & run & 20.9 & 25.1\\
qwen3.6-35b-a3b & 31.2 & 37.8 & 34.5 & run & 27.0 & 38.5\\
\bottomrule
\end{tabular}
}
\caption*{\textbf{Supplementary Table S23.} Thinking-mode pilot on training items: share of outputs at the \thpGsmCap-token cap, and the share of run outputs cut at the cap for the checkpoints that ran.}

\end{table}

\begin{table}[t]
\centering\small
\resizebox{\columnwidth}{!}{%
\begin{tabular}{@{}lrrrrrrrr@{}}
\toprule
& \multicolumn{4}{c}{GSM8K} & \multicolumn{4}{c}{MATH-500}\\
\cmidrule(lr){2-5}\cmidrule(lr){6-9}
Checkpoint & Forced & CoT & Native & Cut (\%) & Forced & CoT & Native & Cut (\%)\\
\midrule
gemma-4-31b-it & 67.5 & 95.8 & 95.5 & 0.3 & 43.9 & 98.0 & 98.6 & 0.9\\
qwen3-14b & 36.1 & 94.8 & 96.3 & 0.1 & 34.5 & 87.8 & 97.5 & 0.7\\
qwen3-30b-a3b & 37.3 & 95.2 & 96.2 & 0.1 & 32.0 & 87.3 & 97.3 & 0.8\\
qwen3-32b & 40.3 & 94.1 & 96.5 & 0.1 & 36.1 & 88.9 & 97.0 & 0.9\\
qwen3-8b & 25.9 & 94.0 & 91.9 & 0.1 & 28.3 & 83.1 & 92.9 & 1.4\\
qwen3.5-35b-a3b & 41.0 & 96.3 & 55.3 & 42.2 & 42.2 & 92.0 & 65.9 & 34.0\\
qwen3.6-27b & 50.1 & 97.4 & 74.5 & 20.9 & 49.3 & 92.9 & 74.9 & 25.1\\
qwen3.6-35b-a3b & 46.5 & 96.7 & 71.0 & 27.0 & 50.2 & 91.9 & 59.6 & 38.5\\
\bottomrule
\end{tabular}
}
\caption*{\textbf{Supplementary Table S24.} Native reasoning per checkpoint: accuracy (\%) with forced direct answering, prompted chain-of-thought (CoT) and native reasoning, and the share of native outputs cut at the cap.}

\end{table}

\section*{S16. Content scoring}

Every output of the three conditions on GSM8K, BIG-Bench Hard and MATH-500 was classified by written extraction rules as correct, wrong, unparseable, a malformed box or truncated, and the content score is one for a correct answer. The rules parse a boxed answer where one exists and otherwise the answer stated in the text. On BIG-Bench Hard, content scoring credited up to \csBbhCotFormatGapMax\pp{} more correct answers than strict scoring for one model with chain-of-thought. On BIG-Bench Hard they accept option text in place of the option letter and the task's own words for an option, and they leave hedged answers and answers outside the options unparseable. The rules were written from the forms of unparsed outputs, without reference to accuracy or gains. A capped output with a complete box is parsed from the box, so every answer that strict scoring counts as correct is correct on content. The same parsing applies to every model, panel and larger model alike. A box that holds only a placeholder (the instruction's \texttt{<answer>} or \texttt{<number>}, another name for the answer's slot such as \texttt{<final answer>} or \texttt{<letter>}, an option template such as \texttt{<Option A>}, or nothing) is not an answer: it is skipped and the reply is parsed without it. Angle brackets around an answer, as in \texttt{<B>} or \texttt{<14>}, are parsed as the answer, and a box written with other brackets, as in \texttt{\textbackslash boxed<Paris>} or \texttt{<boxed>Paris</boxed>}, is parsed as a box. An answer cue at the start of a box, as in \texttt{\textbackslash boxed\{Answer: 12\}}, is not part of the answer, and on GSM8K a clock time such as 2:00~pm is parsed as its hour on the 12-hour clock, as the benchmark writes such answers. On GSM8K and BIG-Bench Hard the forced baseline stops at the first closing brace, so a model that writes \texttt{\textbackslash boxed\{(\textbackslash text\{C\})\}} is stopped inside its own text command; such a reply is parsed as the closed box unless its text ends in a word or sign that asks for more (and, or, to, a comma) or names two options. The one exception is temporal\_sequences, whose options all begin at different times: a box the stop ended after ``to'', as in \texttt{\textbackslash boxed\{6am \textbackslash text\{ to \}}, is parsed as the option that begins at that time. An answer written between \texttt{<answer>} and \texttt{</answer>} tags is parsed as a box. LaTeX spacing commands inside a box are parsed as spaces. On BIG-Bench Hard, a box that opens with an option letter followed by text is parsed as that letter, unless the text is that of another option, in which case the box names two answers and is wrong; an option letter written in lowercase, as in (b), is parsed as its capital; a letter that is not among the item's options is not an answer; on sports\_understanding, the verdict written as one sentence, as in ``The sentence is not plausible.'', is parsed as the answer; and a last sentence that states a condition or a doubt is not parsed as an answer on web\_of\_lies. On GSM8K, a fraction or a mixed number, as in \texttt{\textbackslash frac\{30\}\{2\}}, is parsed as one number with its value. On HotpotQA, typographic quotes, apostrophes and dashes are parsed as their plain forms in the answer and in the reference. On the two larger models served through APIs, whose whole replies were kept, this parsing gave another answer than the box the model closed in \fsCheckDiffer{} of \fsCheckBoxes{} compared boxes. Every prompt of BIG-Bench Hard lettered its options as the benchmark and the scorer do, so the \bbhConflictBoxes{} boxes whose letter and text name different options are the models' own; scoring them by their letter or by their text instead changed none of the BIG-Bench Hard counts of reversed pairs (largest change \bbhConflictMaxRevChange), changed any other count of reversed pairs by \bbhConflictOtherRevMaxChange{} pair at most, and moved no decision cost by more than \bbhConflictDecMaxChange\pp. A box that the model opens and closes inside the prefilled one is parsed as the reply's last complete box. On HotpotQA, content scoring parses the last box or, without one, the answer written outside the box, and scores it by exact match after the official normalization. On MATH-500, only the \mtItems{} items whose reference answer is one rational number are used. Strict scoring parses only the content of the reply's last complete \texttt{\textbackslash boxed\{\}}, found by brace matching; when that box does not hold a number, or the reply has no complete box, the reply has no strict answer. In the forced baseline the prefilled \texttt{\textbackslash boxed\{} is put in front of the model's continuation, and no stop at a closing brace is used, so that fractions such as \texttt{\textbackslash frac\{14\}\{3\}} are not cut; the longest reference answer fits in the 16 tokens under every model's tokenizer. Before parsing, dashes and minus signs are normalized, math delimiters, degree signs, a trailing unit in a text command, a trailing percent sign, LaTeX spacing, a dollar sign and a trailing period are removed. What remains must be an integer, a decimal, a \texttt{\textbackslash frac}, \texttt{\textbackslash dfrac} or \texttt{\textbackslash tfrac} of integers, or a slash fraction of integers; anything else, such as a mixed number, a radical, an expression or an equation, is no answer. Every number parsed is reduced to a fraction, and a reply is correct when it equals the reference exactly, with no numeric tolerance.

Table~S25 gives the class shares, format compliance (the share of outputs in the strict format) and the format gap (content minus strict accuracy).

\begin{table}[t]
\centering\small
\setlength{\tabcolsep}{2.4pt}
\resizebox{\columnwidth}{!}{%
\begin{tabular}{@{}llrrrrrrr@{}}
\toprule
 & Condition & Corr. & Wrong & Unp. & Malf. & Trunc. & Compl. & Gap\\
\midrule
GSM8K & requested & \csGsmReqCorrectPct & \csGsmReqWrongPct & \csGsmReqUnreadablePct & \csGsmReqMalformedPct & \csGsmReqTruncatedPct & \csGsmReqCompliantMean & \csGsmReqFormatGapMean\\
 & forced & \csGsmForcedCorrectPct & \csGsmForcedWrongPct & \csGsmForcedUnreadablePct & \csGsmForcedMalformedPct & \csGsmForcedTruncatedPct & \csGsmForcedCompliantMean & \csGsmForcedFormatGapMean\\
 & CoT & \csGsmCotCorrectPct & \csGsmCotWrongPct & \csGsmCotUnreadablePct & \csGsmCotMalformedPct & \csGsmCotTruncatedPct & \csGsmCotCompliantMean & \csGsmCotFormatGapMean\\
BBH & requested & \csBbhReqCorrectPct & \csBbhReqWrongPct & \csBbhReqUnreadablePct & \csBbhReqMalformedPct & \csBbhReqTruncatedPct & \csBbhReqCompliantMean & \csBbhReqFormatGapMean\\
 & forced & \csBbhForcedCorrectPct & \csBbhForcedWrongPct & \csBbhForcedUnreadablePct & \csBbhForcedMalformedPct & \csBbhForcedTruncatedPct & \csBbhForcedCompliantMean & \csBbhForcedFormatGapMean\\
 & CoT & \csBbhCotCorrectPct & \csBbhCotWrongPct & \csBbhCotUnreadablePct & \csBbhCotMalformedPct & \csBbhCotTruncatedPct & \csBbhCotCompliantMean & \csBbhCotFormatGapMean\\
\bottomrule
\end{tabular}}
\caption*{\textbf{Supplementary Table S25.} Content scoring. Class shares over all scored outputs (\%); compliance: mean share of outputs in the strict format (\%); gap: mean of content minus strict accuracy (pp).}

\end{table}

\section*{S17. Human check of the scorers}

Raters A and B each rated \hraRaterAAnswers{} outputs, \hraShared{} of them rated by both, and rater C (D.A.) rated the \hrtAnswersC{} outputs of rater B. Of the \hrtDisagreeN{} GSM8K and BIG-Bench Hard outputs that strict scoring marked wrong and content scoring marked correct, the raters read \hrtDisagreeContentAgreePct\% as correct (95\% Wilson interval \hrtDisagreeContentAgreePctLo{} to \hrtDisagreeContentAgreePctHi), taking the majority of three on shared outputs and rater C in place of rater B elsewhere; with rater B kept, the share was \hrtWithBDisagreeContentAgreePct\%. On the shared outputs, raters A and C agreed on \hrtAgreePctAC\% (Cohen's $\kappa$ \hrtKappaAC, Gwet's agreement coefficient AC1 \citep{gwet2008computing} \hrtAcOneAC), A and B on \hrtAgreePctAB\% ($\kappa$ \hrtKappaAB, AC1 \hrtAcOneAB), and B and C agreed on \hrtAgreePctBC\% of rater B's outputs ($\kappa$ \hrtKappaBC). On control outputs that both scorers marked wrong, raters A, B and C read \hraRaterACtrlWrongReadAll{} of \hraRaterACtrlWrongNAll, \hraRaterBCtrlWrongReadAll{} of \hraRaterBCtrlWrongNAll{} and \hrtRaterCCtrlWrongReadAll{} of \hrtRaterCCtrlWrongNAll{} as correct.

On the \hrtSharedThree{} outputs rated by all three raters, \hrtMajUnanimous{} verdicts were unanimous and \hrtMajTwoOfThree{} were decided two to one; of the \hrtABDisagree{} outputs on which raters A and B disagreed, rater C sided with A on \hrtABSideA{} and with B on \hrtABSideB. Rater C agreed with content scoring on \hrtRaterCContentAgreePct\% of their GSM8K and BIG-Bench Hard outputs and read none of the control outputs differently from both scorers.

\section*{S18. The split pilot}

Each pilot is one discovery draw and 100 questions from the pilot half of the benchmark's questions (on BIG-Bench Hard, 4 from each task). Every model answers the 100 questions with chain-of-thought. For the baseline, the 100 questions are divided once into 50 scored under the requested and 50 under the forced baseline, the same division for every model in the pilot (on BIG-Bench Hard, two of each task's four questions to each baseline). Each model's estimated gain is the mean, over the 100 questions, of its chain-of-thought score minus its score under the baseline assigned to that question, so it subtracts a baseline accuracy that mixes the two baselines half and half. Chain-of-thought goes to the third of the models with the largest estimated gains; models tied at the cut share the remaining places. Each choice is valued under each deployment baseline on the held-out half of the questions and on validation templates absent from the discovery draws. The two single-baseline pilots use the same draws, questions and chain-of-thought answers with all 100 baseline answers from one baseline, so all three spend the same calls.

\end{document}